\documentclass{article} 
\usepackage{iclr2020_conference,times}

\usepackage{amsmath,amsfonts,bm}

\def\eqref#1{equation~\ref{#1}}

\def\1{\bm{1}}

\DeclareMathAlphabet{\mathsfit}{\encodingdefault}{\sfdefault}{m}{sl}
\SetMathAlphabet{\mathsfit}{bold}{\encodingdefault}{\sfdefault}{bx}{n}

\newcommand{\R}{\mathbb{R}}

\DeclareMathOperator*{\argmin}{arg\,min}

\usepackage{hyperref}
\usepackage{url}

\usepackage{cleveref}       
\usepackage{booktabs}       
\usepackage{amsfonts}       
\usepackage{nicefrac}       
\usepackage{microtype}      

\usepackage{amsmath, amsthm}
\usepackage{amssymb}
\usepackage{mathtools}

\usepackage{algorithm}
\usepackage{algorithmic}

\usepackage{enumitem}
\setlist{noitemsep,topsep=0pt,parsep=0pt,partopsep=0pt,leftmargin=*}

\DeclarePairedDelimiter\norm\lVert\rVert

\theoremstyle{definition}
\newtheorem{definition}{Definition}
\newtheorem{assumption}{Assumption}
\theoremstyle{plain}

\newtheorem{theorem}{Theorem}

\theoremstyle{remark}

\theoremstyle{corollary}
\newtheorem{corollary}{Corollary}

\usepackage{subcaption}

\title{Gap-Aware Mitigation of Gradient Staleness}

\author{
Saar Barkai\thanks{Equal contribution.}\\
Department of Electrical Engineering\\
Technion - Israel Institute of Technology\\
Haifa, Israel \\
\texttt{saarbarkai@gmail.com}\\
\And
Ido Hakimi\footnotemark[1]\\
Department of Computer Science\\
Technion - Israel Institute of Technology\\
Haifa, Israel \\
\texttt{idohakimi@gmail.com}\\
\And
Assaf Schuster\\
Department of Computer Science\\
Technion - Israel Institute of Technology\\
Haifa, Israel \\
\texttt{assaf@sc.technion.ac.il}\\
}

\iclrfinalcopy 
\begin{document}

\maketitle

\begin{abstract}
Cloud computing is becoming increasingly popular as a platform for distributed training of deep neural networks. Synchronous stochastic gradient descent (SSGD) suffers from substantial slowdowns due to stragglers if the environment is non-dedicated, as is common in cloud computing. Asynchronous SGD (ASGD) methods are immune to these slowdowns but are scarcely used due to gradient staleness, which encumbers the convergence process. Recent techniques have had limited success mitigating the gradient staleness when scaling up to many workers (computing nodes). 
In this paper we define the \emph{Gap} as a measure of gradient staleness and propose Gap-Aware (GA), a novel asynchronous-distributed method that penalizes stale gradients linearly to the Gap and performs well even when scaling to large numbers of workers. Our evaluation on the CIFAR, ImageNet, and WikiText-103 datasets shows that GA outperforms the currently acceptable gradient penalization method, in final test accuracy. We also provide convergence rate proof for GA. Despite prior beliefs, we show that if GA is applied, momentum becomes beneficial in asynchronous environments, even when the number of workers scales up.
\end{abstract}

\section{Introduction}
The steady growth of deep neural networks over the years has made it impractical to train them from scratch on a single \emph{worker} (i.e., computational device). Distributing the computations over several workers can drastically reduce the training time. However, due to the sequential nature of the widely used stochastic gradient descent (SGD) method, distributing the process is not an easy task.   

Synchronous SGD (SSGD) is the most common method used to distribute the learning process across multiple workers. Several recent works \citep{fast_imagenet_1,fast_imagenet_2,fast_imagenet_3,facebook1hour} have shown that SSGD can achieve large speedups while maintaining high accuracy. The major drawback of SSGD is that its speed is confined to the slowest worker in every iteration. This shortcoming is magnified in non-dedicated\footnote{An environment of computation nodes who are not specifically optimized to work together} environments such as cloud computing. For this reason, all the above mentioned works were forced to use homogeneous workers in a dedicated network, which serves to reduce the variance in the workers' iteration times. Unlike cloud computing, dedicated networks are expensive and therefore not available to most users.

In asynchronous SGD (ASGD), each worker communicates independently of the others, thereby addressing the major drawback of SSGD. ASGD enjoys linear speedup in terms of the number of workers, even on non-dedicated networks. This makes ASGD a potentially better alternative to SSGD when using cloud computing. Unfortunately, ASGD also has a significant weakness known as \emph{gradient staleness}; the gradients used to update the parameter server's (master) parameters are often based on older parameters and therefore are inaccurate.
Prior works have shown that gradient staleness severely hinders the convergence process by reaching reduced final accuracy \citep{chen2016revisiting,cui2016geeps}. \citet{begets} showed that gradient staleness also induces \emph{implicit momentum}, thus the momentum coefficient $\gamma$ must be decayed when scaling up the number of workers. Most research works measure the gradient staleness of a worker according to the \emph{delay}: the number of master updates since the worker began calculating the stochastic gradient $g$, until $g$ is used to update the master. To overcome gradient staleness, \citet{stale_aware} proposed \emph{Staleness-Aware} (SA), which penalizes the step size of stale gradients linearly to their \emph{delay}. This method was later embraced by other works \citep{jiang2017heterogeneity,edge,giladi2019stability} and is currently the common method for penalizing stale gradients.
Unfortunately, this method suffers from a degradation of the final accuracy, especially when scaling up the number of workers. In \Cref{sec:SA}, we show that the main reason for this degradation is the over-penalization and under-penalization caused by SA.

SSGD and ASGD rely on hyperparameter tuning for every different number of workers \citep{data_parallelism}. Tuning is extremely time-consuming, thus avoiding it is beneficial, whenever possible. 

\textbf{Our contribution:} To mitigate gradient staleness while minimizing the degradation of final accuracy, we define a measure of gradient staleness we refer to as the \emph{Gap}. The Gap is based on the difference between the parameters used to calculate the gradient and the parameters on which the gradient is applied. We propose a new method called \emph{Gap-Aware} (GA) that penalizes the step size of stale gradients linearly to their Gap, while eliminating the over-penalization or under-penalization of SA. No new hyperparameters are introduced using the GA method.
\begin{itemize}
	\item We show that GA out-performs SA, especially as the number of workers scales up.
 	\item We prove that the convergence rate of the GA-ASGD algorithm with a non-convex loss function is consistent with SGD: $\mathcal{O}\left(\frac{1}{\sqrt{BK}}\right)$ where $K$ is the total number of steps and B is the batch size.
 	\item We show that penalizing the gradient itself rather than the step-size, eliminates under-penalization.
 	\item Our results suggest that GA can be used without re-tuning the hyperparameters.
 	\item As opposed to conclusions by \citet{begets}, we show that applying momentum in an asynchronous environment is advantageous (using GA), even when multiple workers are used.
    \item We combine GA with Adam \citep{adam} (Adam-GA), and show that Adam-GA achieves almost two orders of magnitude better perplexity than Adam or Adam-SA (which combines Adam with SA) using several workers on the Transformers-XL model.
\end{itemize}
Our results establish GA as a superior gradient-penalizing option to SA and suggest that using GA is a preferable alternative to SSGD in non-dedicated networks such as cloud computing, even when scaling to large numbers of workers. 
To validate our claims, we performed experiments on the CIFAR10, CIFAR100 \citep{cifar}, ImageNet \citep{imagenet}, and WikiText-103 \citep{wikitext-103} datasets, using several state-of-the-art architectures. A version of GA has reached 72.18\% final test accuracy on the ImageNet dataset using 128 simulated asynchronous workers. As far as we know, this is the largest number of asynchronous workers reported to converge on ImageNet. 

\section{Related Work}
Eliminating gradient staleness is a challenging task and several papers suggested techniques to reduce its detrimental effects.
\citet{dcasgd} proposed DC-ASGD, which uses a Taylor expansion to mitigate the gradient staleness. EASGD \citep{elastic} uses a \emph{center force} to pull the workers' parameters toward the master's parameters. Both DC-ASGD and EASGD achieve high accuracy on small numbers of workers, but fall short when trained on large clusters.
\citet{chan2014distributed} proposed penalizing stale gradients by reducing their size and thus limiting their effect on the learning process. They suggest decaying the learning rate exponentially to the \emph{delay}; this makes the step size arbitrarily small when the amount of workers grows, virtually ceasing the learning process.
As part of their convergence analysis, \citet{dutta2018slow} suggest a penalizing method that is linear to the norm between the master and worker's parameters. However, their method introduces another hyperparameter, which requires additional time to tune. As opposed to other methods, GA performs well even when the number of workers is large, without introducing new hyperparameters.

\citet{gupta2016model} as well as \citet{async_transformer} suggest collecting several gradients before updating the master to reduce the effects of gradient staleness. \citet{terngrad} propose minimizing the size of the gradients to reduce communication times. GA is orthogonal to both of these approaches.

\section{Background}
\label{sec:background}
The goal of an optimization procedure is to minimize $f(\theta)$, where $f$ is a smooth, but not necessarily convex, objective function (a.k.a. loss) and the vector $\theta \in \R^d$ is the model's parameters:

\begin{equation}
  \theta_* = \argmin_{\theta\in \R^d}  f(\theta) := \mathbb{E}_\xi[F(\theta; \xi )]
  \label{eq:main}
\end{equation}
where $\xi \in \Xi$ is a random variable from $\Xi$, the entire set of training samples $\Xi = \{1, 2, \cdots, M\}$. $F(\theta; \xi)$ is the stochastic loss function with respect to the training sample indexed by $\xi$. SGD is commonly the workhorse behind the optimization of deep neural networks. Denoting $\eta$ as the learning rate and $k$ as the step number, SGD's iterative update rule is: $\theta_{k+1}=\theta_k-\eta_k \nabla f(\theta_k)$.
We denote $X_k$ as the variable $X$ at the $k^{th}$ step, where $X$ is any variable.

\paragraph{Momentum}
Momentum \citep{momentum} is a widely adopted optimization technique due to its accelerated convergence and oscillation reduction \citep{sutskever}. 
Instead of simply using the gradient, the momentum iterative update rule\footnote{We consider the version of momentum without dampening} uses an exponentially-weighted moving average of gradients called the \emph{update vector}: $v_{k+1}=\gamma v_k+\nabla f(\theta_k)$. The update rule is: $\theta_{k+1}=\theta_k-\eta_k v_{k+1}$.
\emph{Nesterov's Accelerated Gradient} (NAG) \citep{nesterov} is a well-used variation of momentum that has been proven to achieve quadratic speedup in convergence rate compared to SGD.

\section{Asynchronous SGD (ASGD)}
\label{sec:asgd}
We consider the commonly used ASGD, which operates with a parameter-server (master), used to keep the model's most up-to-date parameters. Each worker maintains a replica of the model's parameters. The workers run in parallel and synchronize with the master independently from each other at the beginning of each batch iteration. We denote $\tau_k$ as the delay at the $k^{th}$ step.
The worker and master algorithms are given by \Cref{alg:NAG-ASGD_worker} and \ref{alg:NAG-ASGD_master}, respectively, where B is the batch-size and $\xi_{k,b}$ denotes the $b^{th}$ sample in the batch sampled at the $k^{th}$ iteration. \emph{Gradient staleness} appears when a gradient $g_k$ is computed on parameters $\theta_{k-\tau_k}$ but applied to different parameters $\theta_{k}$.

\begin{minipage}[t]{0.55\textwidth}
\vspace{-4ex}
\centering
\begin{algorithm}[H]
	\caption{Momentum-ASGD: worker $i$}
	\label{alg:NAG-ASGD_worker}
	\centering
	\begin{algorithmic}
	    \setlength{\itemindent}{-0.5em} 
	    \STATE Always do:
		\STATE \quad Receive parameters $\theta_{k-\tau_k}$ from the master
		\STATE \quad Get $B$ training samples $\xi_{k-\tau_k,[1\dots B]}$
		\STATE \quad Compute gradient: $g^i_k \gets \sum_{b=1}^B\frac{\nabla F(\theta_{k-\tau_k}; \xi_{k-\tau_k,b})}{B}$
		\STATE \quad Send $g^i_k$ to the master
	\end{algorithmic}
\end{algorithm}
\end{minipage}
\hfill
\begin{minipage}[t]{0.42\textwidth}
\vspace{-4ex}
\centering
\begin{algorithm}[H]
\caption{Momentum-ASGD: master}
\label{alg:NAG-ASGD_master}
\begin{algorithmic}
    \setlength{\itemindent}{-0.5em} 
    \STATE For k = 1...K do:
    \STATE \quad Receive gradient $g^i_k$ from worker $i$
    \STATE \quad Update momentum $v_{k+1} \gets \gamma v_k+g^i_k$
    \STATE \quad Update master's weights: \\ \quad $\theta_{k+1} \gets \theta_k-\eta_k \cdot v_{k+1}$
    \STATE \quad Send $\theta_{k+1}$ to worker $i$
\end{algorithmic}
\end{algorithm}
\end{minipage}

\subsection{Staleness-Aware}
\label{sec:SA}
\citet{stale_aware} proposed a gradient penalization method called \emph{Staleness-Aware} (SA). SA aims to reduce the effects of \emph{gradient staleness} by dividing the step size by its corresponding $\tau$ (\Cref{alg:Original_Stale-Aware_master}). The worker algorithm remains unchanged. The two ideas behind SA is that stale gradients should be penalized to reduce their impact and that gradient staleness scales up with $\tau$ . 
SA successfully mitigates the gradient staleness when $N$ is small. Although commonly used, SA can potentially over-penalize as well as under-penalize stale gradients as we show below. 

\begin{minipage}[t]{0.48\textwidth}
\vspace{-4ex}
\centering
\begin{algorithm}[H]
\caption{Staleness-Aware: master}
\label{alg:Original_Stale-Aware_master}
\begin{algorithmic}
    \setlength{\itemindent}{-0.5em} 
    \STATE Initialize an iteration array: $iter = [0]*N$
    \STATE For k = 1...K do:
    \STATE \quad Receive gradient $g^i_k$ from worker $i$
    \STATE \quad Calculate worker $i$'s delay $\tau_k \gets k-iter[i]$
    \STATE \quad Update momentum $v_{k+1} \gets \gamma v_k+g^i_k$
    \STATE \quad Update master $\theta_{k+1} \gets \theta_k-\frac{\eta_k}{\tau_k} v_{k+1}$
    \STATE \quad Send $\theta_{k+1}$ to worker $i$
    \STATE \quad Save current iteration $iter[i] \gets k$
\end{algorithmic}
\end{algorithm}
\end{minipage}
\hfill
\begin{minipage}[t]{0.51\textwidth}
\vspace{-4ex}
\centering
\begin{algorithm}[H]
    \caption{Gap-Aware: master}
    \label{alg:Gap_master}
    \begin{algorithmic}
        \setlength{\itemindent}{-0.5em} 
        \STATE For k = 1...K do:
        \STATE \quad Receive gradient $g_k^i$ from worker $i$
        \STATE \quad Calculate Gap: $G_k = \frac{|\theta_k-\theta_{k-\tau_k}|}{C}+\mathbf{1}^d$
        \STATE \quad Update momentum $v_{k+1} \gets \gamma v_k+\left(\frac{1}{G_k}\right)\odot g^i_k$
        \STATE \quad Update master $\theta_{k+1} \gets \theta_k-\eta_k v_{k+1}$
        \STATE \quad Save and send current parameters 
        \STATE \quad \quad $\theta_{k+1}$ to worker $i$
    \end{algorithmic}
\end{algorithm}
\end{minipage}

\paragraph{Over-Penalizing}
\label{sec:over_penalty}
Let us assume that at some step k, after $\tau$ master updates we get $\theta_k = \theta_{k-\tau}$ just as the gradient calculated on $\theta_{k-\tau}$ is applied. This means there is no gradient staleness for the next gradient update since it was computed using the same parameters on which it is applied. Unfortunately, the delay remains $\tau > 0$, thereby causing over-penalization when SA is used.

Additionally, $\tau$ scales linearly with $N$, which dramatically reduces $\eta$ when $N$ is large. Consequently, on large numbers of workers, the convergence rate of SA is sluggish and its accuracy plummets.

\paragraph{Under-Penalizing}
\label{sec:stale_ver}
The SA method doesn't take into account that when using momentum, the update step also contains past gradients. To emphasize the importance of this issue, let's examine a fictional example: assume that some gradient $g_k$ is very stale ($\tau_k$ is large). Following the SA technique, the update rule is: $\theta_{k+1} = \theta_{k}-\frac{\eta_k}{\tau_k} \cdot v_{k+1} = \theta_{k}-\frac{\eta_k}{\tau_k} \cdot (\gamma v_{k} +g_k)$.
This means the stale gradient $g_k$ is indeed penalized by being multiplied by $\frac{\eta_k}{\tau_{k}}$, which is small; thus $g_k$ doesn't change the parameters much. 
We further assume the next iteration is very fast ($\tau_{k+1}$ is small). The next update will be: $\theta_{k+2}= \theta_{k+1}-\frac{\eta_{k+1}}{\tau_{k+1}} \cdot (\gamma^2 v_{k} +\gamma g_{k} +g_{k+1})$.
The stale gradient $g_{k}$ is multiplied by $\gamma\frac{\eta_{k+1}}{\tau_{k+1}}$, which is large (assuming $\eta_k \approx \eta_{k+1}$). Despite the fact that $g_k$ was stale, it still has a significant impact on the learning process. In other words, $g_k$ is \emph{under-penalized}.

To eliminate this possibility we penalize the stale gradient itself rather than the learning rate.
Using this method, the staleness of each gradient is accounted for within the update vector $v$.

\section{Gap-Aware (GA)}
\label{sec:GA}
In this section we propose a new method called \emph{Gap-Aware} to mitigate over and under-penalization.

\subsection{The Gap as a Measure of Gradient Staleness}
An intuitive method to measure gradient staleness would be: $\norm{\nabla f(\theta_k) - \nabla f (\theta_{k-\tau_k})}$. This essentially measures the difference between the stale gradient and the accurate gradient that is computed on the up-to-date parameters. (Of course, $\nabla f(\theta_k)$ is never calculated in ASGD algorithms.) Commonly used in deep learning is the Lipschitzian gradients assumption:
\begin{equation}
\label{ass:Lip_g}
    \norm{\nabla f\left(x\right) - \nabla f\left(y\right)} \leq L\norm{x-y}, \quad \forall{x}, \forall{y}, L \in \mathbb{R}
\end{equation}
Setting $x = \theta_k, y= \theta_{k-\tau_k}$ into Equation~\ref{ass:Lip_g} we get: $\norm{\nabla f\left(\theta_k\right) - \nabla f\left(\theta_{k-\tau_k}\right)} \leq L\norm{\theta_k-\theta_{k-\tau_k}}$.
This implies that $\norm{\theta_k-\theta_{k-\tau_k}}$ is a valid (and easily calculated) measure of the gradient staleness. This measure also addresses the delay's over-penalization; using the same simple example described in \Cref{sec:over_penalty}, 
the term $\norm{\theta_k-\theta_{k-\tau_k}}$ will now be zero, correctly measuring the gradient staleness.

The learning rate $\eta$, commonly decays as the training progresses. This decay can be viewed as a built-in penalization to reduce variance. 
Following this notion, we suggest reducing the staleness penalization as $\eta$ decays. 
To accommodate all the attributes above, we define the \emph{Gap}: 

\begin{definition} \label{def:gap}
$G_k$, the Gap at the $k^{th}$ step, is defined as the minimal number of updates required to traverse the current distance between the master's and worker's parameters using the maximal learning rate and assuming all gradients have an average norm. $G_k \in \mathbb{R}$ is defined as:
\begin{equation*}
    G_k = \norm{\theta_{k}-\theta_{k-\tau_k}}/{C}+ 1
\end{equation*}
Where $C = \eta_{max}\mathbb{E}_k[\norm{\nabla f(\theta_{k-\tau_k})}]$ is a constant representing the maximal distance the parameters can travel in a single update, given the gradient's norm is the average gradient norm.
\end{definition}

\Cref{def:gap} means that dividing $\eta$ by the Gap produces larger steps than those produced by \emph{SA} 
This allows exploring more distant minimas while still mitigating the gradient staleness. 
Note that $\mathbb{E}[G_k] = \tau_k$ occurs only if all previous $\tau_k$ updates were in the exact same direction, which rarely happens. Empirically, we found that $\mathbb{E}[G_k] < \tau_k$ (See \Cref{sec:delay_gap}).

To mitigate the gradient staleness, while eliminating the over-penalization and under-penalization, we divide the gradients themselves by their respective Gap. 
We refer to this method as \emph{Gap-Aware} (GA).

\subsection{Convergence Analysis}
In this section we provide the outlines of the convergence analysis. The complete proofs are given in \Cref{sec:proofs}. The GA-ASGD update rule (without momentum) is:
\begin{equation} \label{eq:ASGD_update_2}
    \theta_{k+1}=\theta_{k}-\eta_k \left(\frac{1}{G_k}\right)\cdot\sum_{b=1}^B \nabla F(\theta_{k-\tau_k};\xi_{k-\tau_k,b})
\end{equation}
The convergence rate is the speed (or number of steps) at which a convergent sequence approaches its limit.
We follow ideas similar to \citet{asgd_conv} to show that the upper bound of the convergence rate of GA on a non-convex loss function is similar to that of SGD. 
\begin{assumption}
\label{ass:1}
We assume the following, commonly-used assumptions, hold:
\item[$\bullet$ \textbf{Unbiased gradient:}] The stochastic gradient $\nabla F(\theta; \xi)$ is unbiased:
\begin{equation}
\label{ass:unbiased_g}
    \nabla f(\theta) = \mathbb{E}_{\xi} [\nabla F(\theta;\xi)]
\end{equation}
\item[$\bullet$ \textbf{Bounded variance:}] The variance of the stochastic gradient is bounded:
\begin{equation}
\label{ass:bound_var}
   \mathbb{E}_{\xi}[\norm{\nabla F(\theta;\xi) -\nabla f(\theta)}^2] \leq \sigma ^2, \quad \forall \theta
\end{equation}
\item[$\bullet$ \textbf{Lipschitzian gradients:}] See Equation~\ref{ass:Lip_g}. 
\item[$\bullet$ \textbf{Independence:}]  All the random variables $\{\xi_{k,b}\}_{k=1\dots K; b=1\dots B},$ are independent.
\item[$\bullet$ \textbf{Bounded age:}]  All delay variables $\tau_1, ... \tau_K$ are bounded:
\begin{equation}
\label{ass:bound_age}
   \max_k {\tau_k} \leq T
\end{equation}
\end{assumption}

\begin{theorem} 
\label{thm:convergence} Assume that Assumption~\ref{ass:1} holds and the learning rate sequence $\{\eta_k\}_{k=1\cdots K}$ satisfies:
\begin{align}
    \frac{BL\eta_k}{G_k}+ 2B^2L^2T \sum_{t=1}^{T}\frac{\eta_{k+t}^2}{G_{k+t}^2}  \leq 1 \quad \text{for all $k=1,2,...$}
    \label{algo1ass}
\end{align}
We have the following ergodic convergence rate for the iteration of GA-ASGD:
\begin{equation}
\label{eq:thm_1}
    \frac{1}{\sum_{k=1}^K\frac{\eta_k}{G_k}}\sum_{k=1}^K\frac{\eta_k}{G_k} \mathbb{E}(\left\| \nabla f(\theta_{k}) \right\|^2) \leq \frac{\frac{2(f(\theta_1) -f(\theta^*))}{B} + \sum_{k=1}^K\left( \frac{\eta_k}{G_k} + 2B L \sum_{j=k-T}^{k-1} \frac{\eta_j^2}{G_j^2} \right)\frac{\eta_k L\sigma^2}{G_k}}{\sum_{k=1}^K\frac{\eta_k}{G_k}}
\end{equation}
Where $\mathbb{E}[\cdot]$ denotes taking expectation in terms of all random variables.
\end{theorem}
To simplify the upper bound in \Cref{thm:convergence}, we observed that setting the learning rate $\eta_k$ such that the expression $\frac{\eta_k}{G_k}$ is a constant value across all $k$ obtains the following convergence rate:
\begin{corollary} \label{cor:conv}
Assume that Assumption~\ref{ass:1} holds and that $\eta_k$ are set such that $\frac{\eta_k}{G_k}$ is constant for any $k$ as follows:
\begin{align}
    \frac{\eta_k}{G_k} := \eta = \sqrt{\frac{f(\theta_1) - f(\theta^*)} {BLK\sigma^2}}, \quad \forall k \in [1, \dots , K]
    \label{eq:coro_1}
\end{align}
If the maximal delay parameter $T$ satisfies:
\begin{align}
    K \geq \frac{4BL(T+1)^2(f(\theta_1)- f(\theta^*))}{\sigma^2}
    \label{eq:coro_2}
\end{align}
then the output of GA satisfies the following ergodic convergence rate:
\begin{equation}
\begin{split}
    &\min_{k\in \{1, \cdots, K\}}\mathbb{E}[\|\nabla f(\theta_k)\|^2] \leq {\frac{1}{K}}\sum_{k=1}^K 
    \mathbb{E}[\|\nabla f(\theta_k)\|^2] \leq             4\sqrt{\frac{(f(\theta_1)- f(\theta^*))L\sigma^2}{BK}}
\end{split}
\end{equation}
\end{corollary}
\Cref{cor:conv} claims that if the total iterations $K$ is greater than $\mathcal{O}(BT^2)$, the convergence rate achieves $\mathcal{O}({1 / \sqrt{BK}})$, which is consistent with the convergence rate of ASGD presented in \citet{asgd_conv}, and with the convergence rate of SGD.

\subsection{Gap-Aware Versions}
\label{sec:gap_ver}
We explore three ways to measure $G_k$:
\begin{itemize}
\item \textbf{Global:} $G_k \in \mathbb{R}$ as defined in \Cref{def:gap}.
\item \textbf{Layer-wise:} Every layer is penalized differently and independently. $G_k \in \mathbb{R}^P$ where $P$ is the number of layers in the model. We denote  $\mathbf{1^S}$ as an S-dimensional vector of ones. We denote any vector $X_{*,p}$ is the $p^{th}$ layer in the vector $X_*$. Every element in $G_k$ is calculated per-layer:
\begin{equation}
\label{def:layer_wise}
    G_{k,p} = {\norm{\theta_{k,p}-\theta_{k-\tau_k,p}}}/{C_{p}}+ \mathbf{1^p}
\end{equation}
\item \textbf{Parameter-wise:} Every parameter (element in the parameter vector $\theta$) is penalized differently and independently. $G_k \in \mathbb{R}^d$ where $d$ is the number of parameters. We denote $|\cdot|$ on vector $X$, as the absolute value per element of $X$. Every element in $G_k$ is calculated and applied per-element:
\begin{equation}
\label{def:parameter_wise}
    G_{k} = {|\theta_{k}-\theta_{k-\tau_k}|}/{C}+ \mathbf{1^d}
\end{equation}
Where $C \in \mathbb{R}^d$ is also calculated element-wise.
Specifically, $C = \eta_{max}\mathbb{E}_k[|\nabla f(\theta_{k-\tau_k})|]$.
\end{itemize}

\begin{figure*}[t]
    \centering
    \begin{subfigure}[t]{0.4\textwidth}
            \includegraphics[width=\textwidth]{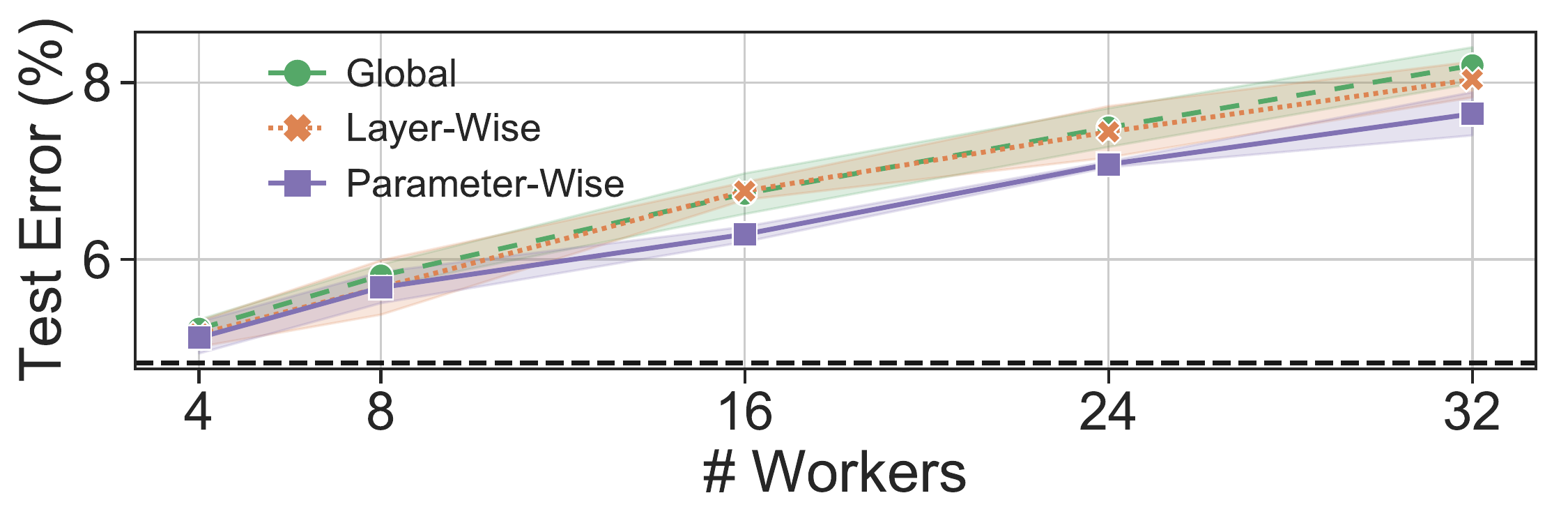}
        	\label{fig:test_gap_CIFAR10wr}
    \end{subfigure}
    \begin{subfigure}[t]{0.4\textwidth}
            \includegraphics[width=\textwidth]{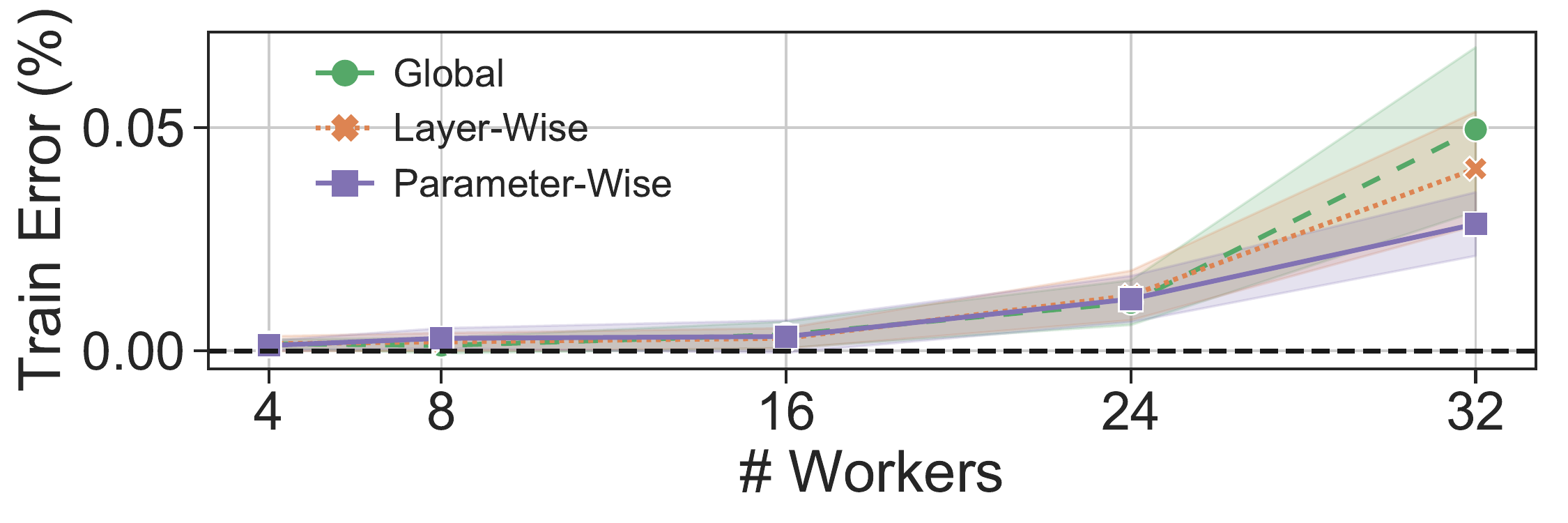}
        	\label{fig:train_gap_CIFAR10wr}
    \end{subfigure}
    \captionsetup{aboveskip=-2pt}
    \captionsetup{belowskip=-2pt}
    \caption{Final test and train error for different numbers of asynchronous workers $N$. The figure shows the average (bold line) and standard deviation (band) of 5 runs on the CIFAR10 dataset using the WideResNet model. The black dashed line is the SGD error using a single worker.}
    \label{fig:gap_versions}
\end{figure*}

We tested these variations on three different frameworks\footnote{A framework is a unique combination of dataset and model. See full experiment in \Cref{sec:gap_ver_exp}.} to determine which technique has the best performance.
\Cref{fig:gap_versions} demonstrates that the \emph{parameter-wise} method (\eqref{def:parameter_wise}) resulted in the best test and train error. Since this phenomenon repeats across all frameworks, we henceforth use the \emph{parameter-wise} method in the Gap-Aware algorithm. We denote $\odot$ as an element-wise multiplication between vectors and describe the final GA algorithm of the master as \Cref{alg:Gap_master}.
The worker algorithm remains the same as in \Cref{alg:NAG-ASGD_worker}.

\section{Experiments}
\label{sec:exp}
We simulated multiple distributed workers\footnote{A worker is a machine with one or more accelerators (i.e., GPU). ASGD methods treat each machine with multiple accelerators, all working synchronously locally, as a single worker.} to measure the final test error, train error, and convergence speed of different cluster sizes. To validate that penalizing linearly to the \emph{Gap} is the factor that leads to better performance, we used the same hyperparameters across all the tested algorithms (see \Cref{sec:hyperparameters}). These hyperparameters are the ones tuned for a single worker, suggested by the authors of the respective papers for each framework. We simulated the workers' execution time using a \emph{gamma-distributed model} \citep{Ali:2000:TET} (see \Cref{sec:gamma_distribution}), where the execution time for each individual batch was drawn from a gamma distribution. The gamma distribution is a well-accepted model for task execution time, which naturally gives rise to stragglers. The importance of asynchronous over synchronous training is explained in \Cref{sec:async_speedup}.

\paragraph{Combining GA with DANA}
One way to verify whether it is better to penalize using the Gap or the delay, is to change the Gap while fixing the delay, and examining the results using GA and SA. Momentum generally increases the norm of the update vector; this in turn, increases the effective step size thus increasing the Gap for a given delay. \citet{dana} introduced DANA, which uses the momentum to estimate the master's parameters at the time of the gradient update, thus decreasing the \emph{Gap}.
The combination of decreasing the \emph{Gap} using DANA and penalizing stale gradients using GA or SA is easily integrated since both methods are orthogonal. In our experiments, we also compared between DANA-Gap-Aware (DANA-GA) and DANA-Staleness-Aware (DANA-SA). We note that since DANA decreases the Gap, DANA-GA penalizes much less than DANA-SA (see \Cref{sec:delay_gap}).

\paragraph{Algorithms}
Our evaluations consist of the following algorithms:
\begin{itemize}
	\item \emph{Baseline:} Single worker with the same tuned hyperparameters as in the respective framework's paper. See \Cref{sec:hyperparameters}.
 	\item \emph{ASGD:} ASGD (\Cref{alg:NAG-ASGD_master}) without momentum ($\gamma = 0$).
	\item \emph{NAG-ASGD:} ASGD with momentum (\Cref{alg:NAG-ASGD_master}) using a NAG optimizer.
	\item \emph{Staleness-Aware:} SA as described in \Cref{alg:Original_Stale-Aware_master}, using a NAG optimizer.
	\item \emph{Gap-Aware:} Parameter-wise GA (\Cref{alg:Gap_master}) as described in Section~\ref{sec:GA}, using a NAG optimizer. 
	\item \emph{DANA:} DANA (\Cref{alg:DANA_master}) as described in \Cref{sec:ap_algorithms}.
	\item \emph{DANA-SA:} DANA-Staleness-Aware (\Cref{alg:DANA-Stale-Aware_master}) as described in \Cref{sec:ap_algorithms}.
	\item \emph{DANA-GA:} DANA-Gap-Aware (\Cref{alg:DANA-Gap-Aware_master}) as described in \Cref{sec:ap_algorithms}.
	\item \emph{Adam:} Adam (\Cref{alg:Adam_master}) as described in \Cref{sec:ap_algorithms}.
	\item \emph{Adam-SA:} Adam-Staleness-Aware (\Cref{alg:Adam-Stale-Aware_master}) as described in \Cref{sec:ap_algorithms}.
	\item \emph{Adam-GA:} Adam-Gap-Aware (\Cref{alg:Adam-Gap-Aware_master}) as described in \Cref{sec:ap_algorithms}.
\end{itemize}

Our evaluation was extensive on image classification tasks such as CIFAR10, CIFAR100 \citep{cifar}, and ImageNet \citep{imagenet}. It also included a language modeling task using the WikiText-103 corpus \citep{wikitext-103}. All datasets and models are detailed in \Cref{sec:data_model}.

\subsection{Evaluation on CIFAR}
\label{sec:exp_cifar}
\begin{figure*}[t]
    \centering
    \begin{subfigure}[t]{0.32\textwidth}
            \includegraphics[width=\textwidth]{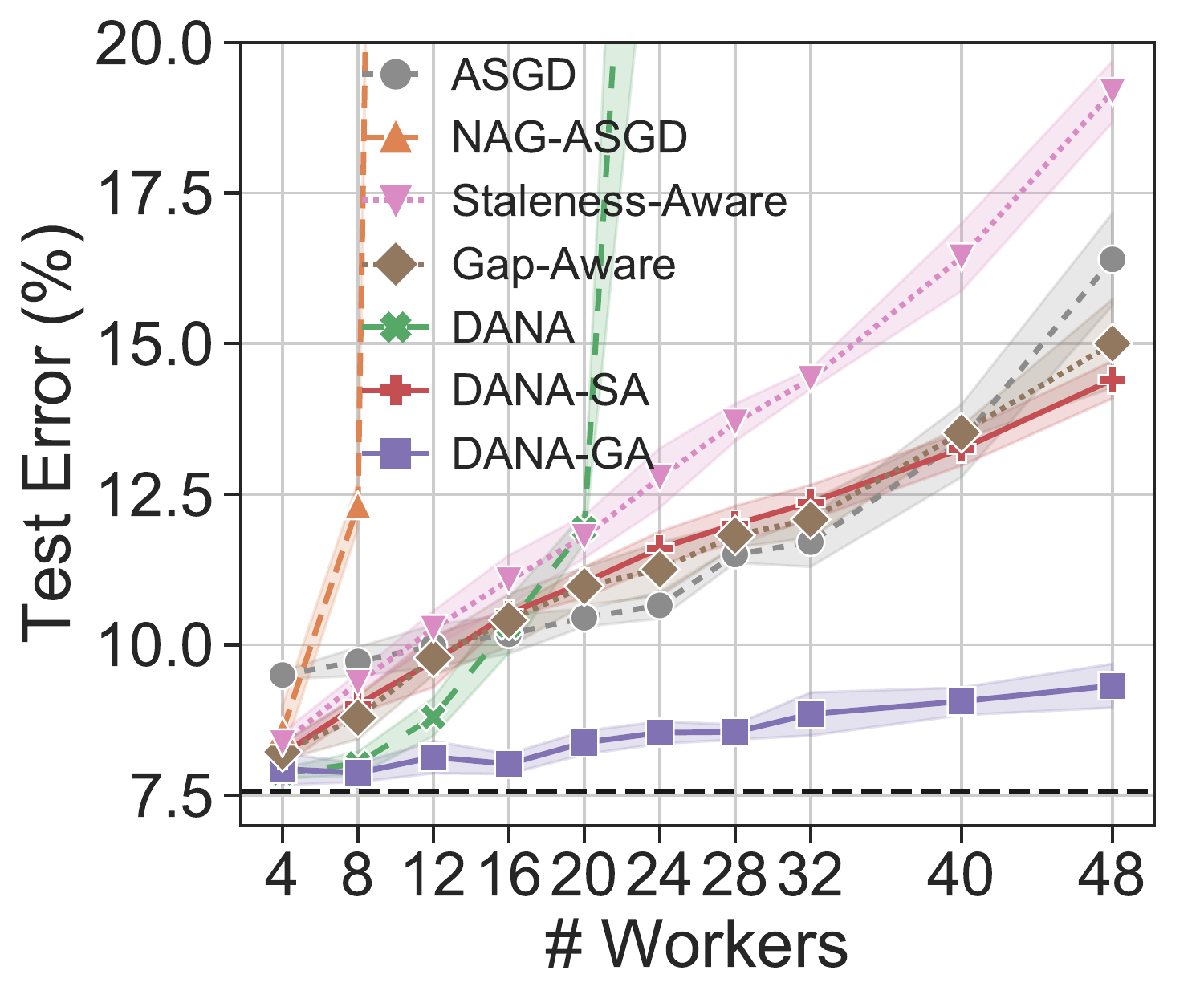}
            \caption{CIFAR10 ResNet-20}
        	\label{fig:test_CIFAR10resnet_all}
    \end{subfigure}
    \begin{subfigure}[t]{0.32\textwidth}
            \includegraphics[width=\textwidth]{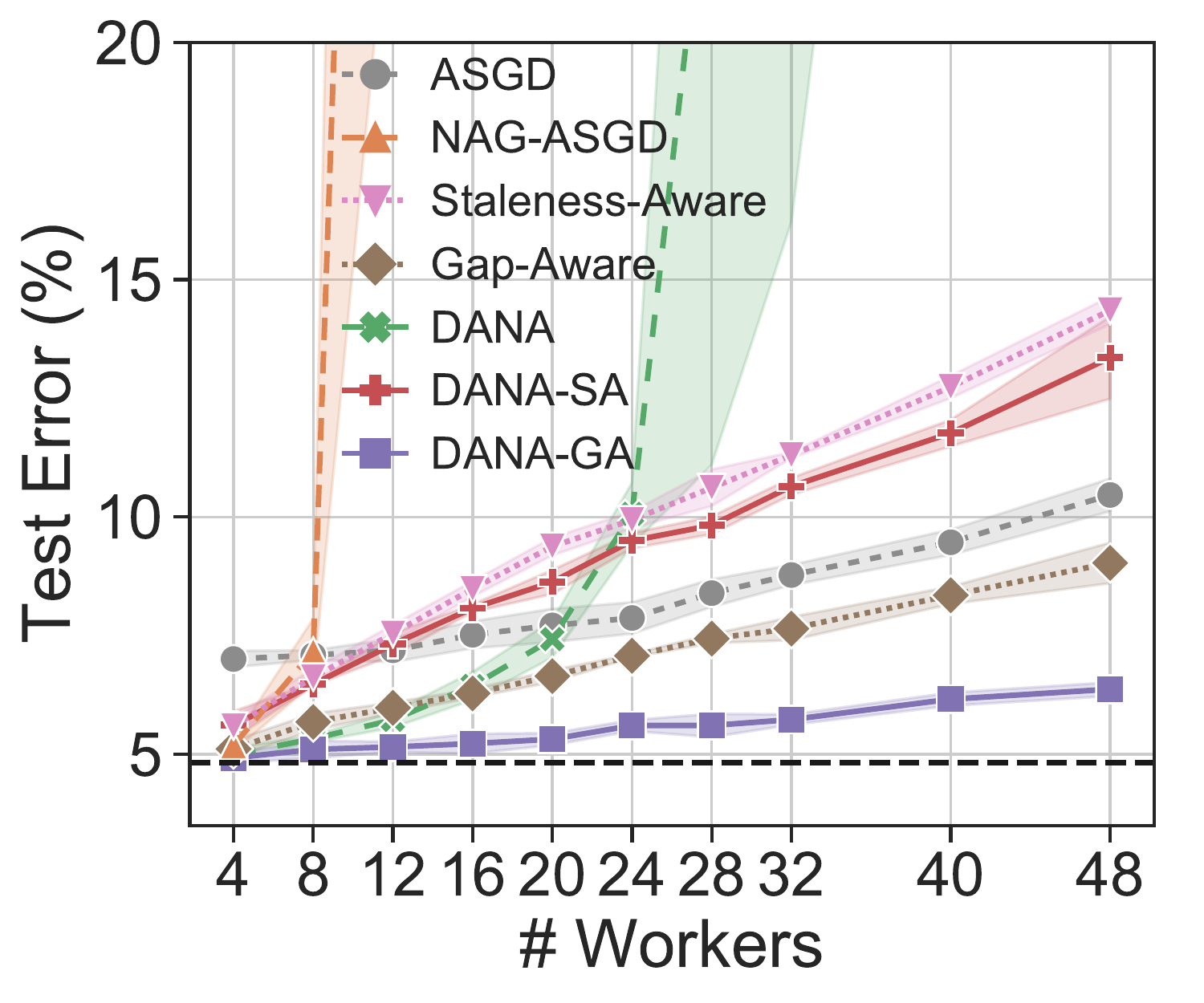}
        	\caption{CIFAR10 WideResNet}
        	\label{fig:test_CIFAR10wr_all}
    \end{subfigure}
        \begin{subfigure}[t]{0.32\textwidth}
            \includegraphics[width=\textwidth]{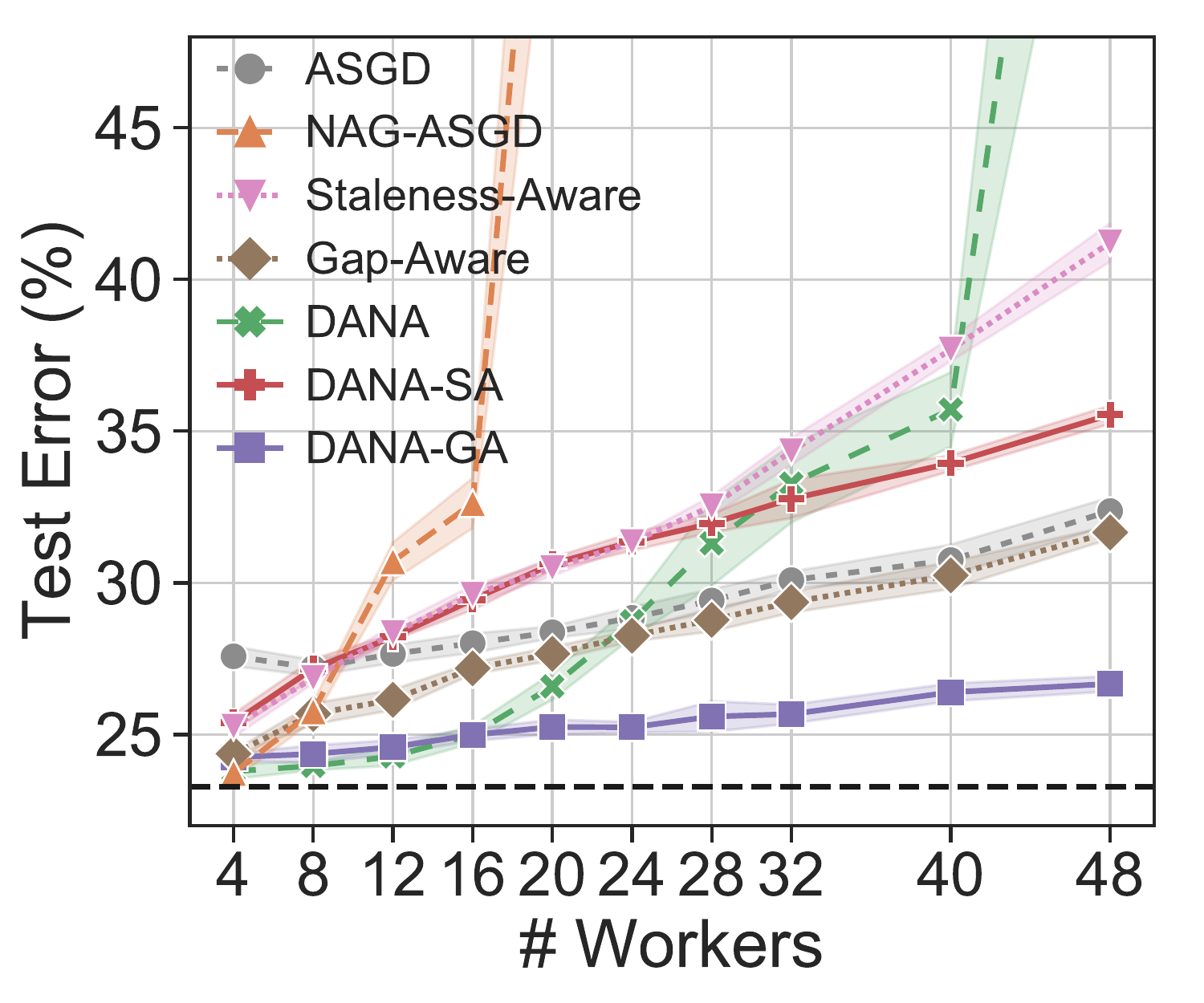}
        	\caption{CIFAR100 WideResNet}
        	\label{fig:test_CIFAR100wr_all}
    \end{subfigure}
    \caption{Final test error for different numbers of asynchronous workers $N$. Each line in the figure represents the average (bold line) and standard deviation (band) of 5 runs on a specific framework. The black dashed line represents the average result of SGD using a single worker.}
    \label{fig:test_all_algos}
\end{figure*}

\paragraph{Gradient Staleness Effects} In \Cref{fig:test_all_algos}, NAG-ASGD shows how gradient staleness is exacerbated by momentum. NAG-ASGD yields high accuracy with few workers, but the test error climbs sharply when more than 16 workers are used. On the other hand, ASGD without momentum performs poorly using few workers. When using many workers, ASGD significantly surpasses NAG-ASGD because of the implicit momentum generated in asynchronous training \citep{begets}.

\paragraph{SA \& GA}\Cref{fig:test_all_algos} also demonstrates that both staleness penalization methods (GA and SA) out-perform the naive NAG-ASGD. GA results in better final test error than SA across all experiments. This empirically proves that GA is the better method for penalizing the gradients. We claim this occurs mainly because SA over-penalizes the gradients, thereby making it impossible to reach any distant, good minima when the number of steps is limited (for more details see \Cref{sec:delay_gap}). 

\paragraph{DANA Versions} \Cref{fig:test_all_algos} shows that DANA potentially diverges when $N$ grows as opposed to DANA-GA and DANA-SA. This shows that DANA benefits from staleness penalization. Furthermore, DANA-GA out-performs all other methods and remains close to the baseline's error across all frameworks. The fact that DANA-GA out-performs DANA-SA validates that GA is superior to SA.

\paragraph{Tuned ASGD}
To validate that the staleness penalization helps overcome the gradient staleness and improve the results, we tuned the momentum and learning rate of ASGD using 32 workers on the 3 frameworks shown in \Cref{fig:test_all_algos}. For each framework, we performed a grid search of 70 perturbations (See \Cref{sec:app_tuned}).
\Cref{tab:tuned-table} shows that GA and DANA-GA, using the same hyperparameters as the baseline, provide similar or better results than tuning both $\gamma$ and $\eta$, which is highly time-consuming. 

\begin{table}[t]
\begin{minipage}[t]{0.56\textwidth}
\centering
\caption[Test accuracy on different frameworks. N=32.]{Test accuracy on different frameworks\footnote{C10/C100=CIFAR10/100, R=ResNet, WR=WideResNet.}. N=32.}
\begin{tabular}{c c c c}
\toprule
Framework & Tuned ASGD  & GA & DANA-GA \\
\midrule
\textbf{C10 R} & 88\% &  87.9\% &   \textbf{91.1\%} \\
\midrule
\textbf{C10 WR} & 91.6\% &  92.3\% &   \textbf{94.3\%} \\
\midrule
\textbf{C100 WR} & 71.1\% &  70.6\% &   \textbf{74.3\%} \\
\bottomrule
\end{tabular}
\label{tab:tuned-table}
\end{minipage}
\hfill
\begin{minipage}[t]{0.43\textwidth}
\centering
\caption{Final test perplexity using Transformer-XL on WikiText-103. (Baseline $24.25$. Lower is better).}
\begin{tabular}{c c c c}
\toprule
N & Adam & Adam-SA & Adam-GA \\
\midrule
4 &  1644.76 &  1210.8 &   \textbf{26.48} \\
8 &  1603.01 &  1129.9 &   \textbf{28.7} \\
\bottomrule
\end{tabular}
\label{tab:NLP}
\end{minipage}
\end{table}

According to \citet{begets}, if momentum is used, the asynchronous implicit momentum should impede the convergence as N increases. However, GA and DANA-GA, which use a large momentum, generally perform better than the tuned ASGD even when $N$ is large. This phenomenon repeats across all frameworks, which suggests that GA, and especially DANA-GA, can mitigate the asynchronous implicit momentum problem.
Tuning these methods should further improve the results.

The graphs of the train error also show the same concepts discussed here regarding the test error and are presented in \Cref{fig:train_all_algos}, \Cref{sec:more_res}. The convergence rate analysis appears in \Cref{sec:conv_rate}.

\subsection{ImageNet Experiments}
\begin{table*}[t]
\centering
\captionof{table}{ResNet-50 ImageNet final test accuracy (Baseline $75.64\%$)}
\begin{tabular}{c c c c c c c c}
\toprule
N & ASGD & NAG-ASGD & SA & GA & DANA & DANA-SA & DANA-GA \\
\midrule
32 & 70.53\% & 70.64\% & 61.73\% & 70.27\%  & 74.89\% &     65.66\% & \textbf{75.06\%}    \\
\midrule
48 &  69.05\% & 66.78\% &  56.22\% & 67.75\% & 73.75\% &     61.16\%  & \textbf{74.23\%}   \\
\midrule
64  & 67.1\% &  59.81\% &  50.79\% & 64.78\% &   69.88\% &     56.98\% & \textbf{74.11\%}   \\ 
\midrule
128 & NaN &  NaN &       NaN & NaN &   NaN &        NaN  & \textbf{72.18\%}   \\ 
\bottomrule
\end{tabular}
\label{tab:imagenet}
\end{table*}

We conducted experiments on the ImageNet dataset using the ResNet-50 model \citep{resnet}. Every asynchronous worker is a machine with 8 GPUs, so the 128 workers in our experiments simulate a total of 1024 GPUs. For reference, \citet{facebook1hour} used 256 GPUs synchronously. The hyperparameters we used are those of the tuned single worker (see \Cref{sec:hyperparameters}). \Cref{tab:imagenet} shows that GA out-performs SA due to the high number of workers, 
which exacerbates the over-penalizing of SA. Unlike SA, GA out-performs NAG-ASGD as $N$ increases due to successful staleness mitigation. DANA-GA remains close to the baseline and better than any other method as $N$ increases. DANA-GA reaches 72.18\% final test accuracy when using 128 workers, which is the most asynchronous workers shown to converge on ImageNet as far as we know.

\subsection{NLP Experiments}
NLP tasks are usually trained using Adam \citep{adam}. To test SA and GA we implemented a version of Adam-SA and Adam-GA given by \Cref{alg:Adam-Stale-Aware_master} and \ref{alg:Adam-Gap-Aware_master}, respectively (\Cref{sec:ap_algorithms}). Transformer-XL \citep{transformer-xl} is a state-of-the-art model for NLP tasks; however, its sensitivity to gradient staleness is catastrophic \citep{async_transformer}. \Cref{tab:NLP} shows that GA successfully mitigates the gradient staleness and achieves near-baseline perplexity while SA results in a higher perplexity by almost two orders of magnitude. In this scenario, SA completely fails to mitigate the gradient staleness, proving the superiority of GA. (See hyperparameters in \Cref{sec:hyperparameters}).

\section{Conclusions}
The goal of this work is to mitigate gradient staleness, one of the main challenges of ASGD. We argue that penalizing stale gradients linearly to the \emph{delay}, as done in the widely used SA method, flounders due to over and under-penalization. We defined the \emph{Gap} to measure gradient staleness and proposed GA, a novel asynchronous distributed technique that mitigates the gradient staleness by penalizing stale gradients linearly to the Gap. We showed that GA surpasses SA across all frameworks, especially in NLP problems or when the number of workers is large. This presents GA as a superior alternative for staleness penalizing. We further introduced DANA-GA and demonstrated that DANA-GA mitigates gradient staleness better than any of the other methods we compared. Despite prior belief, DANA-GA's superb performance enables the use of momentum in asynchronous environments with many workers; it presents a desirable alternative for parallel training with multiple workers, especially on non-dedicated environments such as cloud computing.
In future work, we plan to examine what makes GA perform so well in NLP tasks.

\section*{Acknowledgement}
This work was supported by The Hasso Plattner Institute.

\bibliography{iclr2020_conference}

\begin{thebibliography}{30}
\providecommand{\natexlab}[1]{#1}
\providecommand{\url}[1]{\texttt{#1}}
\expandafter\ifx\csname urlstyle\endcsname\relax
  \providecommand{\doi}[1]{doi: #1}\else
  \providecommand{\doi}{doi: \begingroup \urlstyle{rm}\Url}\fi

\bibitem[Aji \& Heafield(2019)Aji and Heafield]{async_transformer}
Alham~Fikri Aji and Kenneth Heafield.
\newblock Making asynchronous stochastic gradient descent work for
  transformers.
\newblock \emph{arXiv preprint arXiv:1906.03496}, 2019.

\bibitem[Ali et~al.(2000)Ali, Siegel, Maheswaran, Ali, and
  Hensgen]{Ali:2000:TET}
Shoukat Ali, Howard~Jay Siegel, Muthucumaru Maheswaran, Sahra Ali, and Debra
  Hensgen.
\newblock Task execution time modeling for heterogeneous computing systems.
\newblock In \emph{Proceedings of the 9th Heterogeneous Computing Workshop},
  HCW '00, pp.\  185--199, 2000.
\newblock ISBN 0-7695-0556-2.

\bibitem[Chan \& Lane(2014)Chan and Lane]{chan2014distributed}
William Chan and Ian Lane.
\newblock Distributed asynchronous optimization of convolutional neural
  networks.
\newblock In \emph{Fifteenth Annual Conference of the International Speech
  Communication Association}, 2014.

\bibitem[Chen et~al.(2016)Chen, Pan, Monga, Bengio, and
  Jozefowicz]{chen2016revisiting}
Jianmin Chen, Xinghao Pan, Rajat Monga, Samy Bengio, and Rafal Jozefowicz.
\newblock Revisiting distributed synchronous sgd.
\newblock \emph{arXiv preprint arXiv:1604.00981}, 2016.

\bibitem[Cui et~al.(2016)Cui, Zhang, Ganger, Gibbons, and Xing]{cui2016geeps}
Henggang Cui, Hao Zhang, Gregory~R Ganger, Phillip~B Gibbons, and Eric~P Xing.
\newblock Geeps: Scalable deep learning on distributed gpus with a
  gpu-specialized parameter server.
\newblock In \emph{Proceedings of the Eleventh European Conference on Computer
  Systems}, pp.\ ~4. ACM, 2016.

\bibitem[Dai et~al.(2019)Dai, Yang, Yang, Carbonell, Le, and
  Salakhutdinov]{transformer-xl}
Zihang Dai, Zhilin Yang, Yiming Yang, Jaime~G. Carbonell, Quoc~V. Le, and
  Ruslan Salakhutdinov.
\newblock Transformer-xl: Attentive language models beyond a fixed-length
  context.
\newblock \emph{CoRR}, abs/1901.02860, 2019.
\newblock URL \url{http://arxiv.org/abs/1901.02860}.

\bibitem[Dutta et~al.(2018)Dutta, Joshi, Ghosh, Dube, and
  Nagpurkar]{dutta2018slow}
Sanghamitra Dutta, Gauri Joshi, Soumyadip Ghosh, Parijat Dube, and Priya
  Nagpurkar.
\newblock Slow and stale gradients can win the race: Error-runtime trade-offs
  in distributed sgd.
\newblock \emph{arXiv preprint arXiv:1803.01113}, 2018.

\bibitem[Goyal et~al.(2017)Goyal, Doll{\'{a}}r, Girshick, Noordhuis,
  Wesolowski, Kyrola, Tulloch, Jia, and He]{facebook1hour}
Priya Goyal, Piotr Doll{\'{a}}r, Ross~B. Girshick, Pieter Noordhuis, Lukasz
  Wesolowski, Aapo Kyrola, Andrew Tulloch, Yangqing Jia, and Kaiming He.
\newblock Accurate, large minibatch {SGD:} training imagenet in 1 hour.
\newblock \emph{CoRR}, abs/1706.02677, 2017.

\bibitem[Gupta et~al.(2016)Gupta, Zhang, and Wang]{gupta2016model}
Suyog Gupta, Wei Zhang, and Fei Wang.
\newblock Model accuracy and runtime tradeoff in distributed deep learning: A
  systematic study.
\newblock In \emph{2016 IEEE 16th International Conference on Data Mining
  (ICDM)}, pp.\  171--180. IEEE, 2016.

\bibitem[Hakimi et~al.(2019)Hakimi, Barkai, Gabel, and Schuster]{dana}
Ido Hakimi, Saar Barkai, Moshe Gabel, and Assaf Schuster.
\newblock Taming momentum in a distributed asynchronous environment.
\newblock \emph{arXiv preprint arXiv:1907.11612}, 2019.

\bibitem[{Hardy} et~al.(2017){Hardy}, {Le Merrer}, and {Sericola}]{edge}
C.~{Hardy}, E.~{Le Merrer}, and B.~{Sericola}.
\newblock Distributed deep learning on edge-devices: Feasibility via adaptive
  compression.
\newblock In \emph{2017 IEEE 16th International Symposium on Network Computing
  and Applications (NCA)}, pp.\  1--8, Oct 2017.
\newblock \doi{10.1109/NCA.2017.8171350}.

\bibitem[He et~al.(2016)He, Zhang, Ren, and Sun]{resnet}
Kaiming He, Xiangyu Zhang, Shaoqing Ren, and Jian Sun.
\newblock Deep residual learning for image recognition.
\newblock In \emph{2016 {IEEE} Conference on Computer Vision and Pattern
  Recognition}, pp.\  770--778, 2016.

\bibitem[Hinton(2007)]{cifar}
Geoffrey~E Hinton.
\newblock Learning multiple layers of representation.
\newblock \emph{Trends in Cognitive Sciences}, 11\penalty0 (10):\penalty0
  428--434, 2007.

\bibitem[Jiang et~al.(2017)Jiang, Cui, Zhang, and Yu]{jiang2017heterogeneity}
Jiawei Jiang, Bin Cui, Ce~Zhang, and Lele Yu.
\newblock Heterogeneity-aware distributed parameter servers.
\newblock In \emph{Proceedings of the 2017 ACM International Conference on
  Management of Data}, pp.\  463--478. ACM, 2017.

\bibitem[Kingma \& Ba(2015)Kingma and Ba]{adam}
Diederik~P Kingma and Lei Ba.
\newblock J. {ADAM}: a method for stochastic optimization.
\newblock In \emph{International Conference on Learning Representations}, 2015.

\bibitem[Lian et~al.(2015)Lian, Huang, Li, and Liu]{asgd_conv}
Xiangru Lian, Yijun Huang, Yuncheng Li, and Ji~Liu.
\newblock Asynchronous parallel stochastic gradient for nonconvex optimization.
\newblock In \emph{Advances in Neural Information Processing Systems}, pp.\
  2737--2745, 2015.

\bibitem[Merity et~al.(2016)Merity, Xiong, Bradbury, and Socher]{wikitext-103}
Stephen Merity, Caiming Xiong, James Bradbury, and Richard Socher.
\newblock Pointer sentinel mixture models.
\newblock \emph{CoRR}, abs/1609.07843, 2016.
\newblock URL \url{http://arxiv.org/abs/1609.07843}.

\bibitem[Mikami et~al.(2018)Mikami, Suganuma, U.{-}Chupala, Tanaka, and
  Kageyama]{fast_imagenet_1}
Hiroaki Mikami, Hisahiro Suganuma, Pongsakorn U.{-}Chupala, Yoshiki Tanaka, and
  Yuichi Kageyama.
\newblock Imagenet/resnet-50 training in 224 seconds.
\newblock \emph{CoRR}, abs/1811.05233, 2018.
\newblock URL \url{http://arxiv.org/abs/1811.05233}.

\bibitem[Mitliagkas et~al.(2016)Mitliagkas, Zhang, Hadjis, and
  R{\'{e}}]{begets}
Ioannis Mitliagkas, Ce~Zhang, Stefan Hadjis, and Christopher R{\'{e}}.
\newblock Asynchrony begets momentum, with an application to deep learning.
\newblock In \emph{54th Annual Allerton Conference on Communication, Control,
  and Computing}, pp.\  997--1004, 2016.

\bibitem[Nesterov(1983)]{nesterov}
Yurii Nesterov.
\newblock A method of solving a convex programming problem with convergence
  rate o(1/k2).
\newblock In \emph{Soviet Mathematics Doklady}, volume~27, pp.\  372--376,
  1983.

\bibitem[Polyak(1964)]{momentum}
B.T. Polyak.
\newblock Some methods of speeding up the convergence of iteration methods.
\newblock \emph{USSR Computational Mathematics and Mathematical Physics},
  4\penalty0 (5):\penalty0 1 -- 17, 1964.

\bibitem[Russakovsky et~al.(2015)Russakovsky, Deng, Su, Krause, Satheesh, Ma,
  Huang, Karpathy, Khosla, Bernstein, Berg, and Li]{imagenet}
Olga Russakovsky, Jia Deng, Hao Su, Jonathan Krause, Sanjeev Satheesh, Sean Ma,
  Zhiheng Huang, Andrej Karpathy, Aditya Khosla, Michael~S. Bernstein,
  Alexander~C. Berg, and Fei{-}Fei Li.
\newblock {ImageNet} large scale visual recognition challenge.
\newblock \emph{International Journal of Computer Vision}, 115\penalty0
  (3):\penalty0 211--252, 2015.

\bibitem[Shallue et~al.(2018)Shallue, Lee, Antognini, Sohl{-}Dickstein,
  Frostig, and Dahl]{data_parallelism}
Christopher~J. Shallue, Jaehoon Lee, Joseph~M. Antognini, Jascha
  Sohl{-}Dickstein, Roy Frostig, and George~E. Dahl.
\newblock Measuring the effects of data parallelism on neural network training.
\newblock \emph{CoRR}, abs/1811.03600, 2018.
\newblock URL \url{http://arxiv.org/abs/1811.03600}.

\bibitem[Sutskever et~al.(2013)Sutskever, Martens, Dahl, and Hinton]{sutskever}
Ilya Sutskever, James Martens, George~E. Dahl, and Geoffrey~E. Hinton.
\newblock On the importance of initialization and momentum in deep learning.
\newblock In \emph{Proceedings of the 30th International Conference on Machine
  Learning}, pp.\  1139--1147, 2013.

\bibitem[Wen et~al.(2017)Wen, Xu, Yan, Wu, Wang, Chen, and Li]{terngrad}
Wei Wen, Cong Xu, Feng Yan, Chunpeng Wu, Yandan Wang, Yiran Chen, and Hai Li.
\newblock Terngrad: Ternary gradients to reduce communication in distributed
  deep learning.
\newblock In I.~Guyon, U.~V. Luxburg, S.~Bengio, H.~Wallach, R.~Fergus,
  S.~Vishwanathan, and R.~Garnett (eds.), \emph{Advances in Neural Information
  Processing Systems 30}, pp.\  1509--1519. Curran Associates, Inc., 2017.
\newblock URL
  \url{http://papers.nips.cc/paper/6749-terngrad-ternary-gradients-to-reduce-communication-in-distributed-deep-learning.pdf}.

\bibitem[Yamazaki et~al.(2019)Yamazaki, Kasagi, Tabuchi, Honda, Miwa, Fukumoto,
  Tabaru, Ike, and Nakashima]{fast_imagenet_3}
Masafumi Yamazaki, Akihiko Kasagi, Akihiro Tabuchi, Takumi Honda, Masahiro
  Miwa, Naoto Fukumoto, Tsuguchika Tabaru, Atsushi Ike, and Kohta Nakashima.
\newblock Yet another accelerated {SGD:} resnet-50 training on imagenet in 74.7
  seconds.
\newblock \emph{CoRR}, abs/1903.12650, 2019.
\newblock URL \url{http://arxiv.org/abs/1903.12650}.

\bibitem[Ying et~al.(2018)Ying, Kumar, Chen, Wang, and Cheng]{fast_imagenet_2}
Chris Ying, Sameer Kumar, Dehao Chen, Tao Wang, and Youlong Cheng.
\newblock Image classification at supercomputer scale.
\newblock \emph{CoRR}, abs/1811.06992, 2018.
\newblock URL \url{http://arxiv.org/abs/1811.06992}.

\bibitem[Zhang et~al.(2015{\natexlab{a}})Zhang, Choromanska, and
  LeCun]{elastic}
Sixin Zhang, Anna Choromanska, and Yann LeCun.
\newblock Deep learning with elastic averaging {SGD}.
\newblock In \emph{Advances in Neural Information Processing Systems 28: Annual
  Conference on Neural Information Processing Systems}, pp.\  685--693,
  2015{\natexlab{a}}.

\bibitem[Zhang et~al.(2015{\natexlab{b}})Zhang, Gupta, Lian, and
  Liu]{stale_aware}
Wei Zhang, Suyog Gupta, Xiangru Lian, and Ji~Liu.
\newblock Staleness-aware async-sgd for distributed deep learning.
\newblock \emph{CoRR}, abs/1511.05950, 2015{\natexlab{b}}.
\newblock URL \url{http://arxiv.org/abs/1511.05950}.

\bibitem[Zheng et~al.(2017)Zheng, Meng, Wang, Chen, Yu, Ma, and Liu]{dcasgd}
Shuxin Zheng, Qi~Meng, Taifeng Wang, Wei Chen, Nenghai Yu, Zhiming Ma, and
  Tie{-}Yan Liu.
\newblock Asynchronous stochastic gradient descent with delay compensation.
\newblock In \emph{Proceedings of the 34th International Conference on Machine
  Learning}, pp.\  4120--4129, 2017.

\end{thebibliography}
\bibliographystyle{iclr2020_conference}

\newpage
\appendix
\section{Source Code}
The source code of DANA-Gap-Aware is provided via: \\
\href{https://drive.google.com/drive/folders/1z1e_GI-6FZyfROIftoLHqz1X7xvNczWs?usp=sharing}{DOWNLOAD LINK}

\section{Proofs}
\label{sec:proofs}
{\bf Proof for \Cref{thm:convergence}}
\begin{proof}
From the Lipschitzisan gradient assumption \eqref{ass:Lip_g}, we have
  \begin{align}
    \nonumber
    & f(\theta_{k+1}) -f(\theta_{k})  \le \\ 
    \nonumber \le & \langle \nabla f(\theta_{k}), \theta_{k+1} - \theta_{k} \rangle + \frac{L}{2}\|\theta_{k+1} - \theta_{k}\|^2 = \\
    \nonumber = &
        - \left\langle \nabla f(\theta_{k}), \eta_k \left(\frac{1}{G_k}\right)\cdot \sum\limits_{b=1}^{B}\nabla F(\theta_{k-\tau_k}; \xi_{k-\tau_k, b})\right\rangle + \frac{\eta_k^2 L }{2G_k^2} \left\| \sum\limits_{b=1}^{B}\nabla F(\theta_{k-\tau_k}; \xi_{k-\tau_k, b})\right\|^2\\
    = &
        - \frac{B\eta_k}{G_k} \left\langle \nabla f(\theta_{k}), \frac{1}{B}\sum\limits_{b=1}^{B}\nabla F(\theta_{k-\tau_k}; \xi_{k-\tau_k, b})\right\rangle + \frac{\eta_k^2 L }{2G_k^2} \left\| \sum\limits_{b=1}^{B}\nabla F(\theta_{k-\tau_k}; \xi_{k-\tau_k, b})\right\|^2
        \label{eq:proof:con_1}
  \end{align}
  Taking expectation with respect to $\xi_{k, *}$ on both sides of
  \eqref{eq:proof:con_1}, we have
  \begin{align}
    \mathbb{E}_{\xi_{k, *}}[f(\theta_{k+1})] - f(\theta_{k})
    \leq&  - \frac{B\eta_k}{G_k} \left\langle \nabla f(\theta_{k}), \frac{1}{B} \sum_{b=1}^B\nabla f(\theta_{k-\tau_k}) \right\rangle \nonumber\\
        &+ \frac{\eta^2_k L}{2G_k^2}\mathbb{E}_{\xi_k, *}\left[\left\| \sum\limits_{b=1}^{B}\nabla F(\theta_{k-\tau_k}; \xi_{k-\tau_k, b})\right\|^2\right]
    \label{eq:con_2}
  \end{align}
  where we use the unbiased stochastic gradient assumption (\eqref{ass:unbiased_g}).
  
  From the fact that:
  \[
  \langle a ,b \rangle = \frac{1}{2}\left(\|a\|^2 + \|b\|^2 -
    \|a-b\|^2\right)\]
  we have
  \begin{align}
    \label{eq:3}
    &\mathbb{E}_{\xi_{k, *}}[f(\theta_{k+1})] - f(\theta_{k}) \nonumber\\
    \leq& \nonumber - \frac{B\eta_k}{2G_k} \left(\left\| \nabla f(\theta_{k}) \right\|^2 + \left\| \frac{1}{B} \sum_{b=1}^B\nabla f(\theta_{k-\tau_k}) \right\|^2 - \underbrace{\left\|\nabla f(\theta_{k})-  \frac{1}{B} \sum_{b=1}^B\nabla f(\theta_{k-\tau_k}) \right\|^2}_{T_1}\right) \\
    & + \frac{\eta^2_kL}{2G_k^2}\underbrace{\mathbb{E}_{\xi_k,*}\left[\left\|\sum\limits_{b=1}^{B} \nabla F(\theta_{k-\tau_k}; \xi_{k-\tau_k,b}) \right\|^2\right]}_{T_2}
  \end{align}

  Next we estimate the upper bound of $T_1$ and $T_2$. For $T_2$ we have
  \begin{align} T_2 =&\mathbb{E}_{\xi_{k, *}}\left[\left\| \sum\limits_{b=1}^{B}\nabla F\left(\theta_{k-\tau_k}; \xi_{k-\tau_k, b}\right)\right\|^2\right]\nonumber\\
    =& \mathbb{E}_{\xi_{k, *}} \left[ \left\|   \left( \sum_{b=1}^B \left(\nabla F(\theta_{{k-\tau_k}}; \xi_{k-\tau_k, b}) - \nabla f(\theta_{{k-\tau_k}})\right) + \sum_{b=1}^B \nabla f(\theta_{{k-\tau_k}}) \right) \right\|^2\right]\nonumber\\
    =& \mathbb{E}_{\xi_{k, *}} \Bigg[ \left\|   \sum_{b=1}^B \left(\nabla F(\theta_{{k-\tau_k}}; \xi_{k-\tau_k, b}) - \nabla f(\theta_{{k-\tau_k}})\right) \right\|^2 \nonumber + \left \|   \sum_{b=1}^B \nabla f(\theta_{k-\tau_k}) \right\|^2 
    \\&+ 2\left\langle \sum_{b=1}^B
    \left(\nabla F(\theta_{{k-\tau_k}}; \xi_{k-\tau_k, b}) - \nabla f(\theta_{{k-\tau_k}})\right), \sum_{b=1}^B \nabla f(\theta_{{k-\tau_k}}) \right\rangle\Bigg]\nonumber\\
    =&  \mathbb{E}_{\xi_{k, *}} \left[ \left\|   \sum_{b=1}^B \left(\nabla F(\theta_{{k-\tau_k}}; \xi_{k-\tau_k, b}) - \nabla f(\theta_{{k-\tau_k}})\right) \right\|^2 + \left\|   \sum_{b=1}^B \nabla f(\theta_{{k-\tau_k}}) \right\|^2\right]\nonumber\\
    =&  \mathbb{E}_{\xi_{k, *}} \left[ \left\| \sum_{b=1}^B   \left(\nabla F(\theta_{{k-\tau_k}}; \xi_{k-\tau_k, b}) - \nabla f(\theta_{{k-\tau_k}})\right) \right\|^2 + \left\|   \sum_{b=1}^B \nabla f(\theta_{{k-\tau_k}}) \right\|^2\right]\nonumber\\
    =& \mathbb{E}_{\xi_{k, *}} \Bigg[  \sum_{b=1}^B \left\|  \left(\nabla F(\theta_{{k-\tau_k}}; \xi_{k-\tau_k, b})  - \nabla f(\theta_{{k-\tau_k}})\right) \right\|^2 \nonumber\\
    &+  2\sum_{1\le b < b^\prime \le B} \left\langle \nabla F(\theta_{{k-\tau_k}}; \xi_{k-\tau_k, b})  - \nabla f(\theta_{{k-\tau_k}}) , \nabla F(\theta_{{k-\tau_k}}; \xi_{k-\tau_k, {b^\prime}})  - \nabla f(\theta_{{k-\tau_k}}) \right\rangle \nonumber\\
    &+ \left\|   \sum_{b=1}^B \nabla f(\theta_{{k-\tau_k}}) \right\|^2\Bigg]\nonumber\\
    \leq& \mathbb{E}_{\xi_{k, *}} \Bigg[  \sum_{b=1}^B \left( \left\|\left(\nabla F(\theta_{{k-\tau_k}}; \xi_{k-\tau_k, b}) - \nabla f(\theta_{{k-\tau_k}})\right) \right\|^2\right) + \left\|   \sum_{b=1}^B \nabla f(\theta_{{k-\tau_k}}) \right\|^2 \Bigg]\nonumber\\
    \le &B \sigma^2 + \left\| \sum_{b=1}^B \nabla f(\theta_{k-\tau_k}) \right\|^2
    \label{eq:2}
  \end{align}
  where the fifth equality is due to
  \begin{align*}
    &\mathbb{E}_{\xi_{k, *}} \left\langle \sum_{b=1}^B \left(\nabla F(\theta_{{k-\tau_k}}; \xi_{k-\tau_k, b}) - \nabla f(\theta_{{k-\tau_k}})\right), \sum_{b=1}^B \nabla f(\theta_{{k-\tau_k}}) \right \rangle \\
    =&  \left\langle \sum_{b=1}^B \mathbb{E}_{\xi_{k, *}}\left(\nabla F(\theta_{{k-\tau_k}}; \xi_{k-\tau_k, b}) - \nabla f(\theta_{{k-\tau_k}})\right), \sum_{b=1}^B \nabla f(\theta_{{k-\tau_k}}) \right \rangle \\
    =& 0
  \end{align*}
  and the last inequality is due to the bounded variance assumption \eqref{ass:bound_var} and due to:
  \begin{align}
    &\mathbb{E}_{\xi_{k, *}}\left[\sum_{1\le b < b^\prime \le B} \left\langle \nabla F(\theta_{{k-\tau_k}}; \xi_{k-\tau_k, b})  - \nabla f(\theta_{{k-\tau_k}}), \nabla F(\theta_{k-\tau_k}; \xi_{k-\tau_k, {b^\prime}})  - \nabla f(\theta_{k-\tau_k}) \right\rangle\right]\nonumber\\
    =&  \mathbb{E}_{\xi_{k, *}}\left[\sum_{1\le b < b^\prime \le B} \mathbb{E}_{k-\tau_k,b^\prime}\left[ \left\langle \nabla F(\theta_{{k-\tau_k}}; \xi_{k-\tau_k, b})  - \nabla f(\theta_{{k-\tau_k}}), \nabla F(\theta_{k-\tau_k}; \xi_{k-\tau_k, {b^\prime}})  - \nabla f(\theta_{k-\tau_k} \right\rangle\right] \right]\nonumber\\
    =& \mathbb{E}_{\xi_{k, *}}\left[ \sum_{1\le b < b^\prime \le B} \left\langle \nabla F(\theta_{{k-\tau_k}}; \xi_{k-\tau_k, b})  - \nabla f(\theta_{{k-\tau_k}}), \mathbb{E}_{k-\tau_k,b^\prime}[ \nabla F(\theta_{k-\tau_k}; \xi_{k-\tau_k, {b^\prime}})  - \nabla f(\theta_{k-\tau_k})] \right\rangle \right]\nonumber\\
    =& 0.
       \label{cross10}
  \end{align}

  We next turn to $T_1$:

  \begin{align*}
    T_1  =& \left\|\nabla f(\theta_{k})- \frac{1}{B} \sum\limits_{b=1}^{B}\nabla f\left(\theta_{k-\tau_k}\right)\right\|^2 \\
    = & \left\|\nabla f(\theta_{k})- \nabla f\left(\theta_{k-\tau_k}\right)\right\|^2\\
    \leq & L^2 \|\theta_{k} - \theta_{{k-\tau_k}}\|^2 \\
  \end{align*}
  where the last inequality is from the Lipschitzian gradient
  assumption (\eqref{ass:Lip_g}). It follows that
  \begin{align}
    T_1 \leq& L^2\left\|\theta_{k} - \theta_{k-\tau_k}\right\|^2\nonumber\\
    =& L^2 \left\| \sum_{j=k-\tau_k}^{k-1} \left(\theta_{j+1} - \theta_j\right) \right\|^2\nonumber\\ 
    =& L^2 \left\|\sum_{j=k-\tau_k}^{k-1} \eta_j \left(\frac{1}{G_j}\right)\cdot \sum\limits_{b=1}^{B}\nabla F \left(\theta_{j-\tau_j}; \xi_{j-\tau_j,b}\right)\right\|^2 \nonumber\\
    =& L^2  \left\|\sum_{j=k-\tau_k}^{k-1}\eta_j \left(\frac{1}{G_j}\right)\cdot \sum\limits_{b=1}^{B} \left[\nabla F \left(\theta_{j-\tau_j}; \xi_{j-\tau_j,b}\right)- \nabla f\left(\theta_{j-\tau_j}\right) \right] + \sum_{j=k-\tau_k}^{k-1}  \eta_j \left(\frac{1}{G_j}\right)\cdot \sum\limits_{b=1}^{B}\nabla f\left(\theta_{j-\tau_j}\right) \right\|^2\nonumber \\
    \leq &2L^2  \left( \underbrace{\left\|\sum_{j=k-\tau_k}^{k-1} \eta_j \left(\frac{1}{G_j}\right)\cdot \sum\limits_{b=1}^{B} \left[\nabla F \left(\theta_{j-\tau_j}; \xi_{j-\tau_j,b}\right)- \nabla f\left(\theta_{j-\tau_j}\right) \right]\right\|^2}_{T_3} + \underbrace{\left\|\sum_{j=k-\tau_k}^{k-1} \eta_j \left(\frac{1}{G_j}\right)\cdot \sum\limits_{b=1}^{B}\nabla f\left(\theta_{j-\tau_j}\right) \right\|^2}_{T_4} \right)
           \label{eq:t1}
  \end{align}
  where the last inequality uses the fact that $\|a + b\|^2 \le 2\|a\|^2 + 2\|b\|^2$ for any real vectors $a$ and $b$. Taking the expectation in terms of $\{\xi_{j-\tau_j,*}|j\in \{k-\tau_k, ..., k-1\}\}$ for $T_3$, we have
  \begin{align}
    \nonumber    &\mathbb{E}_{\xi_{j-\tau_j,*}|j\in \{k-\tau_k, ..., k-1\}}(T_3)
    \\      =&  \mathbb{E}_{\xi_{j-\tau_j,*}|j\in \{k-\tau_k, ..., k-1\}}\left[ \left\|\sum_{j=k-\tau_k}^{k-1}\eta_j \left(\frac{1}{G_j}\right)\cdot \sum\limits_{b=1}^{B} \left(\nabla F(\theta_{j-\tau_j}; \xi_{j-\tau_j,b}) - \nabla f(\theta_{j-\tau_j})\right )\right\|^2\right]
               \nonumber \\
    =& \mathbb{E}_{\xi_{j-\tau_j,*}|j\in \{k-\tau_k, ..., k-1\}}\left[  \sum_{j=k-\tau_k}^{k-1}\frac{\eta_j^2}{G_j^2} \left\| \sum\limits_{b=1}^{B}\left(\nabla F (\theta_{j-\tau_j}; \xi_{j-\tau_j,b}) - \nabla f(\theta_{j-\tau_j})\right )\right\|^2 \right]
       \nonumber \\
            &+2\mathbb{E}_{\xi_{j-\tau_j,*}|j\in \{k-\tau_k, ..., k-1\}}\Bigg[\nonumber \sum_{k-1\ge j^{\prime\prime}
            >
            j^\prime\ge k-\tau_k}\frac{\eta_{j^{\prime}}\eta_{j^{\prime\prime}}}{G_j^{\prime} G_j^{\prime\prime}} \Bigg\langle \sum\limits_{b=1}^{B}\left(\nabla F (\theta_{j^{\prime\prime}- \tau_{j^{\prime\prime}}}; \xi_{j^{\prime\prime}-
            \tau_{j^{\prime\prime}},b}) - \nabla f(\theta_{j^{\prime\prime}-
            \tau_{j^{\prime\prime}} })\right) ,\nonumber\\
                 & \sum\limits_{b=1}^{B}\left(\nabla F (\theta_{j^\prime-
                \tau_{j^{\prime}}}; \xi_{j^\prime-
                \tau_{j^{\prime}},b}) - \nabla f(\theta_{j^{\prime}-
                \tau_{j^{\prime}}})\right) \Bigg\rangle \Bigg]
                   \nonumber\\
    =& \mathbb{E}_{\xi_{j-\tau_j,*}|j\in \{k-\tau_k, ..., k-1\}}\left[  \sum_{j=k-\tau_k}^{k-1}\frac{\eta_j^2}{G_j^2} \left\| \sum\limits_{b=1}^{B}\left( \nabla F (\theta_{j-\tau_j}; \xi_{j-\tau_j,b}) - \nabla f(\theta_{j-\tau_j})\right) \right\|^2 \right]
       \nonumber \\
    =& \mathbb{E}_{\xi_{j-\tau_j,*}|j\in \{k-\tau_k, ..., k-1\}}\left[  \sum_{j=k-\tau_k}^{k-1}\frac{\eta_j^2}{G_j^2} \sum\limits_{b=1}^{B}\left\| \left(\nabla F (\theta_{j-\tau_j}; \xi_{j-\tau_j,b}) - \nabla f(\theta_{j-\tau_j})\right) \right\|^2 \right]
       \nonumber \\
    \leq & \mathbb{E}_{\xi_{j-\tau_j,*}|j\in \{k-\tau_k, ..., k-1\}}\left[ \sum_{j=k-\tau_k}^{k-1} \frac{B\eta_j^2}{G_j^2} \left\| \left(\nabla F (\theta_{j-\tau_j}; \xi_{j-\tau_j,b}) - \nabla f(\theta_{j-\tau_j})\right) \right\|^2 \right]
       \nonumber \\
    \leq & B\sum_{j=k-\tau_k}^{k-1}\frac{\eta_j^2}{G_j^2}\sigma^2
           \label{eq:t3}
  \end{align}
  where the second to last equality is due to the last lines in \eqref{eq:2} and the
  third equality is due to
  \begin{align*}
    &\mathbb{E}_{\xi_j,k -1 \ge
      j
      \ge
      k-\tau_k} \Bigg[\sum_{k+\tau_{k} -1\ge
      j^{\prime\prime}
      >
      j^\prime\ge
      k}\frac{\eta_{j^{\prime}}\eta_{j^{\prime\prime}}}{G_j^{\prime} G_j^{\prime\prime}} \Bigg\langle \sum\limits_{b=1}^{B}\left(\nabla F (\theta_{j^{\prime\prime}-
      \tau_{j^{\prime\prime} }}; \xi_{j^{\prime\prime}-
      \tau_{j^{\prime\prime}},b}) - \nabla f(\theta_{j^{\prime\prime}- \tau_{j^{\prime\prime}}})\right) ,\nonumber\\
    & \sum\limits_{b=1}^{B}\left(\nabla F(\theta_{j^\prime-\tau_{j^{\prime}}};
      \xi_{j^\prime-\tau_{j^{\prime}},b}) - \nabla f(\theta_{j^{\prime}-\tau_{j^{\prime}}})\right) \Bigg\rangle \Bigg]\\
    = &\mathbb{E}_{\xi_j,k -1 \ge j \ge k-\tau_k}         \Bigg[\sum_{k -1\ge j^{\prime\prime} >
      j^\prime\ge k-\tau_k}\frac{\eta_{j^{\prime}} \eta_{j^{\prime\prime}}} {G_j^{\prime} G_j^{\prime\prime}} \Bigg\langle \sum\limits_{b=1}^{B}\mathbb{E}_{j^{\prime\prime,*}}\left[\nabla F (\theta_{j^{\prime\prime}-\tau_{j^{\prime\prime}} }; \xi_{j^{\prime\prime}-\tau_{j^{\prime\prime}},b}) - \nabla f(\theta_{j^{\prime\prime}-\tau_{j^{\prime\prime}}})\right ],\nonumber\\
    &  \sum\limits_{b=1}^{B}\left(\nabla F(\theta_{j^\prime-\tau_{j^{\prime}}};
      \xi_{j^\prime-\tau_{j^{\prime}},b}) - \nabla f(\theta_{j^{\prime}-\tau_{j^{\prime}}})\right) \Bigg\rangle \Bigg]\\
    =&0.
  \end{align*}
  Taking the expectation in terms of $\xi_{j-\tau_j,*}$ for $T_4$, we have
  \begin{align}
    &\mathbb{E}_{\xi_{j-\tau_j,*}|j\in \{k-\tau_k, ..., k-1\}}[T_4] \nonumber\\
    =&\mathbb{E}_{\xi_{j-\tau_j,*}|j\in \{k-\tau_k, ..., k-1\}}\left[ \left\|\sum_{j=k-\tau_k}^{k-1} \eta_j \left(\frac{1}{G_j}\right)\cdot \sum\limits_{b=1}^{B}\nabla f\left(\theta_{j-\tau_j}\right) \right\|^2 \right]
       \nonumber\\
    =&\mathbb{E}_{\xi_{j-\tau_j,*}|j\in \{k-\tau_k, ..., k-1\}}\left[ \left\|\sum_{j=k-\tau_k}^{k-1} \frac{B\eta_j}{G_j} \cdot \nabla f\left(\theta_{j-\tau_j}\right) \right\|^2 \right]
       \nonumber\\
    \leq &
           T\sum_{j=k-\tau_k}^{k-1} \frac{B^2\eta_j^2}{G_j^2} \mathbb{E}_{\xi_{j-\tau_j,*}|j\in \{k-\tau_k, ..., k-1\}}\left[\left\| \nabla f(\theta_{j-\tau_j})\right\|^2\right]
           \label{eq:t4}
  \end{align}
   where the last inequality uses the fact that $\|\sum_{i=1}^N a_i\|^2 \le N \sum_{i=1}^N \|a_i\|^2$ for any real vectors $a_i$ and the bounded age assumption \eqref{ass:bound_age}.

  We take full expectation on both sides of \eqref{eq:t1} and
  substitute $\mathbb{E}[T_{3}]$ and $\mathbb{E}[T_{4}]$ by their
  upper bounds, \eqref{eq:t3} and \eqref{eq:t4} respectively:

  \begin{align}
    \mathbb{E}[T_1] \leq 2L^2  \left( B\sum_{j=k-\tau_k}^{k-1}\frac{\eta_j^2}{G_j^2}\sigma^2 + T\sum_{j=k-\tau_k}^{k-1} \frac{B^2\eta_j^2}{G_j^2} \mathbb{E}\left[\left\|  \nabla f(\theta_{j-\tau_j})\right\|^2\right]\right)
    \label{eq:1}
  \end{align}
  Substituting $\mathbb{E}[T_1]$ and
  $\mathbb{E}[T_2]$ with their upper bounds \eqref{eq:1} and \eqref{eq:2} respectively, and taking full expectation on both sides in \eqref{eq:3}, we obtain
  \begin{align}
    \nonumber&\mathbb{E}[f(\theta_{k+1})] - f(\theta_{k})\\
    \nonumber\leq& - \frac{B\eta_k}{2G_k} \left(\mathbb{E}[\left\| \nabla f(\theta_{k}) \right\|^2] + \mathbb{E}\left[\left\| \frac{1}{B} \sum_{b=1}^B\nabla f(\theta_{k-\tau_k}) \right\|^2\right]\right)\\
    \nonumber& + \frac{B\eta_kL^2}{G_k}  \left( B\sum_{j=k-\tau_k}^{k-1}\frac{\eta_j^2}{G_j^2} \sigma^2 + T\sum_{j=k-\tau_k}^{k-1} \frac{B^2\eta_j^2}{G_j^2} \mathbb{E}\left[\left\| \nabla f(\theta_{j-\tau_j})\right\|^2\right]\right) \\
    \nonumber& + \frac{\eta^2_k L}{2G_k^2} \left( B \sigma^2 + \mathbb{E}\left[\left\|\sum_{b=1}^B \nabla f(\theta_{k-\tau_k}) \right\|^2\right]\right)\\
    \nonumber\le & - \frac{B\eta_k}{2G_k}\mathbb{E}[\left\| \nabla f(\theta_{k}) \right\|^2] + \left(\frac{B^2\eta^2_k L}{2G_k^2}   - \frac{B\eta_k}{2G_k}\right) \mathbb{E}\left[ \left\| \nabla f (\theta_{k-\tau_k}) \right\|^2\right]\\
    &+ \left(\frac{B\eta^2_k L}{2G_k^2} + \frac{B^2\eta_k L^2}{G_k} \sum_{j=k-\tau_k}^{k-1}\frac{\eta_j^2}{G_j^2} \right)\sigma^2+ \frac{B^3\eta_k L^2T}{G_k} \sum_{j=k-\tau_k}^{k-1} \frac{\eta_j^2}{G_j^2} \mathbb{E}\left[\left\|\nabla f(\theta_{j-\tau_j})\right\|^2\right] 
    \label{eq:proof:con:4}
  \end{align}
  Summarizing the inequality \eqref{eq:proof:con:4} from $k=1$ to $k=K$, meaning until the K-th update of the master, we have
  \begin{align}
    \nonumber&\mathbb{E}[f(\theta_{K+1})] - f(\theta_1)\\
    \nonumber\le & - \frac{B}{2}\sum_{k=1}^K\frac{\eta_k}{G_k} \mathbb{E}\left[  \left\| \nabla f (\theta_{k}) \right\|^2\right] + \sum_{k=1}^K \left(\frac{B^2\eta^2_k L}{2G_k^2} - \frac{B\eta_k}{2G_k}\right) \mathbb{E}\left[ \left\|\nabla f (\theta_{k-\tau_k}) \right\|^2\right]\\
    \nonumber&+ \sum_{k=1}^K\left( \frac{B\eta^2_k L}{2G_k^2} + \frac{B^2\eta_k L^2}{G_k} \sum_{j=k-\tau_k}^{k-1} \frac{\eta_j^2}{G_j^2} \right)\sigma^2\\
             \nonumber&+TL^2\sum_{k=1}^K\left(\frac{B^3\eta_k}{G_k} \sum_{j=k-\tau_k}^{k-1} \frac{\eta_j^2}{G_j^2} \mathbb{E}\left[\left\|\nabla f(\theta_{j-\tau_j}) \right\|^2\right]\right)\\
    \le \nonumber& - \frac{B}{2}\sum_{k=1}^K\frac{\eta_k}{G_k} \mathbb{E}\left( \left\| \nabla f (\theta_{k}) \right\|^2\right)\\
             \nonumber& + \sum_{k=1}^K \left( \frac{B^2\eta^2_kL}{2G_k^2} + \frac{B^3\eta_kL^2T}{G_k} \sum_{t=1}^{T}\frac{\eta_{k+t}^2}{G_{k+t}^2} - \frac{B\eta_k}{2G_k} \right) \mathbb{E}\left[  \left\| \nabla f (\theta_{k-\tau_k}) \right\|^2\right]\\
             \nonumber& + \sum_{k=1}^K\left( \frac{B\eta^2_k L}{2G_k^2} + \frac{B^2\eta_k L^2}{G_k} \sum_{j=k-\tau_k}^{k-1} \frac{\eta_j^2}{G_j^2} \right)\sigma^2\\
    \nonumber \le& - \frac{B}{2}\sum_{k=1}^K\frac{\eta_k}{G_k} \mathbb{E}\left( \left\| \nabla f (\theta_{k}) \right\|^2\right)
    + \sum_{k=1}^K\left( \frac{B\eta^2_k L}{2G_k^2} + \frac{B^2\eta_k L^2}{G_k} \sum_{j=k-\tau_k}^{k-1} \frac{\eta_j^2}{G_j^2} \right)\sigma^2\\
    \le& - \frac{B}{2}\sum_{k=1}^K\frac{\eta_k}{G_k} \mathbb{E}\left( \left\| \nabla f (\theta_{k}) \right\|^2\right)
    + \sum_{k=1}^K\left( \frac{B\eta^2_k L}{2G_k^2} + \frac{B^2\eta_k L^2}{G_k} \sum_{j=k-T}^{k-1} \frac{\eta_j^2}{G_j^2} \right)\sigma^2
    \label{eq:4}
  \end{align}
  where the second to last inequality is due to \eqref{algo1ass} and the last inequality is due to \eqref{ass:bound_age}. Note that
  $\theta^*$ is the global optimization point. By doing a few simple algebraic operations on \eqref{eq:4} we have:
  \begin{align}
    \frac{1}{\sum_{k=1}^K\frac{\eta_k}{G_k}}\sum_{k=1}^K\frac{\eta_k}{G_k} \mathbb{E}(\left\| \nabla f(\theta_{k}) \right\|^2) \leq \frac{2(f(\theta_1) -f(\theta^*) ) + \sum_{k=1}^K\left( \frac{B\eta^2_k L}{G_k^2} + \frac{2B^2\eta_k L^2}{G_k} \sum_{j=k-\tau_k}^{k+T-1} \frac{\eta_j^2}{G_j^2} \right)\sigma^2}{B\sum_{k=1}^K\frac{\eta_k}{G_k}}
  \end{align}
  This completes the proof.
\end{proof}

{\noindent \bf Proof for Corollary~\ref{cor:conv}}
\begin{proof}
  Combining \eqref{eq:coro_1} and \eqref{eq:coro_2}, we get
  \begin{align}
    \nonumber K &  \geq \frac{4BL(T+1)^2(f(\theta_1)-f(\theta^*))}{\sigma^2}\\
    \nonumber \frac{1}{4B^2L^2(T+1)^2} &  \geq \frac{(f(\theta_1)-f(\theta^*)}{BLK\sigma^2}\\
    \nonumber \frac{1}{4B^2L^2(T+1)^2} & \geq \eta^2\\
    \eta & \leq \frac{1}{2BL(T+1)}
    \label{eq:proof_con_1}
  \end{align}
  
  If follows from \eqref{eq:proof_con_1} that
  \[ B\eta L + 2B^2L^2 T^2 \eta^2 \le {\frac{1}{2T+2}} + {\frac{T^2}{2(T+1)^2}} \leq \frac{1}{2} + \frac{1}{2} = 1
  \]
  The last inequality holds since naturally $T \geq 0$. This implies that condition \eqref{algo1ass} in \Cref{thm:convergence} is satisfied globally. Then we can safely apply \eqref{eq:thm_1} in \Cref{thm:convergence}:
  \begin{align*}
    \nonumber
    {\frac{1}{K}}\sum_{i=1}^K \mathbb{E}(\|\nabla f(\theta_i)\|^2) \leq &\frac{2(f(\theta_1) -f(\theta^*) ) + K\left(B\eta^2 L +  2B^2L^2T\eta^3 \right)\sigma^2}{BK\eta}
    \\ = &
            \frac{2(f(\theta_1)-f(\theta^*))}{BK\eta} + L\sigma^2\eta \left(1 + 2BLT\eta \right)
    \\ \leq &
            \frac{2(f(\theta_1)- f(\theta^*))}{BK\eta} + 2L\sigma^2\eta
    \\  = &
            2\sqrt{\frac{(f(\theta_1)- f(\theta^*))L\sigma^2}{BK}} + 2\sqrt{\frac{(f(\theta_1)- f(\theta^*))L\sigma^2}{BK}}
    \\  = &
            4\sqrt{\frac{(f(\theta_1)- f(\theta^*))L\sigma^2}{BK}}
  \end{align*}
  where the third inequality is due to \eqref{eq:proof_con_1}
  and the second to last equality uses \eqref{eq:coro_1}. This completes the
  proof.
\end{proof}

{\noindent \bf Proof for equation in \Cref{sec:SA}}
\begin{proof}
\begin{equation}
\begin{aligned}
    \theta_{k+2} & = \theta_{k+1}-\frac{\eta_{k+1}}{\tau_{k+1}} \cdot v_{k+2}\\
    & = \theta_{k+1}-\frac{\eta_{k+1}}{\tau_{k+1}} \cdot (\gamma v_{k+1} +g_{k+1})\\
    & =\theta_{k+1}-\frac{\eta_{k+1}}{\tau_{k+1}} \cdot (\gamma (\gamma v_{k} +g_{k}) +g_{k+1} )\\
    & = \theta_{k+1}-\frac{\eta_{k+1}}{\tau_{k+1}} \cdot (\gamma^2 v_{k} +\gamma g_{k} +g_{k+1})
\end{aligned}
\end{equation}
\end{proof}

\section{Experimental Setup}
\label{sec:exp_setup}
\subsection{Algorithms}
\label{sec:ap_algorithms}
\Cref{alg:Stale-Aware_master,alg:DANA_master,alg:DANA-Stale-Aware_master,alg:DANA-Gap-Aware_master,alg:Adam_master,alg:Adam-Stale-Aware_master,alg:Adam-Gap-Aware_master} only change the master's algorithm; the complementary worker algorithm is the same as ASGD (\Cref{alg:NAG-ASGD_worker}). The master's scheme is a simple FIFO. We consider parameter server optimizations beyond the scope of this paper.

The calculation of the $C$ coefficient is described in \Cref{sec:C_coeff}.

We note that in \Cref{alg:DANA-Gap-Aware_master,alg:Adam-Gap-Aware_master} the term $\theta^i$ is identical to $\theta_{k-\tau_k}$ from the definition of $\tau_k$ in \Cref{sec:asgd}.

\begin{algorithm}
\caption{Staleness-Aware-Gradient: master}
\label{alg:Stale-Aware_master}
\begin{algorithmic}
    \setlength{\itemindent}{-0.5em} 
    \STATE Initialize an iteration array: $iter = [0]*N$
    \STATE For k = 1...K do:
    \STATE \quad Receive gradient $g^i_k$ from worker $i$
    \STATE \quad Calculate worker $i$'s current delay $\tau_k \gets k-iter[i]$
    \STATE \quad Update momentum $v_{k+1} \gets \gamma v_k+\frac{g^i_k}{\tau_k}$
    \STATE \quad Update master's weights $\theta_{k+1} \gets \theta_k-{\eta_k} \cdot v_{k+1}$
    \STATE \quad Send $\theta_{k+1}$ to worker $i$
    \STATE \quad Save current iteration $iter[i] \gets k$
\end{algorithmic}
\end{algorithm}

\begin{algorithm}[H]
    \caption{DANA: master}
    \label{alg:DANA_master}
    \begin{algorithmic}
        \setlength{\itemindent}{-0.5em} 
        \STATE For k = 1...K do:
        \STATE \quad Receive gradient $g^i_k$ from worker $i$
        \STATE \quad Update worker's momentum $v^i \gets \gamma v^i+g^i_k$
        \STATE \quad Update master's weights $\theta_{k+1} \gets \theta_k-\eta_k v^i$
        \STATE \quad Send estimate $\hat{\theta} = \theta_{k+1}-\eta_k \gamma \sum_{j=1}^{N}v^j$ to worker $i$
    \end{algorithmic}
\end{algorithm}

\begin{algorithm}[H]
    \caption{DANA-Staleness-Aware: master}
    \label{alg:DANA-Stale-Aware_master}
    \begin{algorithmic}
        \setlength{\itemindent}{-0.5em} 
        \STATE Initialize an iteration array for the workers: $iter = [0]*N$
        \STATE For k = 1...K do:
        \STATE \quad Receive gradient $g^i_k$ from worker $i$
        \STATE \quad Calculate worker $i$'s current delay $\tau_k \gets k-iter[i]$
        \STATE \quad Update worker's momentum $v^i \gets \gamma v^i+\frac{g^i_k}{\tau_k}$
        \STATE \quad Update master's weights $\theta_{k+1} \gets \theta_k-\eta_k v^i$
        \STATE \quad Send estimate $\hat{\theta} = \theta_{k+1}-\eta_k\gamma \sum_{j=1}^{N}v^j$ to worker $i$
        \STATE \quad Save current iteration $iter[i] \gets k$
    \end{algorithmic}
\end{algorithm}

\begin{algorithm}[H]
    \caption{DANA-Gap-Aware: master}
    \label{alg:DANA-Gap-Aware_master}
    \begin{algorithmic}
        \setlength{\itemindent}{-0.5em} 
        \STATE Initialize the given weights for each worker: $\theta^i = \theta_0$
        \STATE For k = 1...K do:
        \STATE \quad Receive gradient $g^i_k$ from worker $i$
        \STATE \quad Calculate Gap: $G_k = \frac{|\theta_k-\theta^i|}{C}+\mathbf{1}^d$
        \STATE \quad Update worker's momentum $v^i \gets \gamma v^i+\left(\frac{1}{G_k}\right)\odot g^i_k$
        \STATE \quad Update master's weights $\theta_{k+1} \gets \theta_k-\eta v^i$
        \STATE \quad Save and send estimate $\theta^i \gets \theta_{k+1}-\eta\gamma \sum_{j=1}^{N}v^j$ to worker $i$
    \end{algorithmic}
\end{algorithm}

The \emph{Adam}-based algorithm requires a slightly more intelligent integration to the gradient staleness penalizing methods (such as GA and SA). Penalizing the gradient before calculating the first and second moments doesn't affect the update vector since the first and second moments cancel each other out. Therefore, we suggest applying the penalty only on the first moment, thus decreasing the update step's size by the desired amount. Since DANA's integration into the Adam algorithm is convoluted and not straight forward, we chose not to implement the combination in this paper.

\begin{algorithm}[H]
\caption{Adam: master}
\label{alg:Adam_master}
\begin{algorithmic}
    \setlength{\itemindent}{-0.5em} 
    \STATE \textbf{Require:} $\eta_1 \dots \eta_K$: step lengths 
    \STATE \textbf{Require:} $\beta_1, \beta_2 \in [0,1)$: exponential decay rates for the moment estimates
    \STATE \textbf{Initialize:} $m_0 \gets 0,\quad v_0 \gets 0$
    \STATE For k = 1...K do:
    \STATE \quad Receive gradient $g^i_k$ from worker $i$
    \STATE \quad Update biased first moment estimate $m_k \gets \beta_1 m_{k-1} + (1-\beta_1)g^i_k$
    \STATE \quad Update biased second moment estimate $v_k \gets \beta_2 v_{k-1} + (1-\beta_2)(g^i_k)^2$
    \STATE \quad Compute bias-corrected first order moment estimate $\hat{m}_k \gets \frac{m_k}{1-\beta_1^k}$
    \STATE \quad Compute bias-corrected second order moment estimate $\hat{v}_k \gets \frac{v_k}{1-\beta_2^k}$
    \STATE \quad Update master's weights $\theta_{k+1} \gets \theta_k-\frac{\eta_k \cdot \hat{m}_k}{\sqrt{\hat{v}_k}+\epsilon}$
    \STATE \quad Send $\theta_{k+1}$ to worker $i$
\end{algorithmic}
\end{algorithm}

\begin{algorithm}[H]
\caption{Adam-Staleness-Aware: master}
\label{alg:Adam-Stale-Aware_master}
\begin{algorithmic}
    \setlength{\itemindent}{-0.5em} 
    \STATE \textbf{Require:} $\eta_1 \dots \eta_K$: step lengths 
    \STATE \textbf{Require:} $\beta_1, \beta_2 \in [0,1)$: exponential decay rates for the moment estimates
    \STATE \textbf{Initialize:} $m_0 \gets 0,\quad v_0 \gets 0$
    \STATE \textbf{Initialize:} $iter = [0]*N$: an iteration array for the workers
    \STATE For k = 1...K do:
    \STATE \quad Receive gradient $g^i_k$ from worker $i$
    \STATE \quad Calculate worker $i$'s current delay $\tau_k \gets k-iter[i]$
    \STATE \quad Update biased first moment estimate $m_k \gets \beta_1 m_{k-1} + (1-\beta_1)\frac{g^i_k}{\tau_k}$
    \STATE \quad Update biased second moment estimate $v_k \gets \beta_2 v_{k-1} + (1-\beta_2)(g^i_k)^2$
    \STATE \quad Compute bias-corrected first order moment estimate $\hat{m}_k \gets \frac{m_k}{1-\beta_1^k}$
    \STATE \quad Compute bias-corrected second order moment estimate $\hat{v}_k \gets \frac{v_k}{1-\beta_2^k}$
    \STATE \quad Update master's weights $\theta_{k+1} \gets \theta_k-\frac{\eta_k \cdot \hat{m}_k}{\sqrt{\hat{v}_k}+\epsilon}$
    \STATE \quad Send $\theta_{k+1}$ to worker $i$
    \STATE \quad Save current iteration $iter[i] \gets k$
\end{algorithmic}
\end{algorithm}

\begin{algorithm}[H]
\caption{Adam-Gap-Aware: master}
\label{alg:Adam-Gap-Aware_master}
\begin{algorithmic}
    \setlength{\itemindent}{-0.5em} 
    \STATE \textbf{Require:} $\eta_1 \dots \eta_K$: step lengths 
    \STATE \textbf{Require:} $\beta_1, \beta_2 \in [0,1)$: exponential decay rates for the moment estimates
    \STATE \textbf{Initialize:} $m_0 \gets 0,\quad v_0 \gets 0$
    \STATE \textbf{Initialize:} $\theta^i = \theta_0$: parameters for every worker
    \STATE For k = 1...K do:
    \STATE \quad Receive gradient $g^i_k$ from worker $i$\STATE \STATE \quad Calculate Gap: $G_k = \frac{|\theta_k-\theta^i|}{C}+\mathbf{1}^d$
    \STATE \quad Update biased first moment estimate $m_k \gets \beta_1 m_{k-1} + \left(\frac{1-\beta_1}{G_k}\right)\odot g^i_k$
    \STATE \quad Update biased second moment estimate $v_k \gets \beta_2 v_{k-1} + (1-\beta_2)(g^i_k)^2$
    \STATE \quad Compute bias-corrected first order moment estimate $\hat{m}_k \gets \frac{m_k}{1-\beta_1^k}$
    \STATE \quad Compute bias-corrected second order moment estimate $\hat{v}_k \gets \frac{v_k}{1-\beta_2^k}$
    \STATE \quad Update master's weights $\theta_{k+1} \gets \theta_k-\frac{\eta_k \cdot \hat{m}_k}{\sqrt{\hat{v}_k}+\epsilon}$
    \STATE \quad Send $\theta_{k+1}$ to worker $i$
    \STATE \quad Save worker $i$'s given parameters $\theta^i \gets \theta_{k+1}$
\end{algorithmic}
\end{algorithm}

\subsection{Datasets \& Models}
\label{sec:data_model}
\paragraph{CIFAR} The CIFAR-10 \citep{cifar} dataset comprises 60K RGB images partitioned into 50K training images and 10K test images. Each image contains 32x32 RGB pixels and belongs to 1 of 10 equal-sized classes. CIFAR-100 is similar but has 100 classes. \href{https://www.cs.toronto.edu/~kriz/cifar.html}{Link}.
\paragraph{ImageNet} The ImageNet dataset~\citep{imagenet}, known as ILSVRC2012, consists of RGB images, each labeled as one of 1000 classes. The images are partitioned into 1.28 million training images and 50K validation images; each image is randomly cropped and re-sized to 224x224 (1-crop validation). \href{http://www.image-net.org/}{Link}.
\paragraph{WikiText-103} The WikiText language modeling dataset is a collection of over 100 million tokens extracted from the set of verified \textit{Good} and \textit{Featured} articles on Wikipedia. Compared to the preprocessed version of Penn Treebank (PTB), WikiText-103 is over 110 times larger. The WikiText dataset also features a far larger vocabulary and retains the original case, punctuation, and numbers \citep{wikitext-103}. \href{https://blog.einstein.ai/the-wikitext-long-term-dependency-language-modeling-dataset/}{Link}.
\paragraph{Transformer-XL} The \emph{WikiText-103} dataset is trained on the \emph{Transformer-XL} model \citep{transformer-xl}. The hyperparameters are the ones suggested in the original paper (also see \Cref{sec:hyperparameters}) and the implementation is taken from their repository. \href{https://github.com/kimiyoung/transformer-xl}{Link}.

\subsection{Gamma Distribution}
\label{sec:gamma_distribution}
\citet{Ali:2000:TET} suggest a method called \emph{CVB} to simulate the run-time of a distributive network of computers. The method is based on several definitions:
\begin{definition} Task execution time variables:
\begin{itemize}
    \item $\mu_{task}$ - mean time of tasks
    \item $V_{task}$ - variance of tasks
    \item $\mu_{mach}$ - mean computation power of machines
    \item $V_{mach}$ - variance of computation power of machines
    \item $\alpha_{task} = \frac{1}{V_{task}^2}$
    \item $\alpha_{mach} = \frac{1}{V_{mach}^2}$
\end{itemize}
$G(\alpha, \beta)$ is a random number generated using a gamma distribution, where $\alpha$ is the shape and $\beta$ is the scale.
\end{definition}

For our case, all tasks are similar and run on a batch size of B. Therefore, the algorithm for deciding the execution-time of every task on a certain machine is reduced to one of the following:
\begin{algorithm}[H]
\caption{Task execution time - homogeneous machines}
\label{alg:task_exe_homo}
\begin{algorithmic}
    \setlength{\itemindent}{-0.5em} 
    \STATE $\beta_{task}$ = $\frac{\mu_{task}}{\alpha_{task}}$
    \STATE $q = G(\alpha_{task}, \beta_{task})$
	\STATE $\beta_{mach} = \frac{q}{\alpha_{mach}}$
    \STATE for i from 0 to $K-1$:
	\STATE $\quad time = G(\alpha_{mach}, \beta_{mach})$
\end{algorithmic}
\end{algorithm}
\begin{algorithm}[H]
\caption{Task execution time - heterogeneous machines}
\label{alg:task_exe_hetero}
\begin{algorithmic}
    \setlength{\itemindent}{-0.5em} 
    \STATE $\beta_{mach}$ = $\frac{\mu_{mach}}{\alpha_{mach}}$
    \STATE for j from 0 to $M$:
    \STATE $\quad p[j] = G(\alpha_{mach}, \beta_{mach})$ 
	\STATE $\beta_{task}[j] = \frac{p[j]}{\alpha_{task}}$
    \STATE for i from 0 to $K-1$:
	\STATE $\quad time = G(\alpha_{task}, \beta_{task}[curr])$ 
\end{algorithmic}
\end{algorithm}
where $K$ is the total amount of tasks of all the machines combined (the total number of batch iterations), $M$ is the total number of machines (workers), and $curr$ is the machine currently about to run.

We note that \Cref{alg:task_exe_homo,alg:task_exe_hetero} naturally give rise to stragglers. In the homogeneous algorithm, all workers have the same mean execution time but some tasks can still be very slow; this generally means that in every epoch a different machine will be the slowest. In the heterogeneous algorithm, every machine has a different mean execution time throughout the training. We further note that $p[j]$ is the mean execution time of machine $j$ on the average task. 

In our experiments, we simulated execution times using the following parameters as suggested by \citet{Ali:2000:TET}: $\mu_{task} = \mu_{mach} = B\cdot V_{mach}^2$, where $B$ is the batch size, yielding a mean execution time of $\mu$ simulated time units, which is proportionate to $B$. In the homogeneous setting $V_{mach} = 0.1$, whereas in the heterogeneous setting $V_{mach} = 0.6$. For both settings, $V_{task} = 0.1$.

\begin{figure*}[t]
    \centering
    \begin{subfigure}[t]{0.4\textwidth}
            \includegraphics[width=\textwidth]{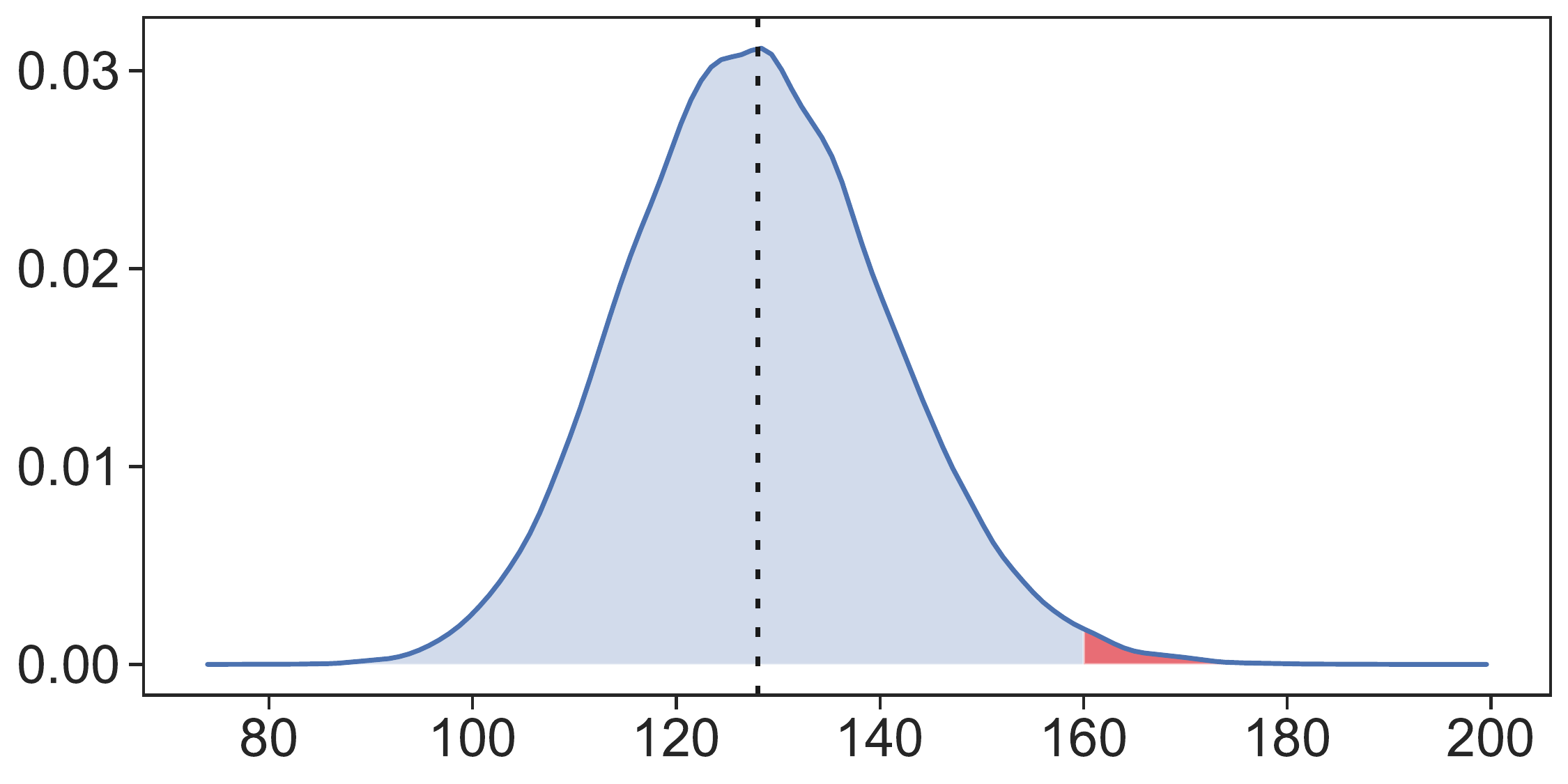}
            \caption{Homogeneous gamma-distribution}
        	\label{fig:homo_gamma}
    \end{subfigure}
    \begin{subfigure}[t]{0.4\textwidth}
            \includegraphics[width=\textwidth]{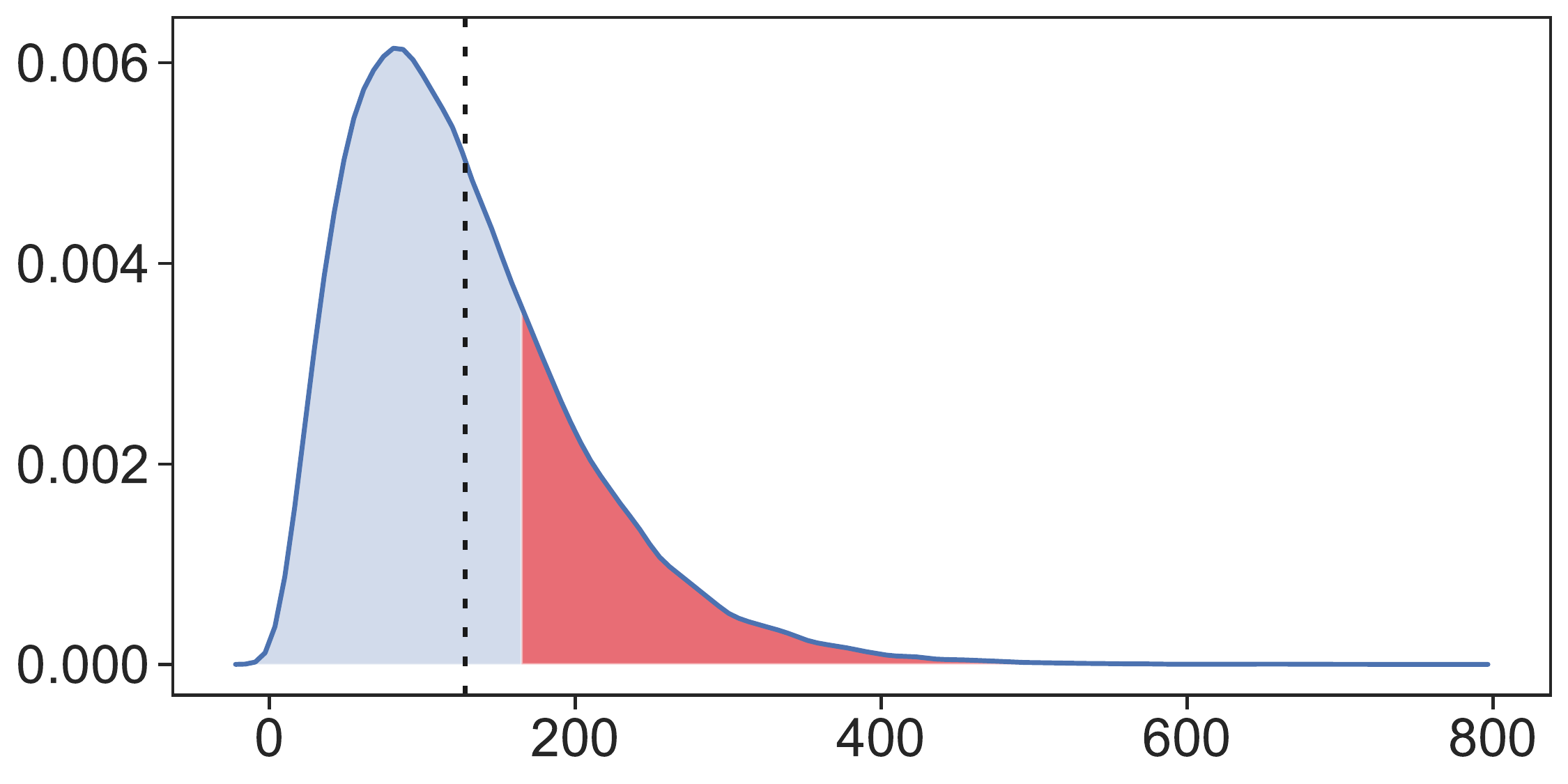}
        	\caption{Heterogeneous gamma-distribution}
        	\label{fig:hetero_gamma}
    \end{subfigure}
    \caption{Gamma-distribution in homogeneous and heterogeneous environments. The \emph{x-axis} is the simulated time units the iteration takes while the \emph{y-axis} is the probability. Both environments have the same mean (128 time units). The red area represents the probability to have an iteration which takes more than 1.25x longer than the mean iteration time.}
    \label{fig:homo_hetero_gamma}
\end{figure*}

\Cref{fig:homo_hetero_gamma} illustrates the differences between the homogeneous and heterogeneous gamma-distribution. Both environments have the same mean (128) but the probability of having an iteration that is at least 1.25x longer than the mean (which means 160 or more) is significantly higher in the heterogeneous environment (27.9\% in heterogeneous environment as opposed to 1\% in the homogeneous environment).

\subsection{Hyperparameters}
\label{sec:hyperparameters}
Since one of our intentions was to verify that penalizing the gradients linearly to the \emph{Gap} is the factor that leads to a better final test error and convergence rate, we used the same hyperparameters across all algorithms tested. These hyperparameters are the original hyperparameters of the respective neural network architecture, which are tuned for a single worker. 

\paragraph{CIFAR-10 ResNet-20}
\begin{itemize}
    \item Initial Learning Rate $\eta$: $0.1$
    \item Momentum Coefficient $\gamma$: $0.9$ with NAG
    \item Dampening: $0$ (no dampening)
    \item Batch Size $B$: $128$
    \item Weight Decay: $0.0005$
    \item Learning Rate Decay: $0.1$
    \item Learning Rate Decay Schedule: Epochs $80$ and $120$
    \item Total Epochs: $160$
\end{itemize}
We note that \cite{cifar} originally suggested $0.0001$ as the Weight Decay hyperparameter for this framework. However, when we tuned the Weight Decay on a single worker we found that $0.0005$ results in the best final test accuracy (92.43\% as opposed to 91.63\%).

\paragraph{CIFAR-10/100 Wide ResNet 16-4}
\begin{itemize}
    \item Initial Learning Rate $\eta$: $0.1$
    \item Momentum Coefficient $\gamma$: $0.9$ with NAG
    \item Dampening: $0$ (no dampening)
    \item Batch Size $B$: $128$
    \item Weight Decay: $0.0005$
    \item Learning Rate Decay: $0.2$
    \item Learning Rate Decay Schedule: Epochs $60$, $120$ and $160$
    \item Total Epochs: $200$
\end{itemize}

\paragraph{ImageNet ResNet-50}
\begin{itemize}
    \item Initial Learning Rate $\eta$: $0.1$
    \item Momentum Coefficient $\gamma$: $0.9$ with NAG
    \item Dampening: $0$ (no dampening)
    \item Batch Size $B$: $256$
    \item Weight Decay: $0.0001$
    \item Learning Rate Decay: $0.1$
    \item Learning Rate Decay Schedule: Epochs $30$ and $60$
    \item Total Epochs: $90$
\end{itemize}

\paragraph{WikiText-103 Transformer-XL}
\begin{itemize}
    \item Initial Learning Rate $\eta$: $0.00025$
    \item Dropout: $0.1$
    \item Dampening: $0$ (no dampening)
    \item Batch Size $B$: $64$
    \item First Moment Coefficient $\beta_1$: $0.9$
    \item Second Moment Coefficient $\beta_2$: $0.999$
    \item $\epsilon$: $10^{-8}$
    \item Learning Rate Decay: Cosine from $0.00025$ to $0$
    \item Gradient Clipping: $0.25$
    \item Total Iteration Steps: $K = 200000$
\end{itemize}

\paragraph{Learning Rate Warm-Up} In the early stages of training, the network changes rapidly, causing error spikes. For all algorithms, we follow the gradual warm-up approach \citep{facebook1hour} to overcome this problem: we divide the initial learning rate by the number of workers $N$ and ramp it up linearly until it reaches its original value after $5$ epochs. We also use momentum correction \citep{facebook1hour} in all algorithms to stabilize training when the learning rate changes. 

\subsection{Weight Decay}
\label{sec:weight_decay}
When GA is used with Weight Decay, the gradients contain a weight decay element which also needs to be divided by the Gap as a part of the gradient.

\subsection{C Coefficient}
\label{sec:C_coeff}
In \Cref{def:gap}, we explained that we use a coefficient $C$ to measure the gap. In \Cref{sec:gap_ver}, using the parameter-wise method, $C$ is calculated per-parameter. To calculate $C$ per-parameter, we used a weighted-average in a manner similar to the technique used in Adam \citep{adam}. The mechanism is described in \Cref{alg:coeff}.

\begin{algorithm}[H]
\caption{C Coefficient Calculation}
\label{alg:coeff}
\begin{algorithmic}
    \setlength{\itemindent}{-0.5em} 
    \STATE \textbf{Require:} $\eta_{max}$ (usually $\eta_{max} = \eta_1$) 
    \STATE \textbf{Require:} $\beta_1 \in [0,1)$: exponential decay rates for the moment estimates
    \STATE \textbf{Initialize:} $C \gets \mathbf{0^d},\quad m_0 \gets 0$
    \STATE For k = 1...K do:
    \STATE \quad Receive gradient $g^i_k$ from worker $i$
    \STATE \quad Calculate update step $v_{k+1} \gets \gamma v_k + g^i_k$
    \STATE \quad Update biased second moment estimate $m_k \gets \beta_1 m_{k-1} + (1-\beta_1)v_{k+1}^2$
    \STATE \quad Compute bias-corrected second order moment estimate $\hat{m}_k \gets \frac{m_k}{1-\beta_1^k}$
    \STATE \quad Calculate Coefficient $C \gets \eta_{max} \left( \sqrt{\hat{m}_k}+\epsilon \right)$
\end{algorithmic}
\end{algorithm}

Where is the parameter-wise method, all operations are executed per-parameter. Throughout our experiments we used $\beta_1 = 0.999$ and $\epsilon = 10^-8$ as suggested by \cite{adam}.
\subsection{Tabled Results}
\label{sec:ap_full_results}

\begin{table}[H]
\captionof{table}{ResNet-20 CIFAR10 Final Test Accuracy (Baseline $92.43\%$)}
\label{tab:CIFAR10_resnet}
\centering
\begin{tabular}{c c c c c c c c}
\toprule
N & SA & DANA-SA & NAG-ASGD & GA &  ASGD &  DANA & DANA-GA \\
\midrule
4  &  91.63$\pm$0.16 &  91.82$\pm$0.17 &  91.43$\pm$0.37 &  91.78$\pm$0.13 &   90.5$\pm$0.06 &   \textbf{92.13$\pm$0.08} &  92.06$\pm$0.24 \\
\midrule
8    &  90.63$\pm$0.14 &  90.99$\pm$0.13 &  87.68$\pm$0.35 &  91.21$\pm$0.32 &  90.28$\pm$0.21 &   91.97$\pm$0.17 &  \textbf{92.13$\pm$0.14} \\
\midrule
12   &  89.73$\pm$0.27 &  90.27$\pm$0.39 &    10.0$\pm$0.0 &  90.12$\pm$0.29 &  90.02$\pm$0.31 &    91.2$\pm$0.28 &  \textbf{91.87$\pm$0.24} \\
\midrule
16   &  88.93$\pm$0.37 &  89.48$\pm$0.06 &    10.0$\pm$0.0 &  89.59$\pm$0.39 &  89.82$\pm$0.29 &   89.68$\pm$0.41 &  \textbf{91.98$\pm$0.15 }\\
\midrule
20   &  88.21$\pm$0.32 &  88.97$\pm$0.23 &    10.0$\pm$0.0 &  89.03$\pm$0.28 &  89.55$\pm$0.13 &   88.06$\pm$0.23 &  \textbf{91.62$\pm$0.17} \\
\midrule
24   &  87.23$\pm$0.44 &   88.4$\pm$0.25 &    10.0$\pm$0.0 &   88.75$\pm$0.4 &   89.35$\pm$0.2 &   67.72$\pm$7.46 &  \textbf{91.46$\pm$0.17} \\
\midrule
28   &  86.31$\pm$0.28 &   88.0$\pm$0.27 &    10.0$\pm$0.0 &  88.18$\pm$0.18 &   88.5$\pm$0.12 &  30.72$\pm$19.74 &  \textbf{91.45$\pm$0.11} \\
\midrule
32   &  85.59$\pm$0.15 &  87.64$\pm$0.26 &    10.0$\pm$0.0 &  87.92$\pm$0.27 &   88.3$\pm$0.37 &    27.6$\pm$14.7 &  \textbf{91.15$\pm$0.32} \\
\midrule
40       &   83.57$\pm$0.5 &  86.75$\pm$0.25 &    10.0$\pm$0.0 &   86.48$\pm$0.1 &  86.61$\pm$0.54 &  22.82$\pm$13.59 &   \textbf{90.94$\pm$0.2}  \\
\midrule
48       &  80.82$\pm$0.46 &   85.6$\pm$0.28 &    10.0$\pm$0.0 &   85.0$\pm$0.66 &   83.6$\pm$0.69 &     10.0$\pm$0.0 &  \textbf{90.68$\pm$0.33} \\
\bottomrule
\end{tabular}
\end{table}

\begin{table}[H]
\captionof{table}{WideResNet 16-4 CIFAR10 Final Test Accuracy (Baseline $95.17\%$)}
\label{tab:CIFAR10_wr}
\centering
\begin{tabular}{c c c c c c c c}
\toprule
N & SA &         DANA-SA &         NAG-ASGD &       GA &            ASGD &             DANA &         DANA-GA \\
\midrule
4        &  94.41$\pm$0.16 &  94.39$\pm$0.26 &   94.81$\pm$0.11 &  94.89$\pm$0.16 &  92.99$\pm$0.15 &    95.0$\pm$0.13 &   \textbf{95.08$\pm$0.1} \\
\midrule
8        &  93.38$\pm$0.16 &  93.52$\pm$0.02 &   92.83$\pm$0.61 &  94.32$\pm$0.16 &  92.91$\pm$0.09 &   94.67$\pm$0.06 &   \textbf{94.9$\pm$0.16} \\
\midrule
12       &  92.46$\pm$0.16 &  92.68$\pm$0.21 &  44.35$\pm$28.22 &   94.02$\pm$0.1 &  92.81$\pm$0.22 &   94.27$\pm$0.15 &   \textbf{94.84$\pm$0.1} \\
\midrule
16       &  91.51$\pm$0.14 &  91.92$\pm$0.07 &  23.36$\pm$26.72 &  93.72$\pm$0.08 &  92.48$\pm$0.25 &   93.56$\pm$0.26 &  \textbf{94.77$\pm$0.19} \\
\midrule
20       &  90.62$\pm$0.17 &  91.37$\pm$0.23 &  33.41$\pm$30.31 &  93.35$\pm$0.11 &  92.28$\pm$0.31 &   92.57$\pm$0.35 &  \textbf{94.68$\pm$0.11} \\
\midrule
24       &  90.06$\pm$0.16 &   90.5$\pm$0.14 &   11.62$\pm$3.23 &  92.92$\pm$0.03 &   92.13$\pm$0.3 &   89.93$\pm$0.59 &   \textbf{94.39$\pm$0.1} \\
\midrule
28       &  89.38$\pm$0.35 &  90.18$\pm$0.17 &  31.91$\pm$17.68 &  92.56$\pm$0.08 &   91.6$\pm$0.26 &  75.32$\pm$12.14 &  \textbf{94.39$\pm$0.22} \\
\midrule
32       &   88.7$\pm$0.06 &  89.35$\pm$0.16 &  19.52$\pm$19.03 &  92.35$\pm$0.22 &  91.22$\pm$0.19 &   68.1$\pm$14.03 &   \textbf{94.27$\pm$0.1} \\
\midrule
40       &  87.26$\pm$0.21 &  88.23$\pm$0.25 &     10.0$\pm$0.0 &  91.65$\pm$0.16 &  90.53$\pm$0.24 &  28.97$\pm$25.07 &  \textbf{93.84$\pm$0.12}  \\
\midrule
48       &  85.65$\pm$0.26 &  86.64$\pm$0.78 &   12.26$\pm$4.52 &  90.97$\pm$0.38 &  89.54$\pm$0.31 &  22.31$\pm$11.43 &  \textbf{93.63$\pm$0.13} \\
\bottomrule
\end{tabular}
\end{table}

\begin{table}[H]
\captionof{table}{WideResNet 16-4 CIFAR100 Final Test Accuracy (Baseline $76.72\%$)}
\label{tab:CIFAR100_wr}
\centering
\begin{tabular}{c c c c c c c c}
\toprule
N & SA &         DANA-SA &        NAG-ASGD &       GA &            ASGD &            DANA &         DANA-GA \\
\midrule
4        &   74.74$\pm$0.3 &  \textbf{76.69$\pm$0.19} &   76.27$\pm$0.2 &  75.64$\pm$0.08 &  72.42$\pm$0.31 &  76.21$\pm$0.22 &  75.75$\pm$0.11 \\
\midrule
8        &  73.14$\pm$0.16 &  75.66$\pm$0.19 &  74.24$\pm$0.27 &  74.32$\pm$0.31 &   72.8$\pm$0.25 &  \textbf{76.03$\pm$0.13} &  75.64$\pm$0.26 \\
\midrule
12       &  71.66$\pm$0.23 &  74.73$\pm$0.38 &  69.29$\pm$0.56 &   73.86$\pm$0.3 &  72.34$\pm$0.28 &  \textbf{75.72$\pm$0.27} &   75.41$\pm$0.2 \\
\midrule
16       &  70.39$\pm$0.27 &  73.76$\pm$0.19 &  67.37$\pm$0.74 &  72.82$\pm$0.17 &   71.99$\pm$0.3 &   75.0$\pm$0.26 &  \textbf{75.01$\pm$0.17} \\
\midrule
20       &  69.51$\pm$0.23 &   72.3$\pm$0.33 &  37.98$\pm$7.21 &   72.34$\pm$0.2 &  71.63$\pm$0.18 &   73.41$\pm$0.4 &  \textbf{74.75$\pm$0.22} \\
\midrule
24       &   68.66$\pm$0.1 &   70.5$\pm$0.44 &   9.67$\pm$4.89 &  71.74$\pm$0.17 &  71.15$\pm$0.34 &  71.26$\pm$0.49 &  \textbf{74.76$\pm$0.16} \\
\midrule
28       &   67.48$\pm$0.3 &   67.67$\pm$0.3 &   6.35$\pm$7.41 &  71.22$\pm$0.34 &  70.58$\pm$0.36 &   68.7$\pm$1.25 &  \textbf{74.41$\pm$0.45} \\
\midrule
32       &  65.67$\pm$0.41 &  63.52$\pm$0.77 &  12.71$\pm$7.69 &  70.64$\pm$0.31 &  69.91$\pm$0.26 &  66.73$\pm$1.15 &  \textbf{74.33$\pm$0.28} \\
\midrule
40       &  62.31$\pm$0.34 &  66.06$\pm$0.23 &    9.56$\pm$5.1 &   69.75$\pm$0.4 &  69.25$\pm$0.48 &    64.29$\pm$1.1 &  \textbf{73.61$\pm$0.26} \\
\midrule
48       &  58.78$\pm$0.57 &  64.46$\pm$0.27 &   4.24$\pm$3.15 &  68.34$\pm$0.19 &  67.63$\pm$0.43 &  27.42$\pm$10.95 &  \textbf{73.33$\pm$0.23} \\
\bottomrule
\end{tabular}
\end{table}

\subsection{Delay vs. Gap}
\label{sec:delay_gap}
To better illustrate that the delay is usually bigger than the Gap, we measured both sizes throughout the training process using different workers. The tested framework was CIFAR100 using WideResNet 16-4. We used the Gap-Aware and DANA-GA algorithms with the same hyperparameters described in \Cref{sec:hyperparameters}. 

\begin{figure*}[t]
    \centering
    \begin{subfigure}[t]{0.32\textwidth}
            \includegraphics[width=\textwidth]{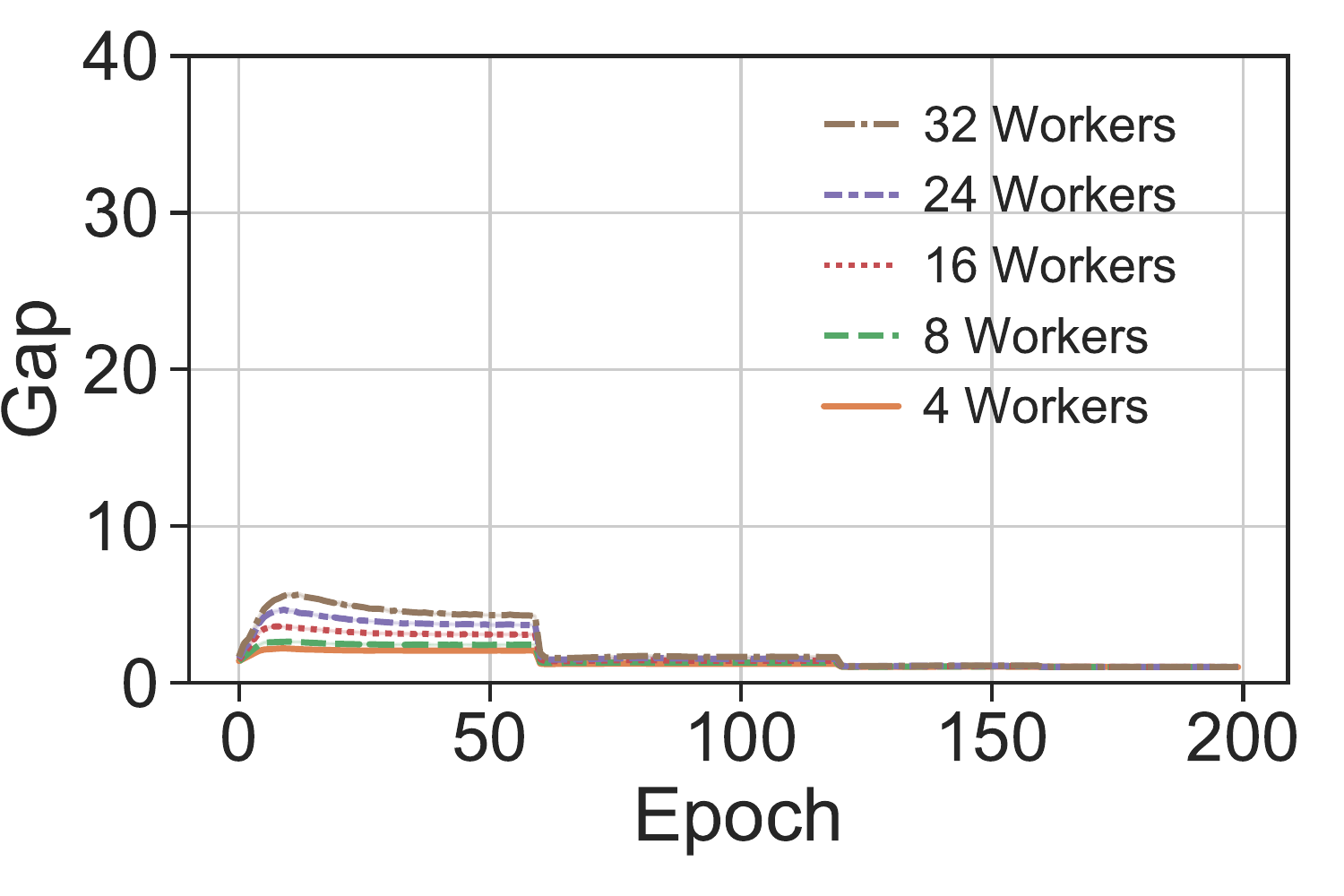}
            \caption{DANA-GA}
        	\label{fig:dana_gap_CIFAR100wr}
    \end{subfigure}
    \begin{subfigure}[t]{0.32\textwidth}
            \includegraphics[width=\textwidth]{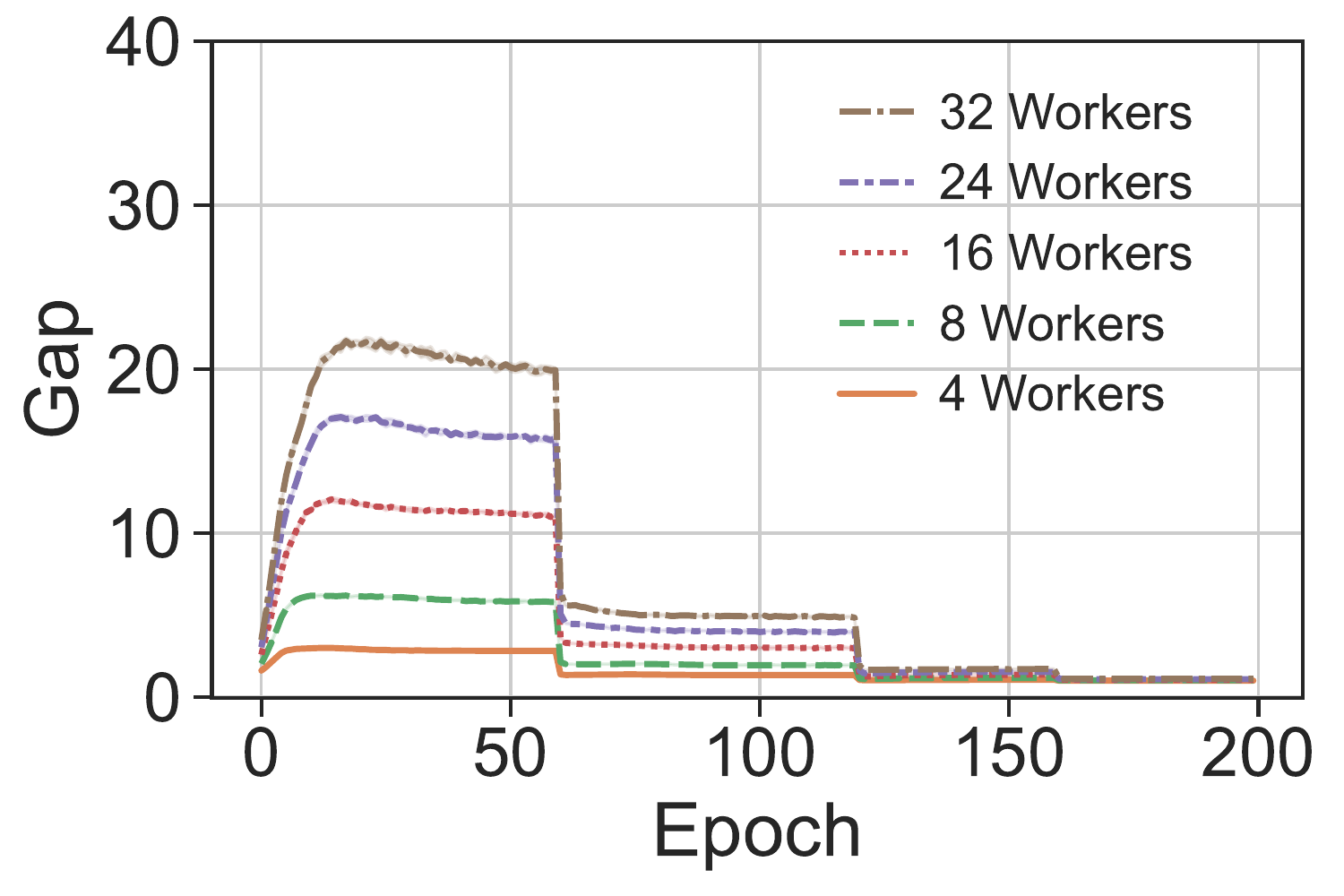}
            \caption{Gap-Aware}
        	\label{fig:gap_CIFAR100wr}
    \end{subfigure}
    \begin{subfigure}[t]{0.32\textwidth}
            \includegraphics[width=\textwidth]{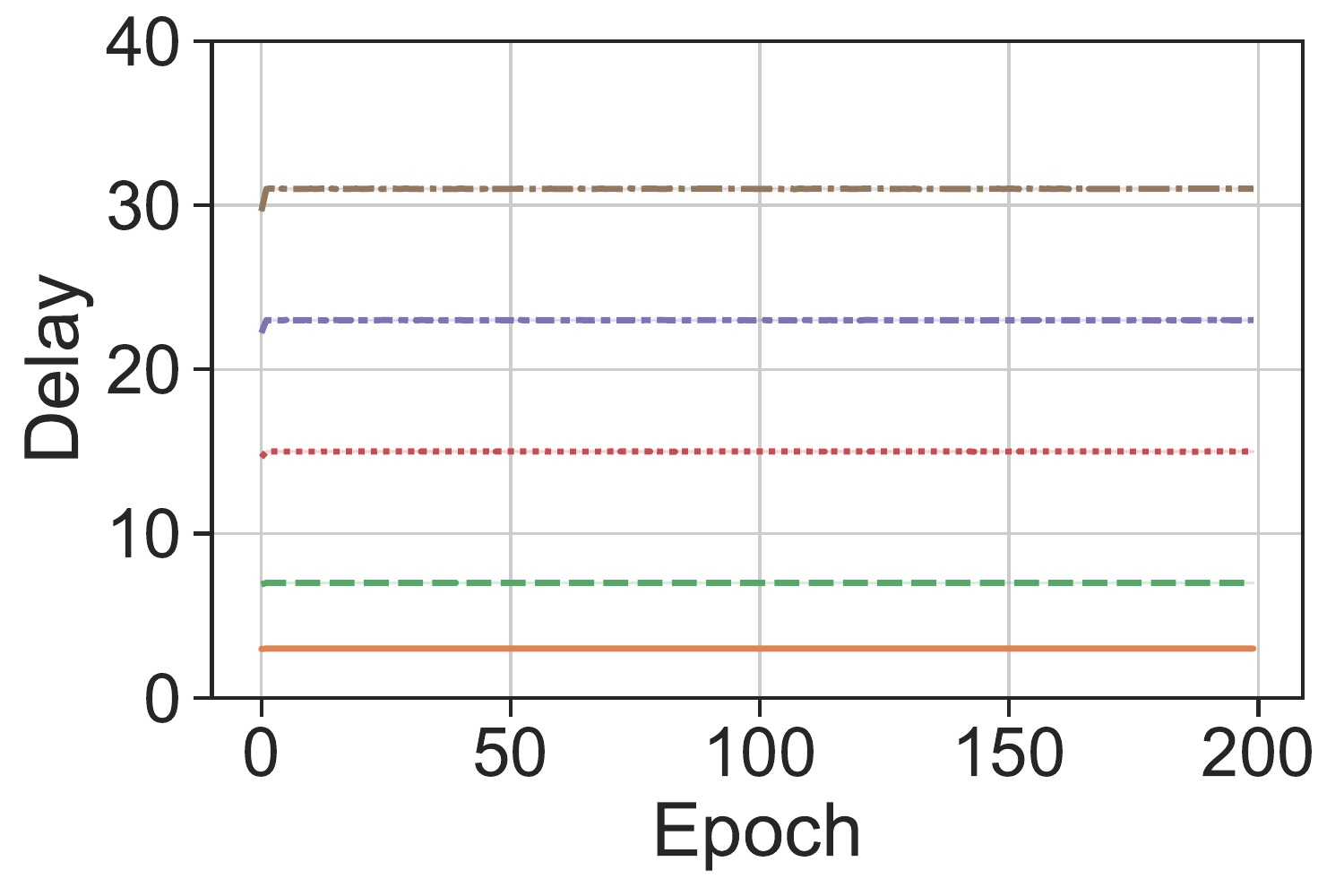}
        	\caption{Delay (SA)}
        	\label{fig:delay_CIFAR100wr}
    \end{subfigure}
    \caption{Delay and Gap throughout the training process for different number of workers using GA and DANA-GA. The figure shows the average (bold line) and standard deviation (band) of 5 runs for $N \in [4,8,16,24,32]$. All sub-figures are equally scaled to easily compare between them. GA penalizes the stale gradient much less than SA. DANA-GA penalizes the stale gradients even less than GA thanks to its approximation.}
    \label{fig:gap_vs_delay}
\end{figure*}

\Cref{fig:gap_vs_delay} shows the average Gap and Delay ($\tau)$ for every epoch throughout the training process. The Gap (\Cref{fig:gap_CIFAR100wr}) is constantly lower than the Delay (\Cref{fig:delay_CIFAR100wr}), especially towards the end of the training where the learning rate is small (which decreases the distance between the master's and worker's parameters). This figure illustrates how penalizing according to the Delay, as in Staleness-Aware, can easily over-penalize when the number of workers is large.

DANA uses an approximation of the master's parameters at the time of the future update, thus mitigating the staleness of the incoming gradients. This mitigation also reduces the Gap, as shown in \Cref{fig:dana_gap_CIFAR100wr}, which reduces the needed penalization when using DANA-GA.

\subsection{Gap Versions}
\label{sec:gap_ver_exp}
\begin{figure*}[t]
    \centering
    \begin{subfigure}[t]{0.4\textwidth}
            \includegraphics[width=\textwidth]{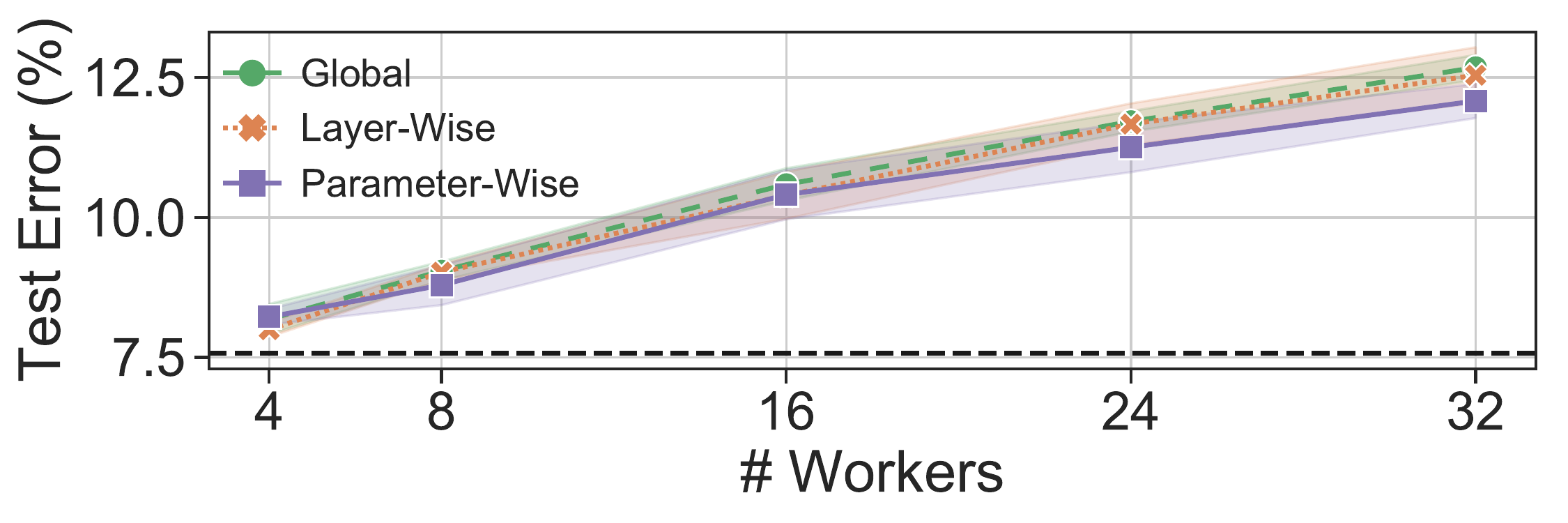}
            \caption{Test error. CIFAR10 ResNet-20}
        	\label{fig:test_gap_CIFAR10resnet}
    \end{subfigure}
    \begin{subfigure}[t]{0.4\textwidth}
            \includegraphics[width=\textwidth]{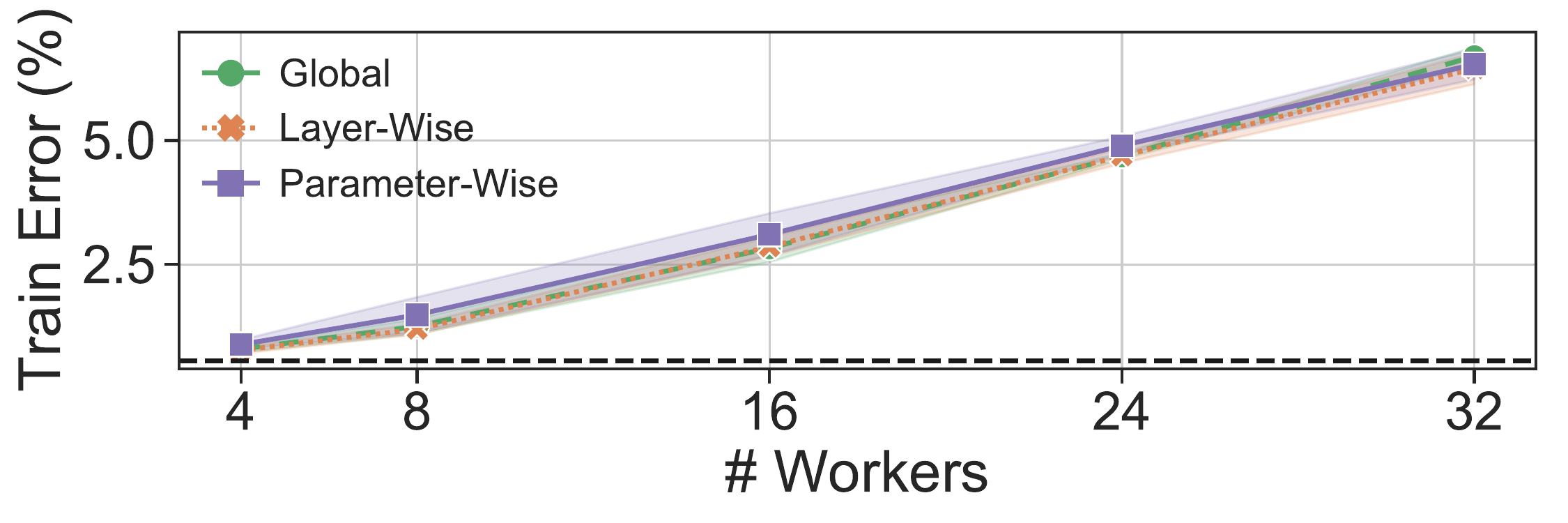}
        	\caption{Train error. CIFAR10 ResNet-20}
        	\label{fig:train_gap_CIFAR10resnet}
    \end{subfigure}
    \begin{subfigure}[t]{0.4\textwidth}
            \includegraphics[width=\textwidth]{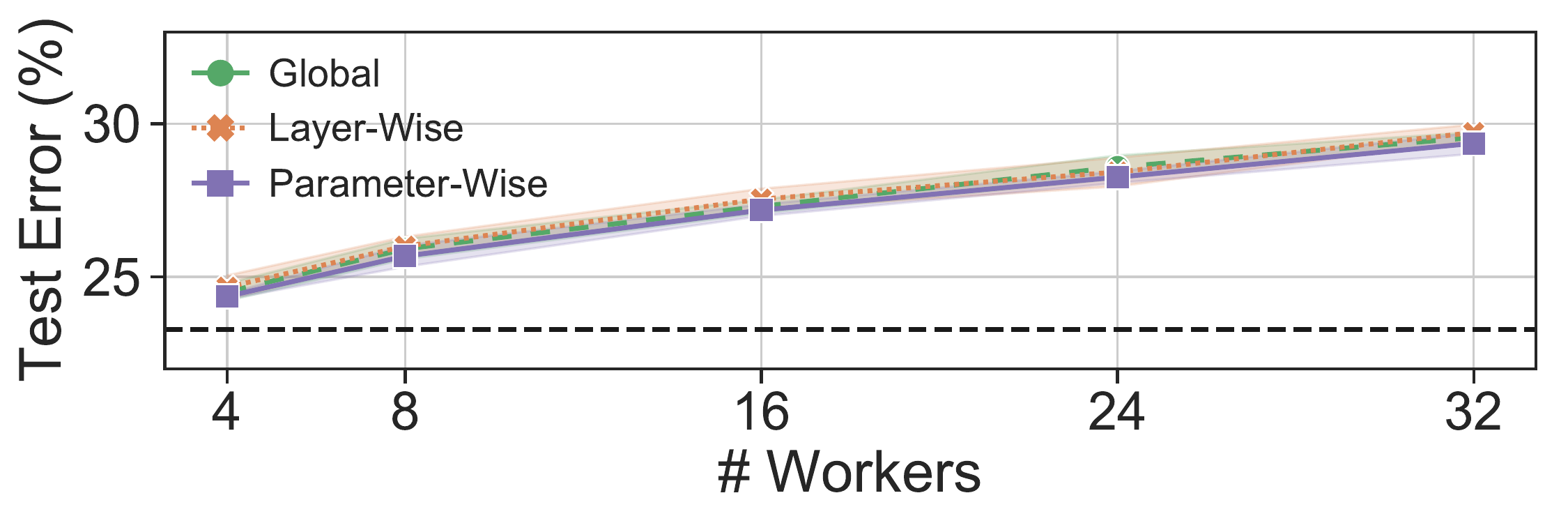}
            \caption{Test error. CIFAR100 WideResNet}
        	\label{fig:test_gap_CIFAR100wr}
    \end{subfigure}
    \begin{subfigure}[t]{0.4\textwidth}
            \includegraphics[width=\textwidth]{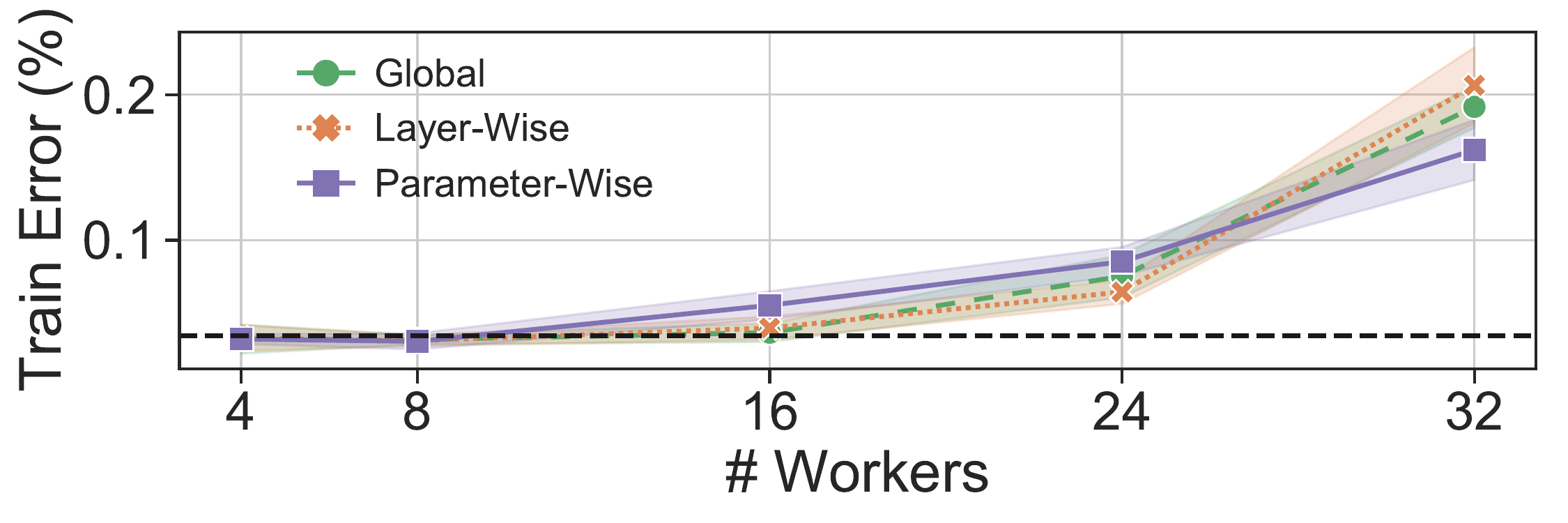}
        	\caption{Train error. CIFAR100 WideResNet}
        	\label{fig:train_gap_CIFAR100wr}
    \end{subfigure}
    \caption{Final test and train error for different numbers of workers $N$. The figure shows the average (bold line) and standard deviation (band) of 5 runs. The black dashed line is the SGD error using a single worker.}
    \label{fig:extra_gap_ver}
\end{figure*}

We tested three different Gap variations (See \Cref{sec:gap_ver}) to determine which technique has the best performance. The variations were tested on three different frameworks:
\begin{itemize}
    \item CIFAR10 using ResNet-20
    \item CIFAR10 using WideResNet 16-4
    \item CIFAR100 using WideResNet 16-4
\end{itemize}
Each framework was trained on $N \in [4,8,16,24,32]$ asynchronous workers. \Cref{fig:extra_gap_ver} contains the experiments not shown in \Cref{sec:gap_ver} and demonstrates that \emph{parameter-wise} Gap-Aware usually reaches better or similar final errors. The hyperparameters are the same ones detailed in \Cref{sec:hyperparameters}.

\subsection{CIFAR Training}
\label{sec:more_res}

\begin{figure}[t]
    \centering
    \begin{subfigure}[t]{0.32\textwidth}
            \includegraphics[width=\textwidth]{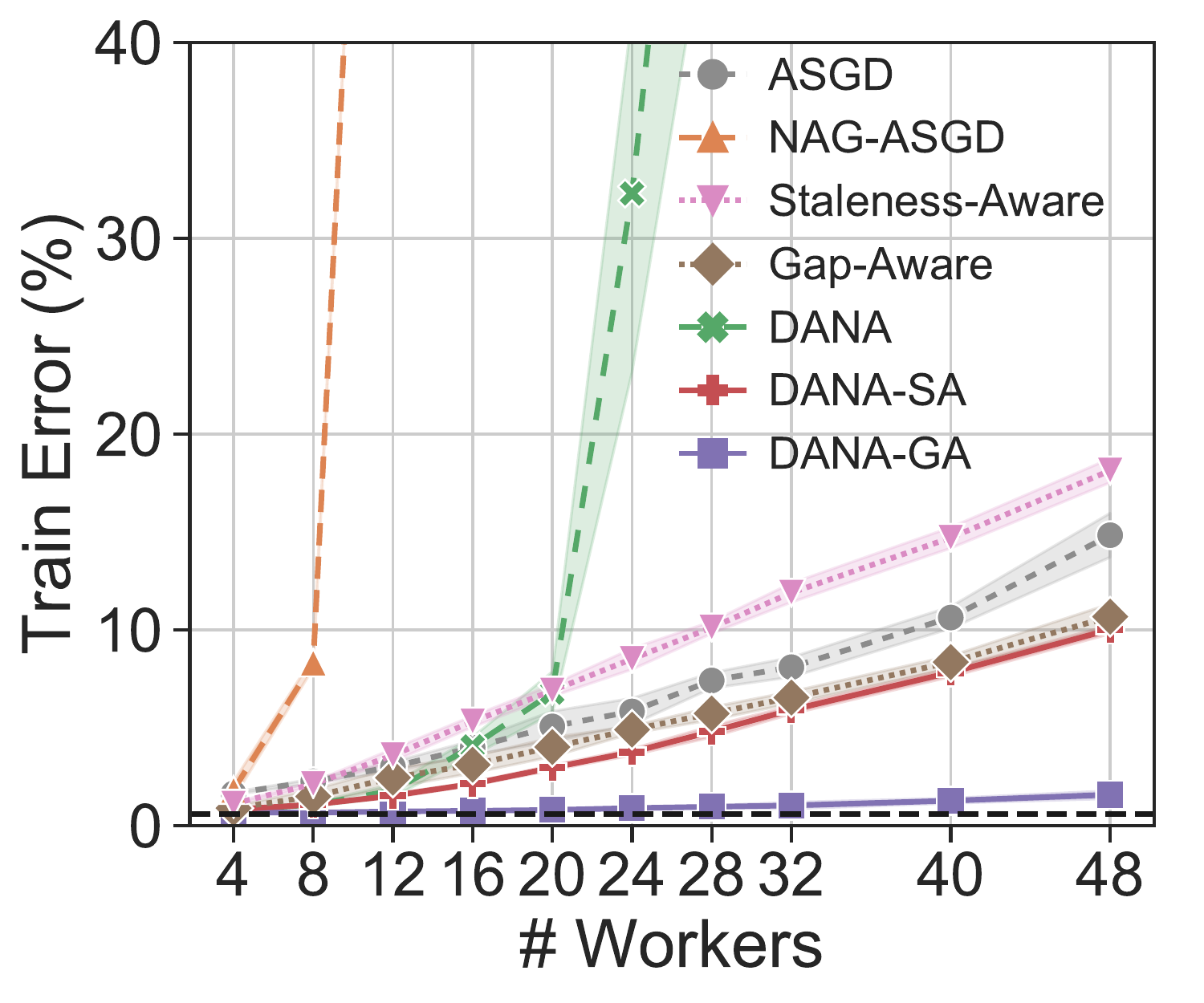}
            \caption{CIFAR10 ResNet-20}
        	\label{fig:train_CIFAR10resnet_all}
    \end{subfigure}
    \begin{subfigure}[t]{0.32\textwidth}
            \includegraphics[width=\textwidth]{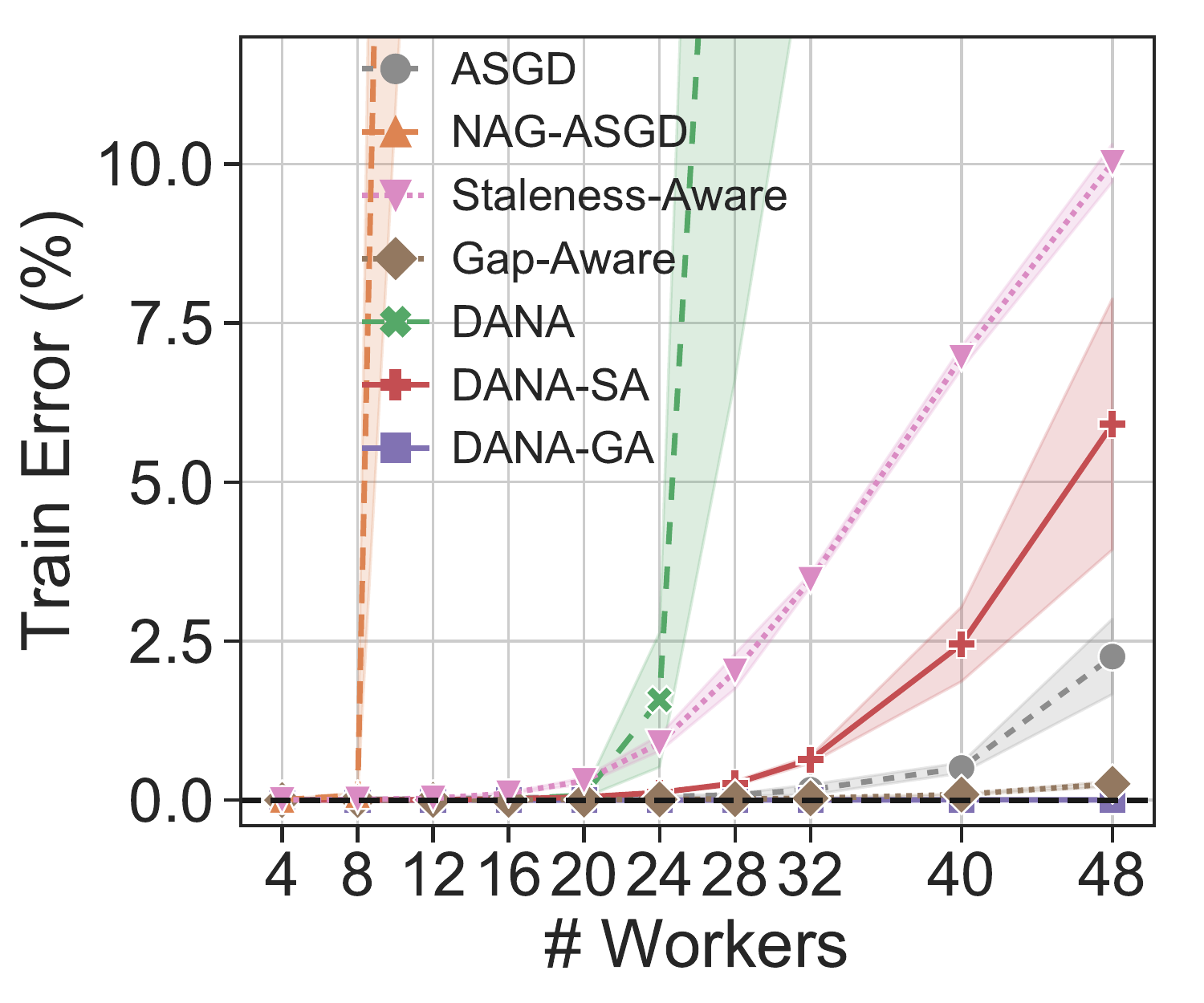}
        	\caption{CIFAR10 WideResNet}
        	\label{fig:train_CIFAR10wr_all}
    \end{subfigure}
        \begin{subfigure}[t]{0.32\textwidth}
            \includegraphics[width=\textwidth]{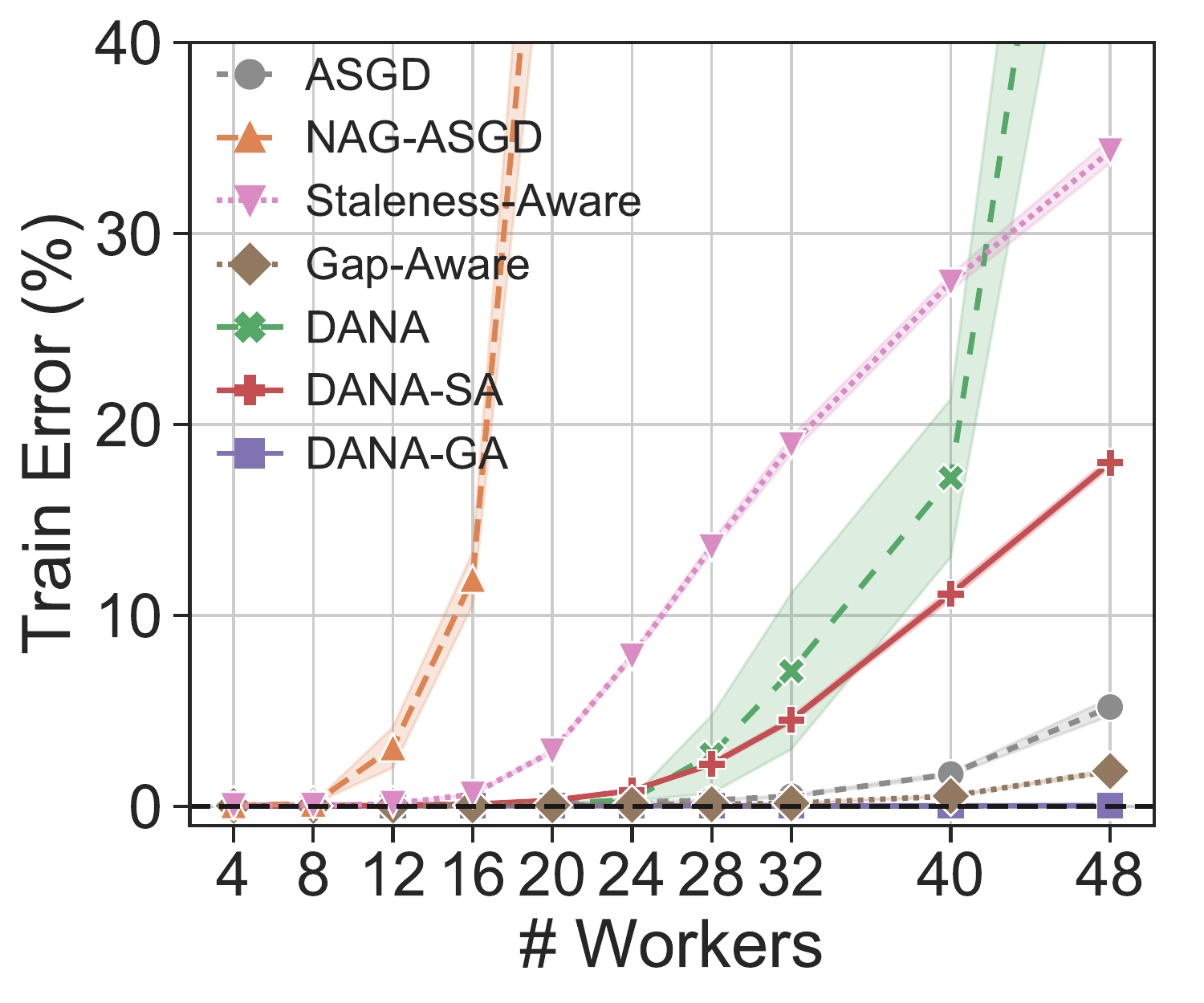}
        	\caption{CIFAR100 WideResNet}
        	\label{fig:train_CIFAR100wr_all}
    \end{subfigure}
    \caption{Final train error for different numbers of workers $N$. The figure shows the average (bold line) and standard deviation (band) of 5 runs on different frameworks.}
    \label{fig:train_all_algos}
\end{figure}
\Cref{fig:train_all_algos} shows that DANA-GA always remains very close to the zero-error region. This means that, unlike other algorithms, DANA-GA is able to converge on the training set despite the gradient staleness. \Cref{fig:train_all_algos} further demonstrates all of the concepts discussed in \Cref{sec:exp_cifar}.

\subsection{Convergence Rate}
\label{sec:conv_rate}
\begin{figure*}[t]
    \centering
    \begin{subfigure}[t]{0.32\textwidth}
            \includegraphics[width=\textwidth]{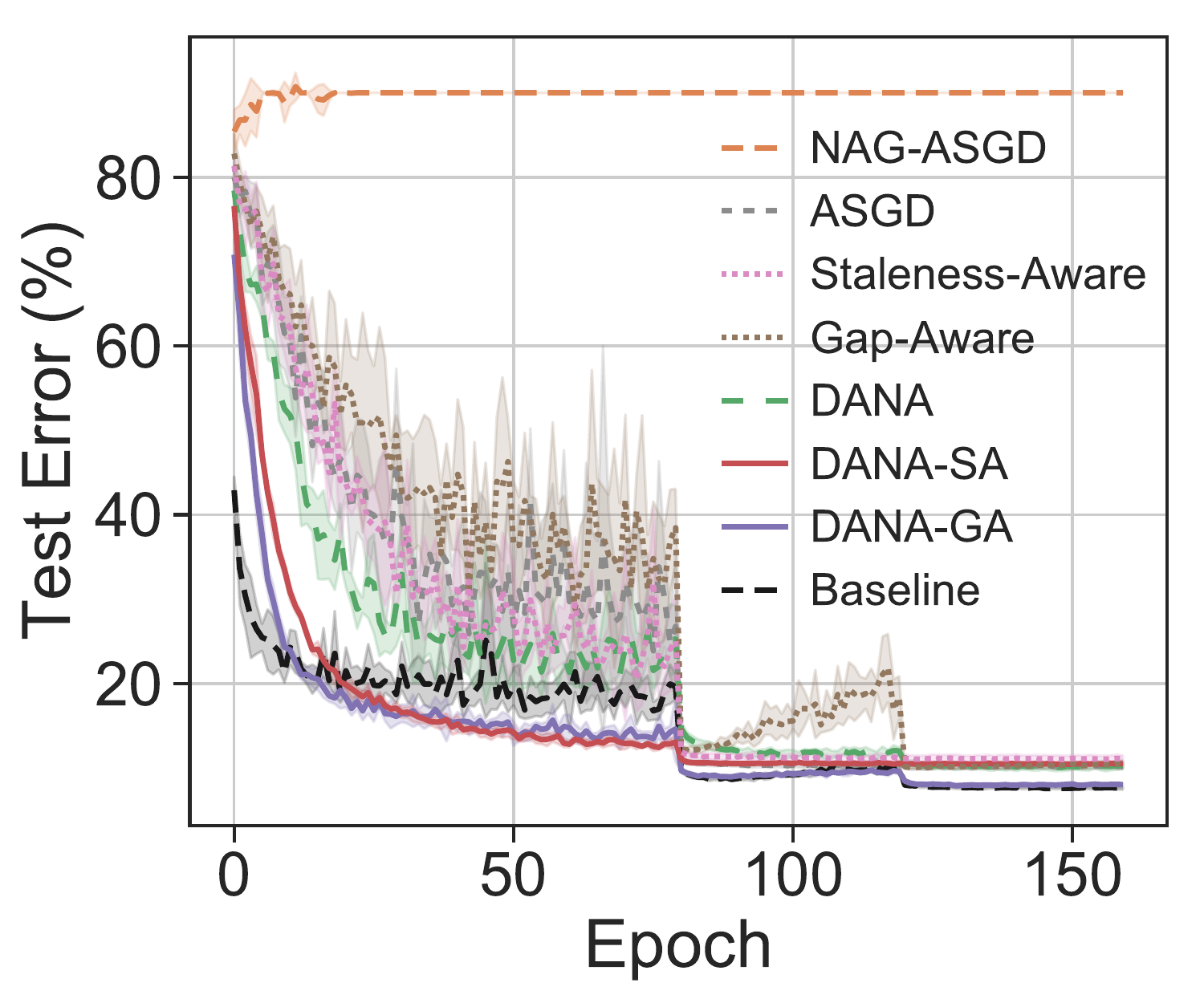}
            \caption{CIFAR10 ResNet-20}
        	\label{fig:conv_CIFAR10resnet_all}
    \end{subfigure}
    \begin{subfigure}[t]{0.32\textwidth}
            \includegraphics[width=\textwidth]{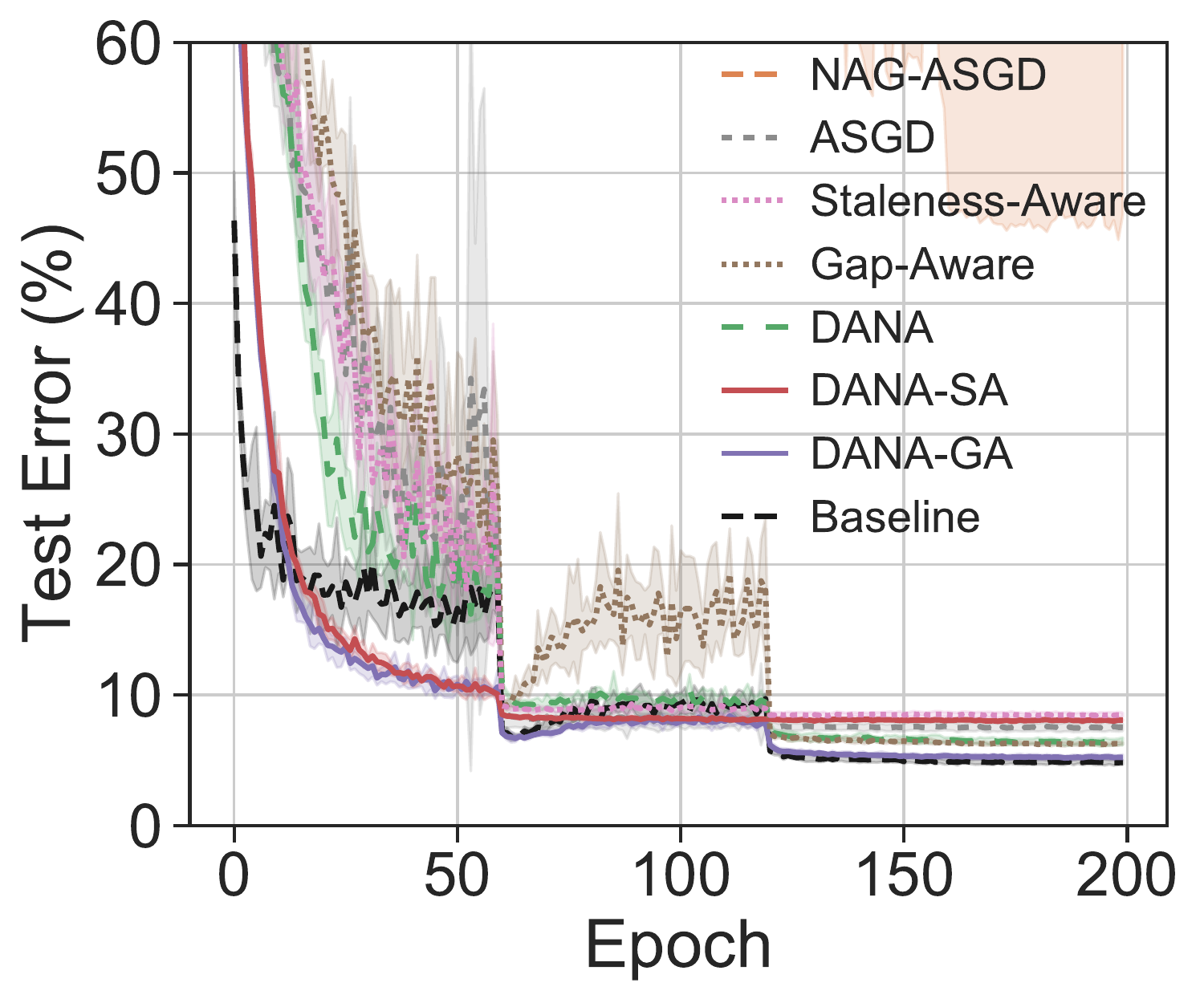}
        	\caption{CIFAR10 WideResNet}
        	\label{fig:conv_CIFAR10wr_all}
    \end{subfigure}
        \begin{subfigure}[t]{0.32\textwidth}
            \includegraphics[width=\textwidth]{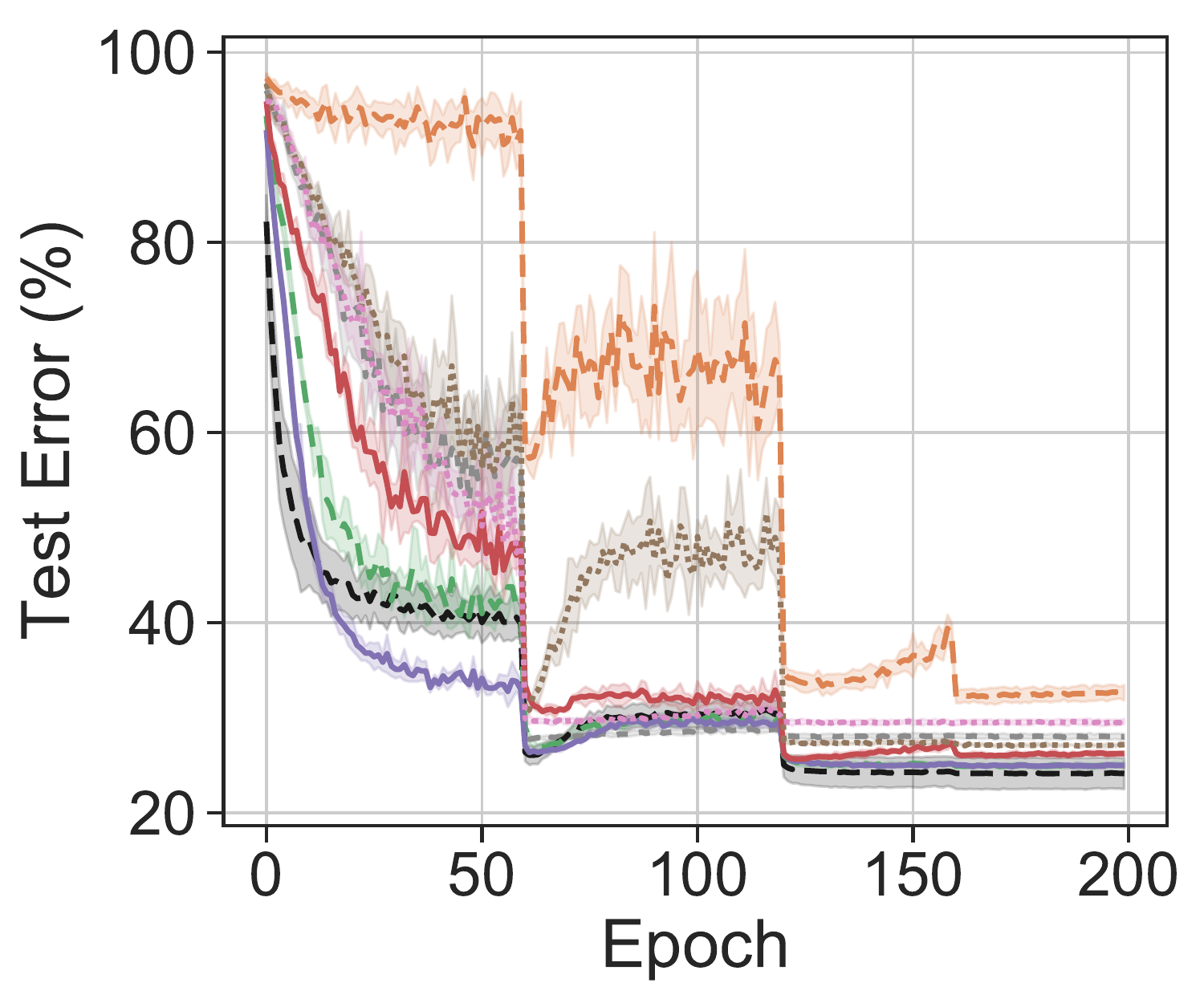}
        	\caption{CIFAR100 WideResNet}
        	\label{fig:conv_CIFAR100wr_all}
    \end{subfigure}
    \caption{Test error throughout the training using 16 workers. The figure shows the average (bold line) and standard deviation (band) of 5 runs on different frameworks.}
    \label{fig:conv_all_algos}
\end{figure*}

\begin{figure*}[t]
    \centering
    \includegraphics[width=\textwidth]{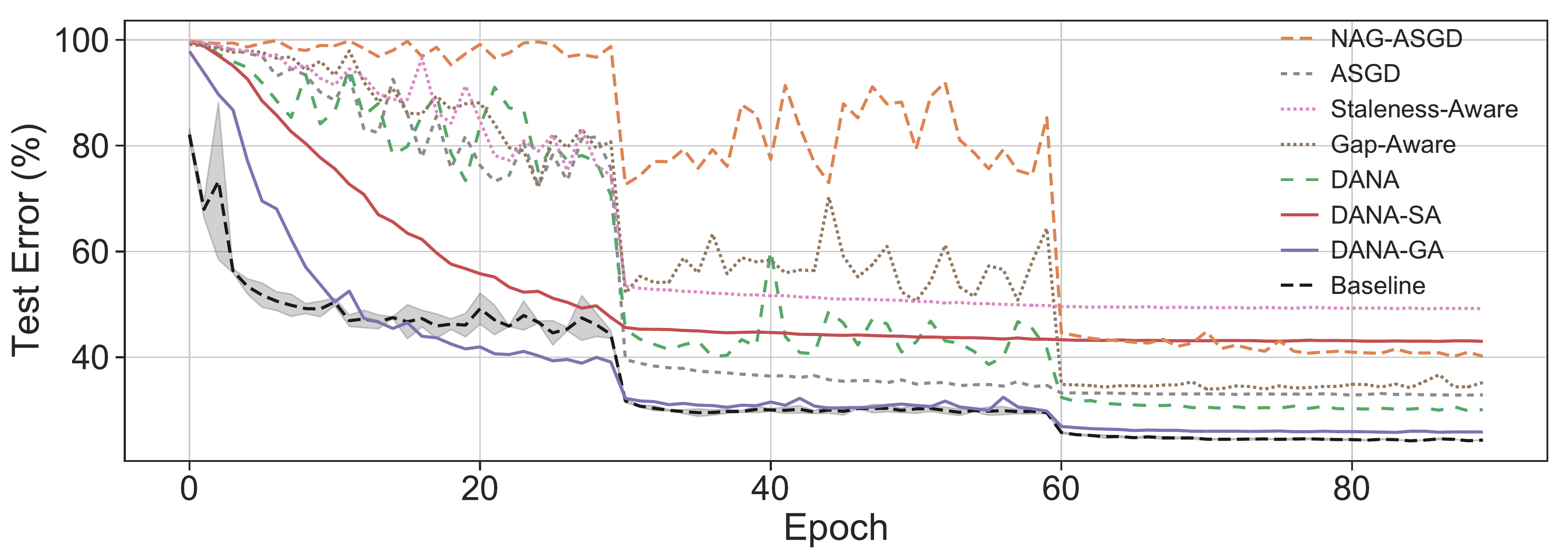}
    \caption{Test error throughout the training using 64 workers on ImageNet. DANA-GA remains close to the baseline. GA surpasses SA and DANA-SA.}
    \label{fig:imagenet_conv_all_algos}
\end{figure*}
\Cref{fig:conv_all_algos} shows that the convergence rate of GA is similar to that of SA, despite using a larger step size. This suggests that GA does not require more steps than SA to reach its (better) minima. DANA-GA's convergence rate remains very close to the baseline, which suggests that it doesn't require more steps to converge. This means that DANA-GA reaches a test error similar to the single worker case in the same number of steps, while enjoying asynchronous speedup.

\Cref{fig:imagenet_conv_all_algos} demonstrates the same ideas discussed in the last paragraph on ImageNet using 64 asynchronous workers. However, since in this case the number of workers is much larger, both SA and DANA-SA perform very poorly due the the over-penalization of SA. This helps illustrate that GA is a better staleness mitigation method than SA. DANA-GA remains very close to the baseline despite the large number of workers used, demonstrating the superiority of DANA-GA.

\subsection{Tuned ASGD}
\label{sec:app_tuned}
\begin{figure*}[t]
    \centering
    \includegraphics[width=\textwidth]{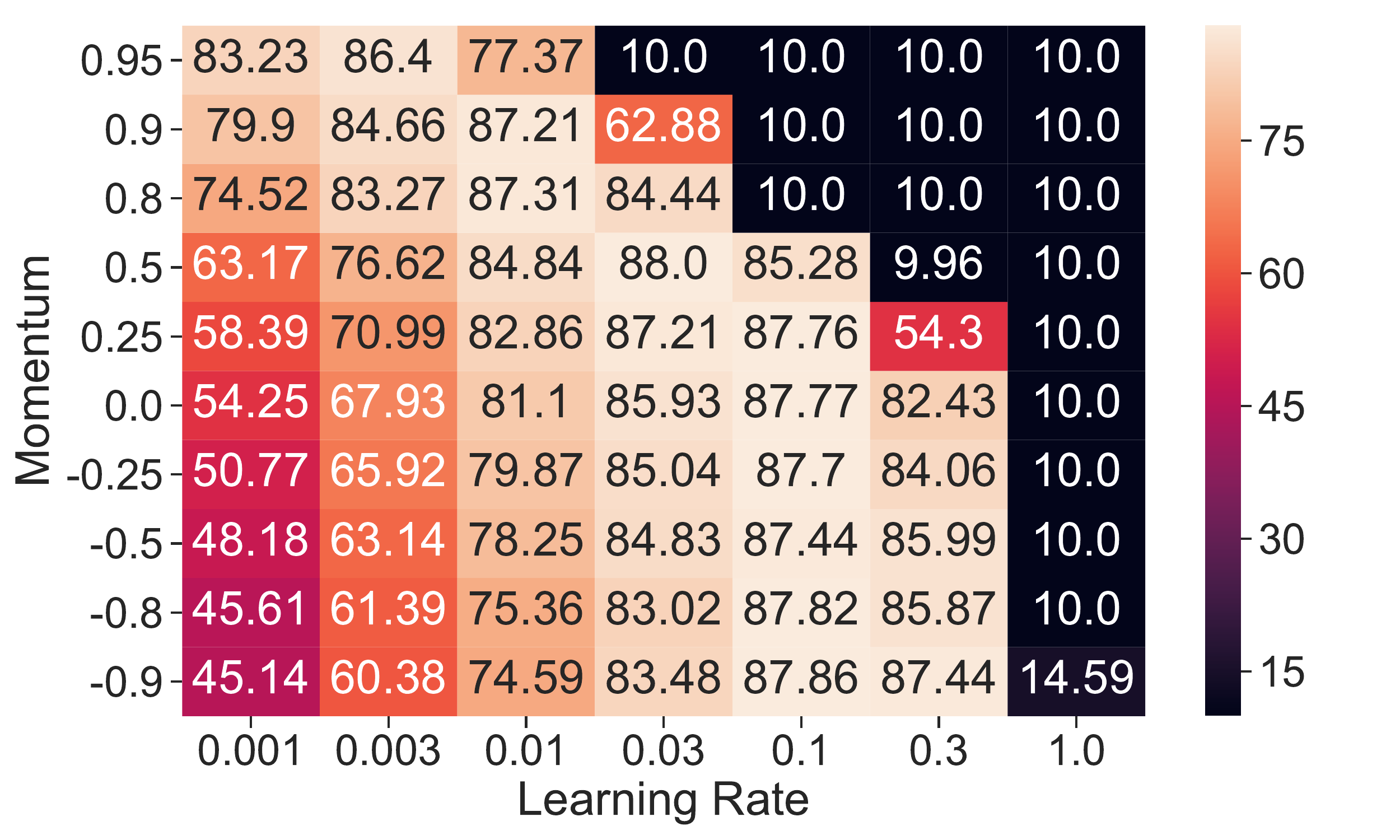}
    \caption{Test Accuracy of ASGD using 32 asynchronous workers on CIFAR10 ResNet-20 using different learning rate and momentum coefficients. The best accuracy achieved is 88\% ($\eta = 0.03, \gamma = 0.5$). The tuning includes negative values of momentum}
    \label{fig:asgd_tuning_C10_R}
\end{figure*}
\begin{figure*}[t]
    \centering
    \includegraphics[width=\textwidth]{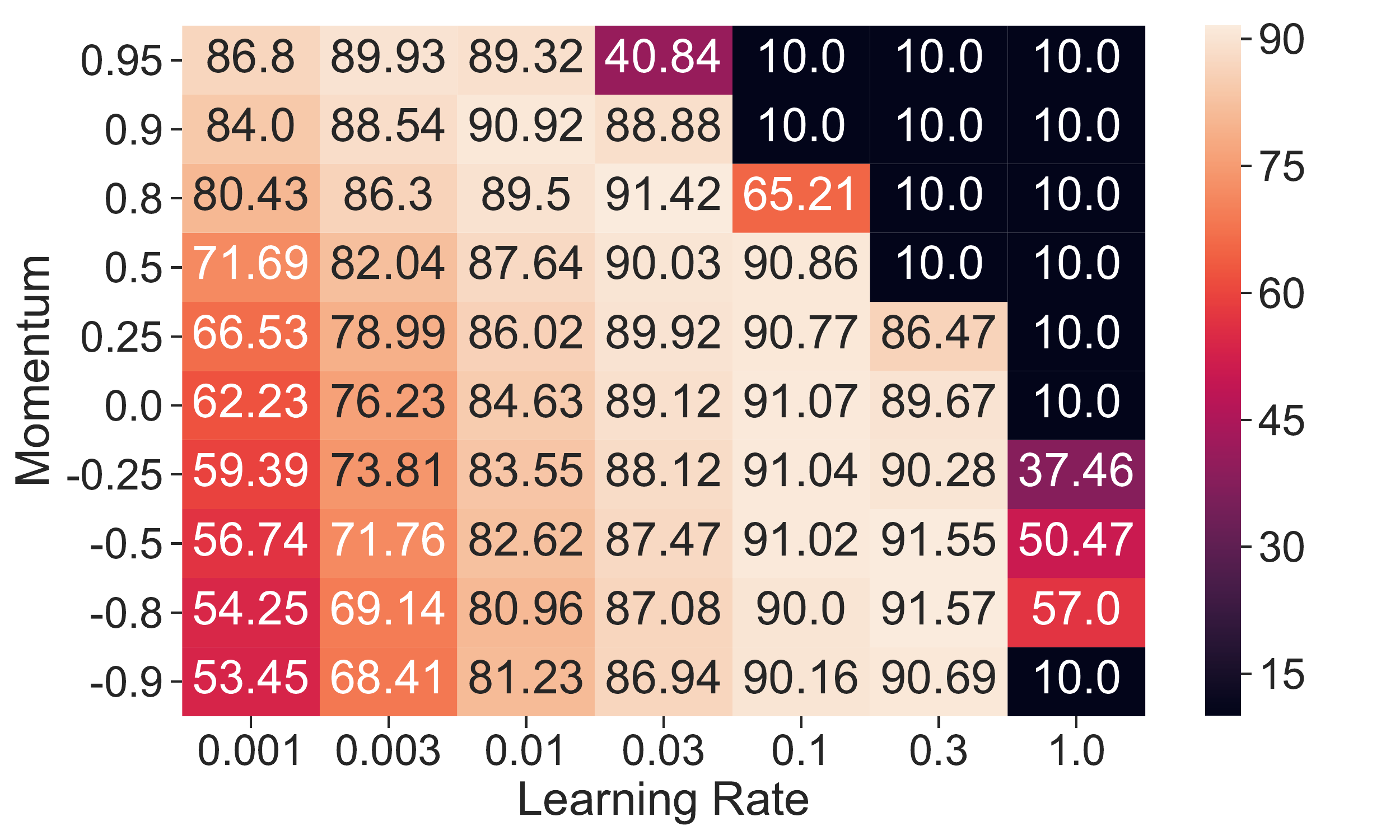}
    \caption{Test Accuracy of ASGD using 32 asynchronous workers on CIFAR10 WideResNet 16-4 using different learning rate and momentum coefficients. The best accuracy achieved is 91.57\% ($\eta = 0.3, \gamma = -0.8$). The tuning includes negative values of momentum}
    \label{fig:asgd_tuning_C10_WR}
\end{figure*}
\begin{figure*}[t]
    \centering
    \includegraphics[width=\textwidth]{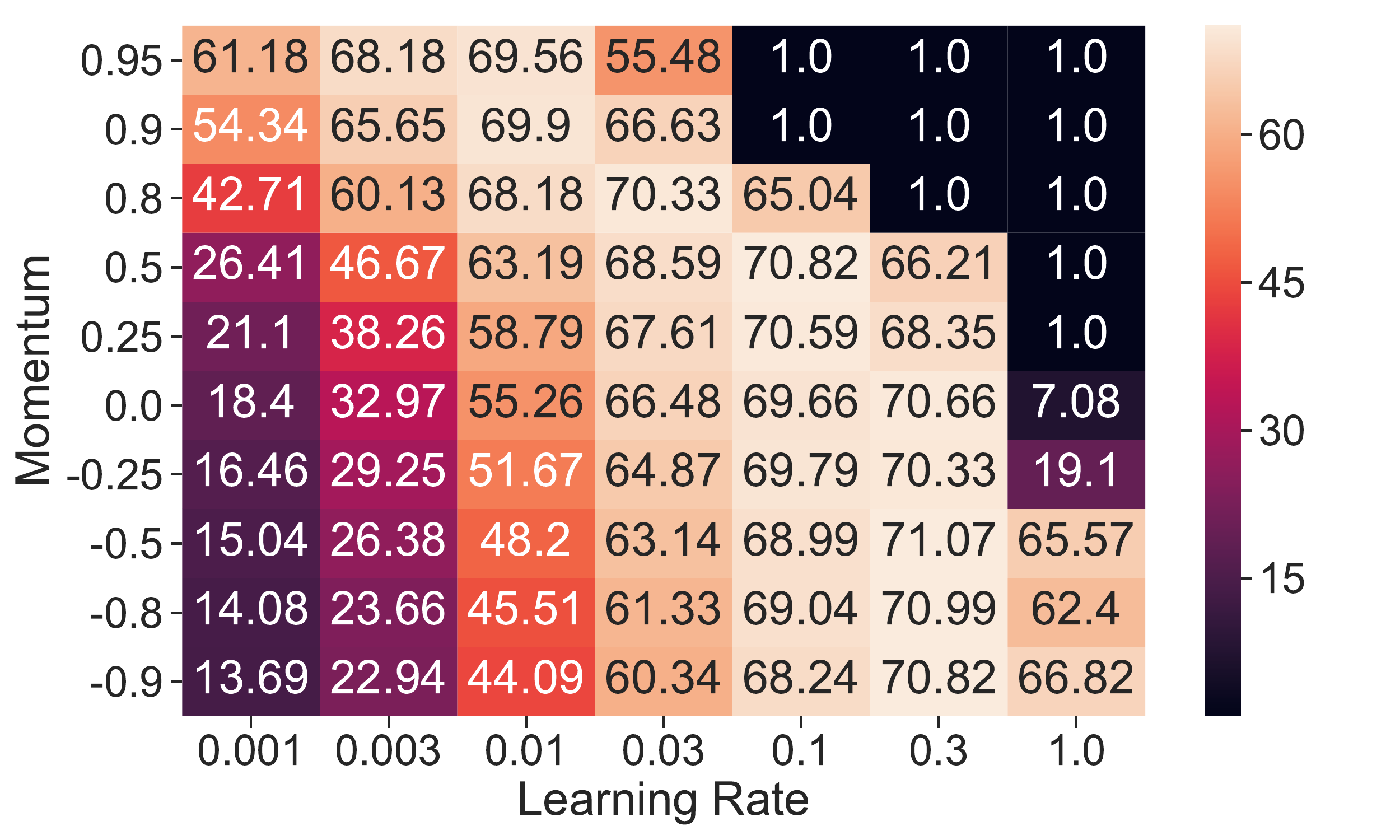}
    \caption{Test Accuracy of ASGD using 32 asynchronous workers on CIFAR10 WideResNet 16-4 using different learning rate and momentum coefficients. The best accuracy achieved is 71.07\% ($\eta = 0.3, \gamma = -0.5$). The tuning includes negative values of momentum}
    \label{fig:asgd_tuning_C100_WR}
\end{figure*}

We tuned the learning rate and momentum of ASGD on the CIFAR10 dataset with the ResNet-20 model using 32 workers.
Tuning was performed using a grid search over 70 perturbations:
\begin{equation*}
   \begin{split}
       \eta & \in [0.001, 0.003, 0.01, 0.03, 0.1, 0.3, 1]\\
       \gamma & \in [-0.9, -0.8, -0.5, -0.25, 0, 0.25, 0.5, 0.8, 0.9, 0.95]
   \end{split} 
\end{equation*}
As suggested by \citet{begets}, we also tested negative values of momentum to mitigate the implicit momentum created by the gradient staleness.
\Cref{fig:asgd_tuning_C10_R,fig:asgd_tuning_C10_WR,fig:asgd_tuning_C100_WR} show the results of the above experiments. The best final test error was given when:
\begin{itemize}
    \item \Cref{fig:asgd_tuning_C10_R}: $(\eta = 0.03, \gamma = 0.5)$
    \item \Cref{fig:asgd_tuning_C10_WR}: $(\eta = 0.3, \gamma = -0.8)$
    \item \Cref{fig:asgd_tuning_C100_WR}: $(\eta = 0.3, \gamma = -0.5)$
\end{itemize}

This shows that the best hyperparameters can vary between frameworks. The best hyperparameters for a specific framework can also vary across different number of workers as all the hyperparameters found in this experiment are different from the best hyperparameters of the single worker. Tuning for the best hyperparameters for every different number of workers for each framework significantly increases the training time and is best avoided if possible.

DANA-GA surpasses the tuned ASGD in every framework, while DANA performs very poorly using 32 workers. This proves that gradient penalization is very beneficial to overcome gradient staleness and specifically that GA is a very successful gradient penalization method. 

\subsection{Gradient Staleness Noise}
\label{sec:ap_noise}
We notice that in the ImageNet experiments (\Cref{tab:imagenet}) NAG-ASGD remains relatively close to the baseline even when the number of workers is large, as opposed to the CIFAR experiments, in which NAG-ASGD severely deteriorates as $N$ scales up. This phenomenon suggests that there is some \textit{gradient staleness noise} that a framework can "tolerate" and still perform well. Following this intuition, it is reasonable that some gradient staleness should be allowed to go "un-penalized" to avoid limiting the step-size needlessly. This idea explains why SA and GA demonstrated relatively poor results in ImageNet, especially when the number of workers was relatively small. Though we consider this analysis beyond the scope of this work, it is relevant for this paper to note that we think that the tolerable \textit{gradient staleness noise} depends on the size of the model and dataset, which suggests that GA can be further improved by correctly analysing the tolerable \textit{gradient staleness noise} and starting the penalization accordingly. We plan to continue our research in this path as well.

\section{Asynchronous Speedup}
\label{sec:async_speedup}
Cloud computing is becoming increasingly popular as a platform to perform distributed training of deep neural networks. Although synchronous SGD is currently the primary method \citep{fast_imagenet_1,fast_imagenet_2,fast_imagenet_3,facebook1hour} used to distribute the learning process, it suffers from substantial slowdowns when run in non-dedicated environments such as the cloud. This shortcoming is magnified in heterogeneous environments and non-dedicated networks. ASGD addresses the SSGD drawback and enjoys linear speedup in terms of the number of workers in both heterogeneous and homogeneous environments even in non-dedicated networks. This makes ASGD a potentially better alternative for cloud computing. 

\begin{figure}
\centering
    \begin{subfigure}[t]{0.48\textwidth}
            \includegraphics[width=\columnwidth]{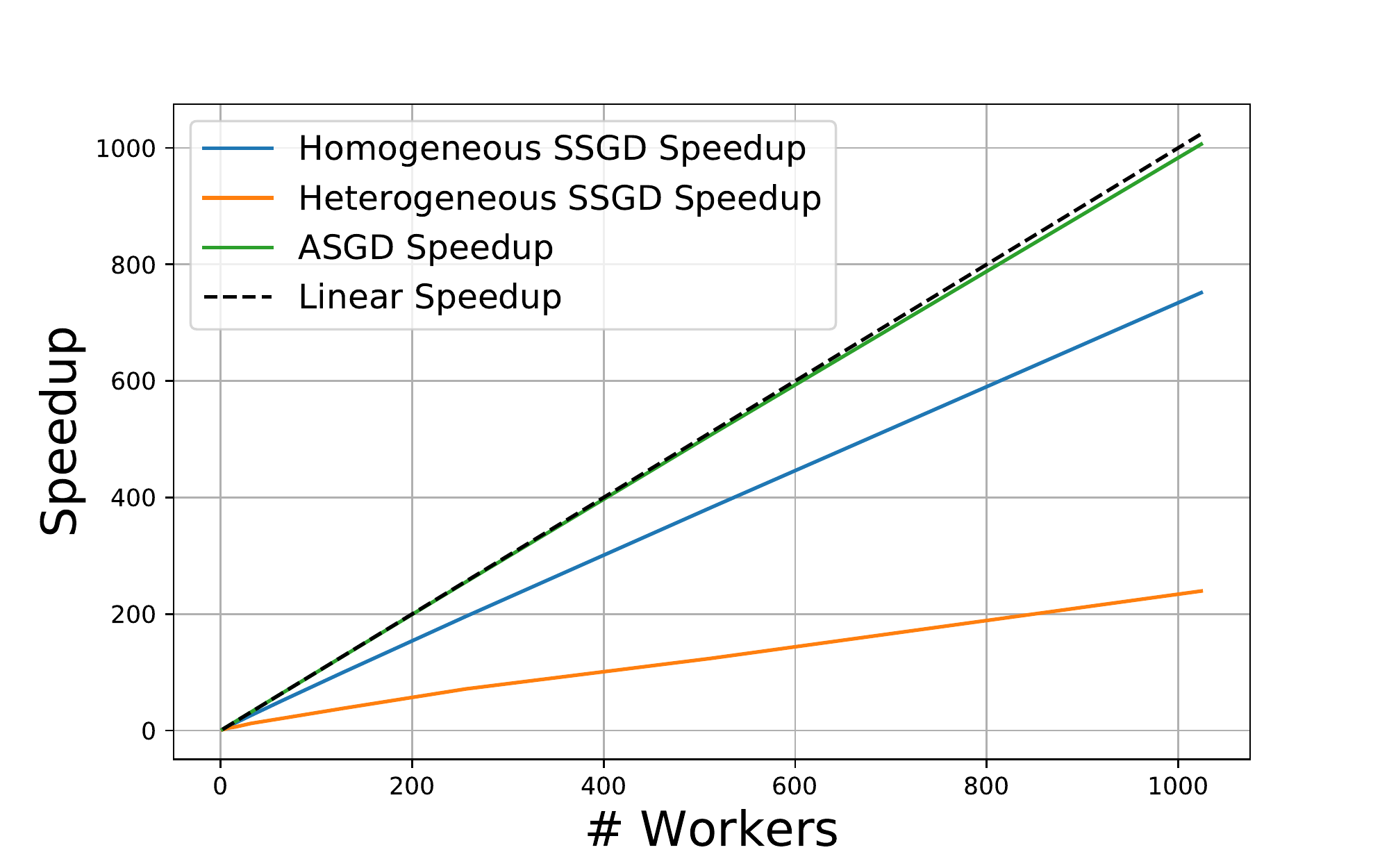}
        	\caption{Async (ASGD) and sync (SSGD) speedups.}
        	\label{fig:GammaSpeedup}
    \end{subfigure}
    \begin{subfigure}[t]{0.48\textwidth}
            \includegraphics[width=\columnwidth]{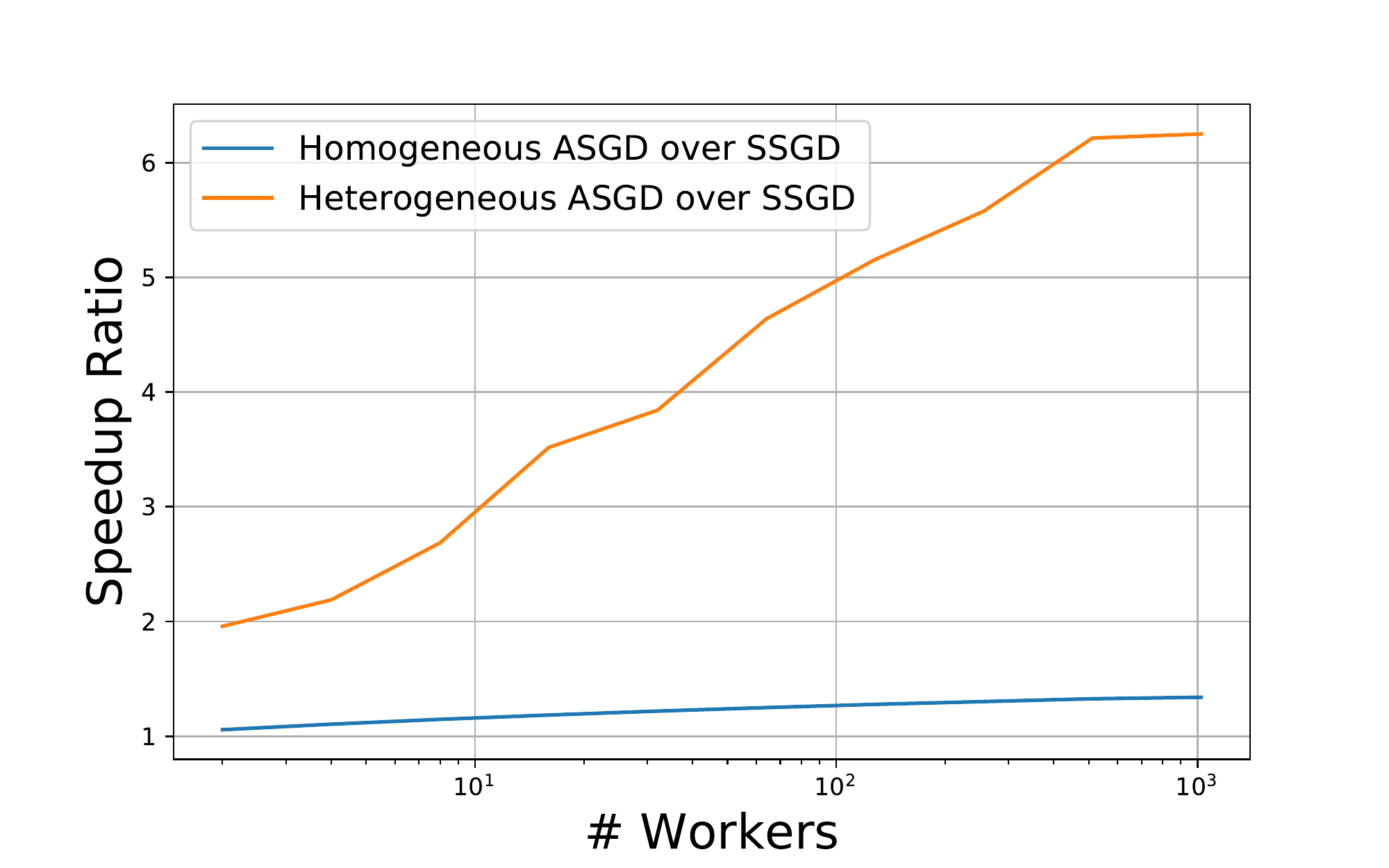}
        	\caption{ASGD speedup over SSGD. X axis is in log scale.}
        	\label{fig:GammaSpeedupRatio}
    \end{subfigure}
    \caption{Theoretical speedups for any ASGD (such as GA, SA or DANA variants) and SSGD algorithms when batch execution times are drawn from a gamma distribution. Each line is an average of 20 runs with 100000 iterations per run. Communication overheads are not modeled; however, asynchronous algorithms are more communication efficient. Accounting for the communication overheads should expand the gap between the asynchronous and synchronous training.}
\label{fig:gamma-speedup}
\end{figure}
\Cref{fig:GammaSpeedup} shows the theoretically achievable speedup, based on the detailed gamma-distributed model, for asynchronous (GA and other ASGD variants) and synchronous algorithms using homogeneous and heterogeneous workers. The asynchronous algorithms can achieve linear speedup while the synchronous algorithm (SSGD) falls short as we increase the number of workers. This occurs because SSGD must wait in each iteration until all workers complete their batch. \Cref{fig:GammaSpeedupRatio} shows that ASGD-based algorithms (including GA, SA and DANA versions) are up to $21\%$ faster than SSGD in homogeneous environments. In heterogeneous environments, ASGD methods can be 6x faster than SSGD. We note that this speedup is an underestimate, since our simulation includes only batch execution times. It does not model the execution time of barriers, all-gather operations, and other overheads which usually increase communication time, especially in SSGD.

\subsection{Gap-Aware Overhead}
The Gap-Aware method requires the master to save the parameters of every worker and adds some simple calculations mentioned in \Cref{alg:Gap_master} and \Cref{alg:coeff}. These calculations add no communication or computation requirements to the workers and introduce only a lightweight overhead to the parameter server compared to vanilla-ASGD. This lightweight overhead is similar to the overhead introduced by other ASGD-based methods such as \citet{zheng2016asynchronous}. The space requirement is not critical since the master is usually implemented in a distributed manner, and the parameters are stored in the CPU-side memory, which is usually substantially larger than the total parameter size.

\subsection{Heterogeneous Experiments}
\label{hetero_exp}
We tested the performance of GA in reference to other algorithms when the asynchronous workers are heterogeneous. The setting was very similar to the one mentioned in \Cref{sec:exp}, except that it this scenario we used the gamma-distribution to model heterogeneous workers (see \Cref{sec:gamma_distribution}).

\begin{figure*}[t]
    \centering
    \begin{subfigure}[t]{0.32\textwidth}
            \includegraphics[width=\textwidth]{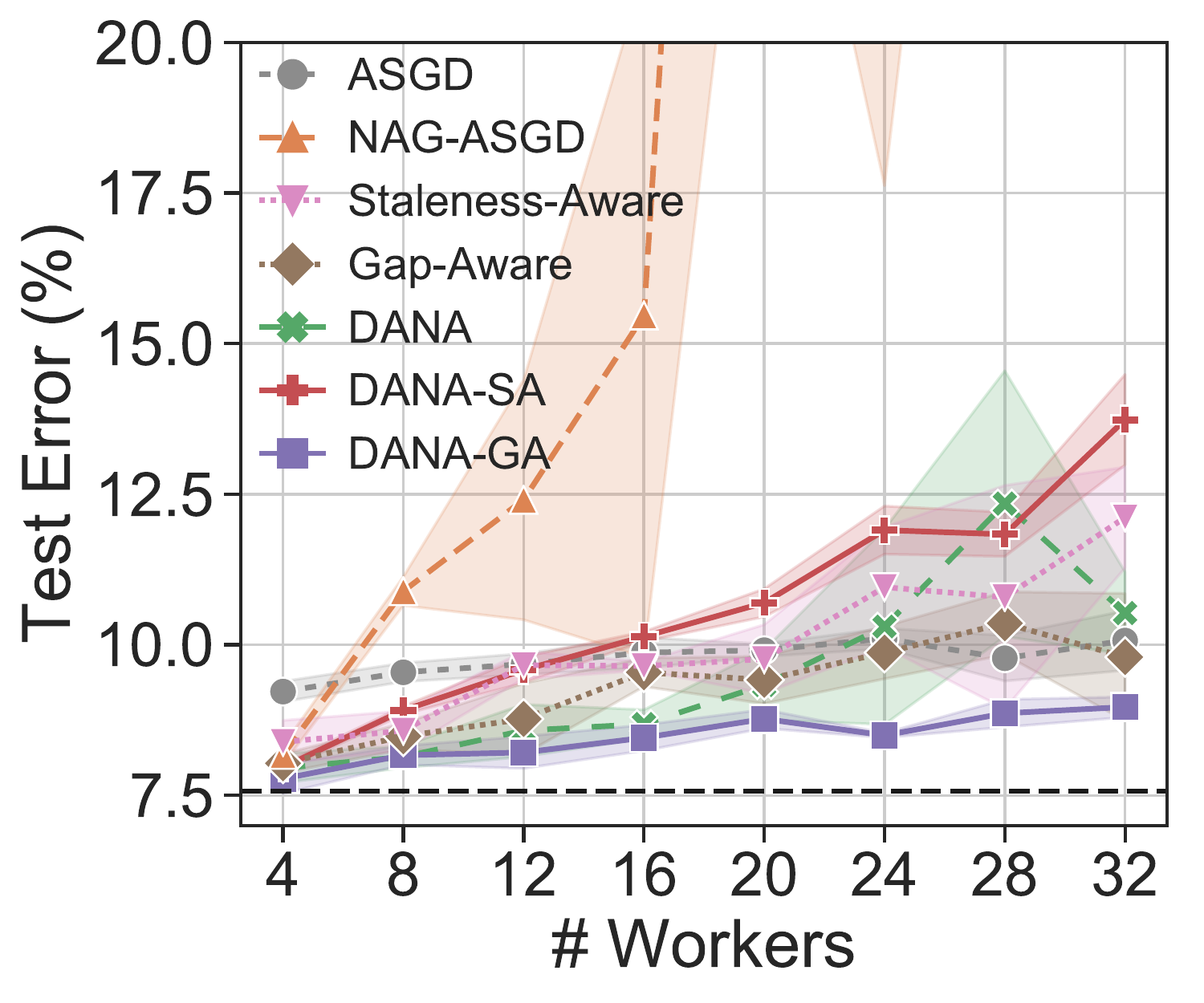}
            \caption{CIFAR10 ResNet-20}
        	\label{fig:hetero_test_CIFAR10resnet_all}
    \end{subfigure}
    \begin{subfigure}[t]{0.32\textwidth}
            \includegraphics[width=\textwidth]{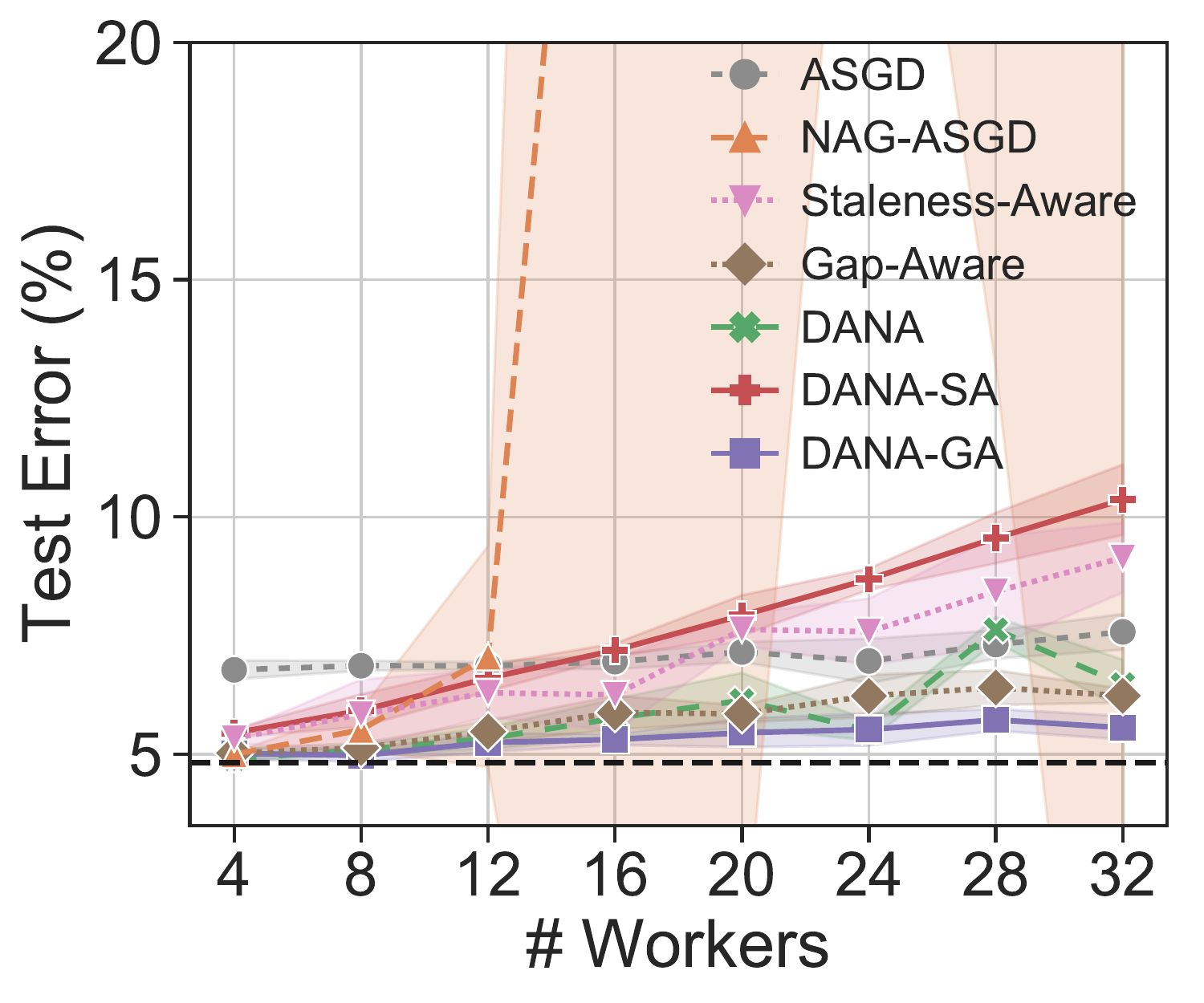}
        	\caption{CIFAR10 WideResNet}
        	\label{fig:hetero_test_CIFAR10wr_all}
    \end{subfigure}
        \begin{subfigure}[t]{0.32\textwidth}
            \includegraphics[width=\textwidth]{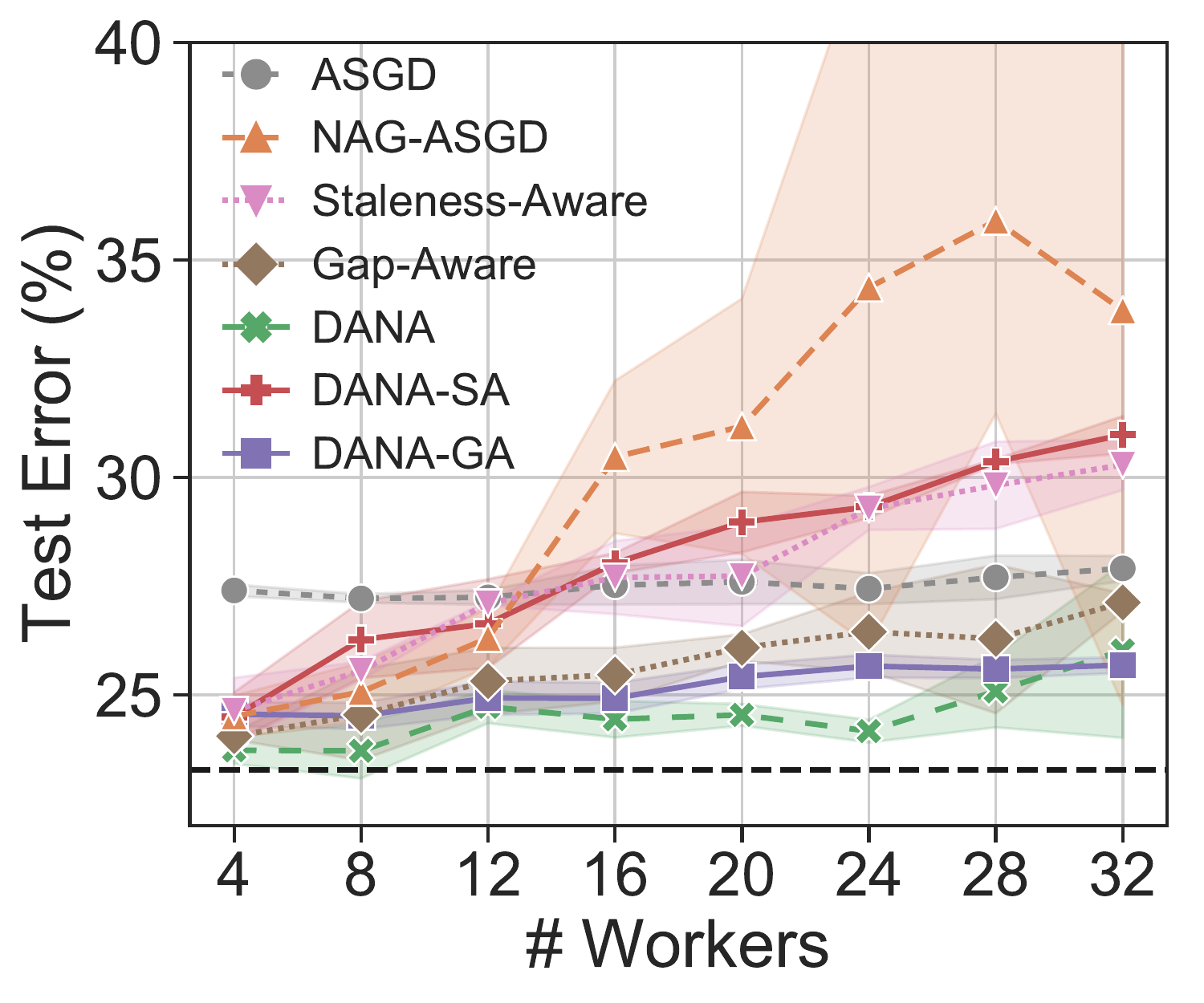}
        	\caption{CIFAR100 WideResNet}
        	\label{fig:hetero_test_CIFAR100wr_all}
    \end{subfigure}
    \caption{Final test error for different numbers of heterogeneous workers $N$. The figure shows the average (bold line) and standard deviation (band) of 5 runs on different frameworks. The black dashed line represents the average result of SGD using a single worker.}
    \label{fig:hetero_test_all_algos}
\end{figure*}

\Cref{fig:hetero_test_all_algos} demonstrates that GA and DANA-GA are superior to the other tested algorithms in heterogeneous environments as well. When comparing between \Cref{fig:hetero_test_all_algos} and \Cref{fig:test_all_algos} it is noticeable that heterogeneous environments reach a higher accuracy. This is because in heterogeneous environments some workers are very fast compared to the other ones, thus their gradients are more accurate and arrive more frequently than the slow workers' gradients. Since in cloud computing, the workers can be either heterogeneous or homogeneous, we suggest using DANA-GA to maximize the results.

\end{document}